%% file: bbo.tex
\documentclass{article}

\PassOptionsToPackage{numbers, compress}{natbib}
\PassOptionsToPackage{hyphens}{url}
\usepackage[preprint]{Style/neurips_2021}

\usepackage[utf8]{inputenc} 
\usepackage[T1]{fontenc}    
\usepackage[backref]{hyperref}       
\usepackage{url}            
\usepackage{booktabs}       
\usepackage{amsfonts}       
\usepackage{nicefrac}       
\usepackage{microtype}      

\usepackage{amsmath} 
\usepackage{amssymb}  
\usepackage{extarrows}
\usepackage{stackrel}
\usepackage{times}
\usepackage{helvet}
\usepackage{courier}
\usepackage{url}
\usepackage{blkarray}

\usepackage{amsthm}
\usepackage[font=small]{caption}
\usepackage{natbib}
\usepackage[capitalize]{cleveref}
\usepackage{autonum}
\usepackage{mathtools}
\usepackage{siunitx}
\usepackage[colorinlistoftodos]{todonotes}
\usepackage{enumitem}
\usepackage{siunitx}
\usepackage{dsfont}
\usepackage[english]{babel}
\usepackage[parfill]{parskip}
\usepackage{bm}
\usepackage{mathrsfs}
\usepackage{multicol}
\usepackage{relsize}
\usepackage{nicefrac}

\usepackage{pdflscape}
\usepackage{afterpage}
\usepackage{graphicx}
\usepackage{makecell}
\usepackage{array}
\usepackage{subcaption}
\usepackage{dblfloatfix}
\usepackage{tablefootnote}
\usepackage[flushleft]{threeparttable}
\usepackage{multirow}
\usepackage{wrapfig}
\graphicspath{{Figures/}}

\usepackage{algorithm}
\usepackage{algorithmic}
\usepackage{changepage}
\usepackage{bbm}

\newtheorem{assumption}{Assumption}

\newtheorem{theorem}{Theorem}
\newtheorem{proposition}{Proposition}
\newtheorem{lemma}{Lemma}
\newtheorem{corollary}{Corollary}[theorem]

\newtheorem{definition}{Definition}

\DeclareMathOperator*{\argmax}{arg\,max}
\DeclareMathOperator*{\argmin}{arg\,min}

\DeclarePairedDelimiterX{\infdivx}[2]{(}{)}{%
  #1\;\delimsize\|\;#2%
}
\makeatletter
\newcommand{\ostar}{\mathbin{\mathpalette\make@circled\star}}
\newcommand{\make@circled}[2]{%
  \ooalign{$\m@th#1\smallbigcirc{#1}$\cr\hidewidth$\m@th#1#2$\hidewidth\cr}%
}
\newcommand{\smallbigcirc}[1]{%
  \vcenter{\hbox{\scalebox{0.77778}{$\m@th#1\bigcirc$}}}%
}
\makeatother

\newcommand{\kl}{\textrm{KL}\infdivx}
\newcommand{\Qw}{\hat{Q}_\omega(s,a)}
\newcommand{\Qwdot}{\hat{Q}_\omega}
\newcommand{\Bp}{\hat{B}_\phi(s,a)}
\newcommand{\Bpdot}{\hat{B}_\phi}
\newcommand{\Dwn}{\mathcal{D}^N_{\omega}}
\newcommand{\msbbe}{\textrm{MSBBE}_N(\omega)}
\newcommand{\Bqw}{\mathcal{B}[\hat{Q}_\omega](s,a)}
\newcommand{\Bqwdot}{\mathcal{B}[\hat{Q}_\omega]}

\usepackage{comment}

\title{Bayesian Bellman Operators}

\author{
  
}

\author{%
Matthew Fellows\thanks{Correspondence to \texttt{matthew.fellows@cs.ox.ac.uk}} \,\,\,\, Kristian Hartikainen \,\,\,\, Shimon Whiteson\\
  Department of Computer Science\\
  University of Oxford \\
}

\begin{document}

\maketitle

\input{Sections/00_abstract}

\input{Sections/01_introduction}
\input{Sections/02_Bayesian_RL}

\input{Sections/03_Bayesian_Bellman_operators}

\input{Sections/04_Approximate_BBO}

\input{Sections/05_related_work}
\input{Sections/06_experiments}

\input{Sections/07_conclusion}

\bibliographystyle{plainnat}
\typeout{} 
\bibliography{bbo}

\section*{Checklist}

\begin{enumerate}

\item For all authors...
\begin{enumerate}
  \item Do the main claims made in the abstract and introduction accurately reflect the paper's contributions and scope?
    \answerYes{In this work we identified a major theoretical issue with existing model-free Bayesian RL approaches that claim to infer a posterior over $Q$-functions. We answered Questions 1 to 2 of \cref{sec:theoretical_issues} in \cref{sec:bbo}} and Questions 3 to 4 in \cref{sec:approximate_bbo}.
  \item Did you describe the limitations of your work?
    \answerYes{The purpose of this work is about addressing a major theoretical limitation with existing model-free Bayesian RL. We show that this limitation can have profound empirical consequences in \cref{sec:experiments}. As we have discussed, the assumptions of our theories are relatively weak and apply to a wide range of settings and function approximators used for RL including nonlinear neural networks.} 
    \item Did you discuss any potential negative societal impacts of your work?
    \answerYes{See \cref{sec:theoretical_issues} and the further discussion in \cref{app:broader_impact}}
  \item Have you read the ethics review guidelines and ensured that your paper conforms to them?
    \answerYes{}
\end{enumerate}

\item If you are including theoretical results...
\begin{enumerate}
  \item Did you state the full set of assumptions of all theoretical results?
    \answerYes{See Assumptions~\ref{ass:ergodic} to~\ref{ass:two_timescale} and our extensive discussion in \cref{app:assumptions_theorem_1,app:assumptions_theorem_2}} 
	\item Did you include complete proofs of all theoretical results?
    \answerYes{We provide a high level intuitive explanation of our theorems in the main text (see \cref{sec:theory} and \cref{sec:approximate_inference}) with rigorous and detailed proofs in \cref{app:proofs}.}
\end{enumerate}

\item If you ran experiments...
\begin{enumerate}
  \item Did you include the code, data, and instructions needed to reproduce the main experimental results (either in the supplemental material or as a URL)?
    \answerYes{Provided in the supplemental material.}
  \item Did you specify all the training details (e.g., data splits, hyperparameters, how they were chosen)?
    \answerYes{}
	\item Did you report error bars (e.g., with respect to the random seed after running experiments multiple times)?
    \answerYes{}
	\item Did you include the total amount of compute and the type of resources used (e.g., type of GPUs, internal cluster, or cloud provider)?
    \answerNo{}
\end{enumerate}

\item If you are using existing assets (e.g., code, data, models) or curating/releasing new assets...
\begin{enumerate}
  \item If your work uses existing assets, did you cite the creators?
    \answerYes{}
  \item Did you mention the license of the assets?
    \answerNA{}
  \item Did you include any new assets either in the supplemental material or as a URL?
    \answerNA{}
  \item Did you discuss whether and how consent was obtained from people whose data you're using/curating?
    \answerNA{}
  \item Did you discuss whether the data you are using/curating contains personally identifiable information or offensive content?
    \answerNA{}
\end{enumerate}

\item If you used crowdsourcing or conducted research with human subjects...
\begin{enumerate}
  \item Did you include the full text of instructions given to participants and screenshots, if applicable?
    \answerNA{}
  \item Did you describe any potential participant risks, with links to Institutional Review Board (IRB) approvals, if applicable?
    \answerNA{}
  \item Did you include the estimated hourly wage paid to participants and the total amount spent on participant compensation?
    \answerNA{}
\end{enumerate}

\end{enumerate}


\newpage
\appendix
\input{Appendix/app_broader_impact}
\input{Appendix/app_proofs}
\input{Appendix/app_derivations}

\input{Appendix/app_linear_bbo}

\input{Appendix/app_randomised_priors}
\input{Appendix/app_algorithms}
\input{Appendix/app_experiments}

\end{document}

%% file: Sections/00_abstract.tex
 \begin{abstract}
We introduce a novel perspective on Bayesian reinforcement learning (RL); whereas existing approaches infer a posterior over the transition distribution or $Q$-function, we characterise the uncertainty in the Bellman operator. Our Bayesian Bellman operator (BBO) framework is motivated by the insight that when bootstrapping is introduced, model-free approaches actually infer a posterior over Bellman operators, not value functions. In this paper, we use BBO to provide a rigorous theoretical analysis of model-free Bayesian RL to better understand its relationship to established frequentist RL methodologies. We prove that Bayesian solutions are consistent with frequentist RL solutions, even when approximate inference is used, and derive conditions for which  convergence properties hold. Empirically, we demonstrate that algorithms derived from the BBO framework have sophisticated deep exploration properties that enable them to solve continuous control tasks at which state-of-the-art regularised actor-critic algorithms fail catastrophically. 
 \end{abstract}

%% file: Sections/01_introduction.tex
\section{Introduction}
A Bayesian approach to reinforcement learning (RL) characterises uncertainty in the Markov decision process (MDP) via a posterior \citep{Ghavamzadeh15,Vlassis2012}. A great advantage of Bayesian RL is that it offers a natural and elegant solution to the exploration/exploitation problem, allowing the agent to explore to reduce uncertainty in the MDP, but only to the extent that exploratory actions lead to greater expected return; unlike in heuristic strategies such as $\varepsilon$-greedy and Boltzmann sampling, the agent does not waste samples trying actions that it has already established are suboptimal, leading to greater sampling efficiency. Elementary decision theory shows that the only admissible decision rules are Bayesian \citep{Cox74} because a non-Bayesian decision can always be improved upon by a Bayesian agent \citep{deFinetti37}. In addition, pre-existing domain knowledge can be formally incorporated by specifying priors. 

In model-free Bayesian RL, a posterior is inferred over the $Q$-function by treating samples from the MDP as stationary labels for Bayesian regression. A major theoretical issue with existing model-free Bayesian RL approaches is their reliance on bootstrapping using a $Q$-function approximator, as samples from the exact $Q$-function are impractical to obtain. This introduces error as the samples are no long estimates of a $Q$-function and their dependence on the approximation is not accounted for. It is unclear what posterior, if any, these methods are inferring and how it relates to the RL problem. 

In this paper, we introduce Bayesian Bellman Operators (BBO), a novel model-free Bayesian RL framework that addresses this issue and facilitates a theoretical exposition of the relationship between model-free Bayesian and  frequentist RL approaches. Using our framework, we demonstrate that, by bootstrapping, model-free Bayesian RL infers a posterior over \emph{Bellman operators}. For our main contribution, we prove that the BBO posterior concentrates on the true Bellman operator (or the closest representation in our function space of Bellman operators). Hence a Bayesian method using the BBO posterior is consistent with the equivalent frequentist solution in the true MDP. We derive convergent gradient-based approaches for Bayesian policy evaluation and uncertainty estimation. Remarkably, our consistency and convergence results still hold when approximate inference is used.

Our framework is general and can recover empirically successful algorithms such as BootDQNprior+ \citep{Osband18}. We demonstrate that BootDQNprior+'s lagged target parameters, which are essential to its performance, arise from applying approximate inference to the BBO posterior. Lagged target parameters cannot be explained by existing model-free Bayesian RL theory. Using BBO, we extend BootDQNprior+ to continuous domains by developing an equivalent Bayesian actor-critic algorithm. Our algorithm can learn optimal policies in domains where state-of-the-art actor-critic algorithms like soft actor-critic \citep{Haarnoja18} fail catastrophically due to their inability to properly explore.

%% file: Sections/02_Bayesian_RL.tex
\vspace{-0.2cm}
\section{Bayesian Reinforcement Learning}

\subsection{Preliminaries} Formally, an RL problem is modelled as a Markov decision process (MDP) defined by the tuple $\langle \mathcal{S},\mathcal{A},r,P,P_0,\gamma \rangle$ \citep{Szepesvari10, Puterman94}, where $\mathcal{S}$ is the set of states and $\mathcal{A}$ the set of available actions. At time $t$, an agent in state $s_t\in \mathcal{S}$ chooses an action $a_t\in \mathcal{A}$ according to the policy $a_t\sim\pi(\cdot\vert s_t)$. The agent transitions to a new state according to the state transition distribution $s_{t+1}\sim P(\cdot\vert s_t,a_t)$ which induces a scalar reward $r_t\coloneqq r(s_{t+1},a_t,s_t)\in\mathbb{R}$ with $\sup_{s',a,s}\lvert r(s',a,s) \rvert<\infty$. The initial state distribution for the agent is $s_0\sim P_0$ and the state-action transition distribution is defined as $P^\pi(s',a'\vert s,a)\coloneqq \pi(a'\vert s')P(s'\vert s,a)$. As the agent interacts with the environment it gathers a trajectory: $(s_0,a_0,r_0,s_1,a_1,r_1,s_2...)$.
We seek an optimal policy $\pi^*\in \argmax_\pi J^\pi$ that maximises the total expected discounted return: $ J^\pi\coloneqq \mathbb{E}_{\pi} \left[\sum_{t=0}^\infty \gamma^t r_t\right]$ 
where $\mathbb{E}_{\pi}$ is the expectation over trajectories induced by $\pi$.
The $Q$-function is the total expected reward as a function of a state-action pair: $Q^\pi(s,a)\coloneqq \mathbb{E}_{\pi_\theta}[\sum_{t=0}^\infty  r_t\vert s_0=s,a_0=a]$. Any $Q$-function satisfies the Bellman equation $\mathcal{B}[Q^\pi]=Q^\pi$ where the Bellman operator is defined as:
\begin{align}
    &\mathcal{B}[Q^\pi](s,a)\coloneqq \mathbb{E}_{P^\pi(s',a\vert s,a)}\left[ r(s',a,s) + \gamma  Q^\pi(s',a')\right]. \label{eq:bellman}
\end{align}
\vspace{-0.5cm}
\subsection{Model-based vs Model-free Bayesian RL}
Bayes-adaptive MDPs (BAMDPs) \citep{Duff02} are a framework for \emph{model-based} Bayesian reinforcement learning where a posterior marginalises over the uncertainty in the unknown transition distribution and reward functions to derive a Bayesian MDP. BAMDP optimal policies are the gold standard, optimally balancing exploration with exploitation but require learning a model of the unknown transition distribution which is typically challenging due to its high-dimensionality and multi-modality \citep{Song13}. Futhermore, planning in BAMDPs requires the calculation of high-dimensional integrals which render the problem intractable. Even with approximation, most existing methods are restricted to small and discrete state-action spaces \citep{Asmuth11,Guez13}. One notable exception is VariBAD  \citep{zintgraf20} which exploits a meta learning setting to carry out approximate Bayesian inference. Unfortunately this approximation sacrifices the BAMDP's theoretical properties and there are no convergence guarantees. 

Existing model-free Bayesian RL approaches attempt to solve a Bayesian regression problem to infer a posterior predictive over a value function \citep{Vlassis2012,Ghavamzadeh15}. Whilst foregoing  the ability to separately model reward uncertainty and transition dynamics, modelling uncertainty in a value function avoids the difficulty of estimating high dimensional conditional distributions and mimics a Bayesian regression problem, for which there are tractable approximate methods \citep{Jordan99,Beal03,kingma2014,rezende15,Gal2016Uncertainty,Lakshminarayanan17}. These methods assume access to a dataset of $N$ samples: $\mathcal{D}^N\coloneqq \{q_i\}_{i=1:N}$ from a distribution over the true $Q$-function at each state-action pair: $q_i\sim P_Q(\cdot\vert s_i,a_i)$. Each sample is an estimate of a point of the true $Q$-function $q_i= Q^\pi(s_i,a_i) + \eta_i $ corrupted by noise $\eta_i$. By introducing a probabilistic model of this random process, the posterior predictive $P(Q^\pi\vert s,a, \mathcal{D}^N)$ can be inferred, which characterises the aleatoric uncertainty in the sample noise and epistemic uncertainty in the model. Modeling aleatoric uncertainty is the goal of distributional RL \citep{bellemare17}. In Bayesian RL we are more concerned with epistemic uncertainty, which can be reduced by exploration \citep{Osband18}.
\vspace{-0.1cm}
\subsection{Theoretical Issues with Existing Approaches}
\label{sec:theoretical_issues}
Unfortunately for most settings it is impractical to sample directly from the true $Q$-function. To obtain efficient algorithms the samples $q_i$ are approximated using bootstrapping: here a parametric function approximator $\Qwdot:\mathcal{S}\times \mathcal{A}\rightarrow \mathbb{R}$ parametrised by $\omega\in\Omega$ is learnt as an approximation of the $Q$-function $\Qwdot\approx Q^\pi$ and then a TD sample is used in place of $q_i$. For example a one-step TD estimate approximates the samples as: $q_i\approx r_i+\gamma \hat{Q}_\omega(s_i,a_i)$, introducing an error that is dependent on $\omega$. Existing approaches do not account for this error's dependency on the function approximator.  Samples are no longer noisy estimates of a point $Q^\pi(s_i,a_i)$ and the resulting posterior predictive is not $P(Q^\pi\vert s,a, \mathcal{D}^N)$ as it has dependence on $\hat{Q}_\omega$ due to the dataset. This is a major theoretical issue that raises the following questions:
\begin{enumerate}
    \item Do model-free Bayesian RL approaches that use bootstrapping still infer a posterior?
    \item If it exists, how does this posterior relate to solving the RL problem? 
    \item What effect does approximate inference have on the solution?
    \item Do methods that sample from an approximate posterior converge?
\end{enumerate}
\paragraph{Contribution:} Our primary contribution is to address these questions by introducing the BBO framework. In answer to Question 1, BBO shows that, by introducing bootstrapping, we actually infer a posterior over Bellman operators. We can use this posterior to marginalise over all Bellman operators to obtain a Bayesian Bellman operator. Our theoretical results provide answers to Questions 2-4, proving that the Bayesian Bellman operator can parametrise a TD fixed point as the number of samples $N\rightarrow\infty$ and is analogous to the projection operator used in convergent reinforcement learning. Our results hold even under posterior approximation. Although our contributions are primarily theoretical, many of the benefits afforded by Bayesian methods play a significant role in a wide range of real-world applications of RL where identifying decisions that are being made under high uncertainty is crucial. We discuss the impact of our work further in \cref{app:broader_impact}.

%% file: Sections/03_Bayesian_Bellman_operators.tex
\vspace{-0.1cm}
\section{Bayesian Bellman Operators}
\label{sec:bbo}
\vspace{-0.1cm}
\emph{Detailed proofs and a discussion of assumptions for all  theoretical results are found in \cref{app:proofs}.}

To introduce the BBO framework we consider the Bellman equation using a function approximator: $\Bqwdot=\Qwdot$. Using \cref{eq:bellman}, we can write the Bellman operator for $\Qwdot$ as an expectation of the empirical Bellman function $b_\omega$:
\begin{align}
    &\Bqw=\mathbb{E}_{P^\pi(s',a'\vert a,s)}\left[  b_\omega(s',a',s,a)\right]\label{eq:bellman_approx},\quad b_\omega(s',a',s,a)\coloneqq r(s',a,s)+\gamma \hat{Q}_\omega(s',a').
\end{align}
When solving the Bellman equation,  the function approximator $\hat{Q}_\omega$ is known but we are uncertain of its value under the Bellman operator due to the reward function and transition distribution. In BBO we capture this uncertainty by treating the empirical Bellman function as a transformation of variables $b_\omega(\cdot,s,a):\mathcal{S}\times\mathcal{A}\rightarrow\mathbb{R}$ for each $(s,a)$. The transformed variable $B:\mathbb{R}\rightarrow\mathbb{R}$ has a conditional distribution $P_B(b\vert s,a,\omega)$ which is the \emph{pushforward} of $P^\pi(s',a',s,a)$ under the transformation $b_\omega(\cdot,s,a)$. For any $P_B$-integrable function $f:\mathbb{R}\rightarrow \mathbb{R}$, the pushforward distribution satisfies: 
\begin{align}
\mathbb{E}_{P_B(b\vert s,a,\omega)}\left[f(b)\right]=\mathbb{E}_{P^\pi(s',a'\vert s,a)}\left[f\circ b_\omega(s',a',s,a)\right]. \label{eq:pushforward}
\end{align}
As the pushforward $P_B(b \vert s,a,\omega)$ is a distribution over empirical Bellman functions, each sample $b\sim P_B(\cdot \vert s,a,\omega)$ is a noisy sample of the Bellman operator at a point: $b_i=\mathcal{B}[\hat{Q}_\omega](s_i,a_i)+\eta_i$. To prove this, observe that taking expectations of $b$ recovers $\Bqw$: 
\begin{align}
    \mathbb{E}_{P_B(b\vert s,a,\omega)}[b]\underbrace{=}_{\textrm{\cref{eq:pushforward}}}\mathbb{E}_{P^\pi(s',a'\vert s,a)}\left[ b_\omega(s',a',s,a)\right]\underbrace{=}_{\textrm{\cref{eq:bellman_approx}}}\Bqw.
\end{align}
As the agent interacts with the environment, it obtains samples from the transition distribution $s'_i\sim P(\cdot\vert s_i,a_i)$ and policy $a_i'\sim \pi(\cdot\vert s_i')$. From \cref{eq:pushforward} a sample from the distribution $b_i\sim P_B(\cdot \vert s_i,a_i,\omega)$ is obtained from these state-action pairs by applying the empirical Bellman function $b_i=  r_i+\gamma \hat{Q}_\omega(s_i',a_i')$. As we discussed in \cref{sec:theoretical_issues}, existing model-free Bayesian RL approaches incorrectly treat each $b_i$ as a sample from a distribution over the value function $P(Q^\pi\vert s,a)$. BBO corrects this by modelling the true conditional distribution: $ P_B(b \vert s,a,\omega)$ that generates the data.\begin{wrapfigure}{r}{0.15\textwidth}
 \centering
   \vspace{-0.35cm}
    \includegraphics[scale=0.61]{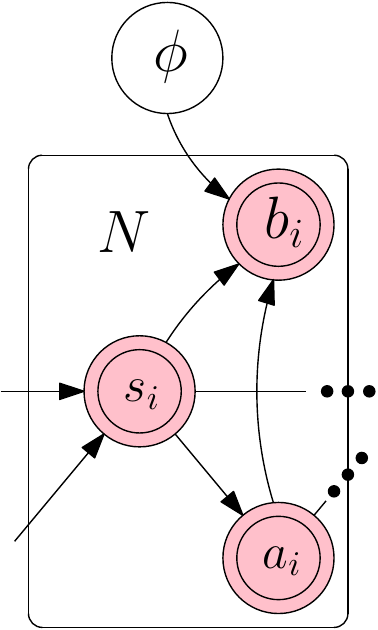}
     \vspace{-0.3cm}
    \caption{Graphical Model for BBO.}
    \label{fig:graphical_bbo}
    \vspace{-1.5cm}
\end{wrapfigure}

The graphical model for BBO is shown in \cref{fig:graphical_bbo}. To model $ P_B(b \vert s,a,\omega)$ we assume a parametric conditional distribution: $P(b\vert s,a,\phi)$ with model parameters $\phi\in\Phi$ and a conditional mean: $\mathbb{E}_{P(b\vert s,a,\phi)}[b]=\hat{B}_\phi(s,a)$. It is also possible to specify a nonparametric model: $P(b\vert s,a)$. The conditional mean of the distribution $\hat{B}_\phi$ defines a function space of approximators that represents a space of Bellman operators, each indexed by $\phi\in\Phi$. The choice of $P(b\vert s,a,\phi)$ should therefore ensure that the space of approximate Bellman operators characterised by $\hat{B}_\phi$ is expressive enough to sufficiently represent the true Bellman operator. As we are not concerned with modelling the transition distribution in our model-free paradigm, we assume states are sampled either from an ergodic Markov chain, or i.i.d.\ from a buffer. Off-policy samples can be corrected using importance sampling.

\begin{assumption}[State Generating Distribution] \label{ass:ergodic} Each state $s_i$ is drawn either i) i.i.d.\ from a distribution $\rho(s)$ with support over $S$ or ii) from an ergodic Markov chain with stationary distribution $\rho(s)$ defined over a $\sigma$-algebra that is countably generated from $S$.
\end{assumption}
 
We represent our preexisting beliefs in the true Bellman operator by specifying a prior $P(\phi)$ with a density $p(\phi)$ which assigns mass over parameterisations of function approximators $\phi\in\Phi$ in accordance with how well we believe they represent $\Bqwdot$. Given the prior and a dataset $\mathcal{D}^N_\omega\coloneqq\{b_i,s_i,a_i\}_{i=1:N} $ of samples from the true distribution $P_B$, we infer the posterior density using Bayes' rule (see \cref{app:posterior_derivation} for a derivation using both state generating distributions of \cref{ass_app:ergodic}):
\begin{align}
     p(\phi\vert \mathcal{D}_\omega^N)=\frac{\prod_{i=1}^N p(b_i\vert s_i,a_i,\phi)p(\phi)}{\int_\Phi \prod_{i=1}^N p(b_i\vert s_i,a_i,\phi)dP(\phi)}.\label{eq:posterior}
 \end{align}
To be able to make predictions, we infer the posterior predictive: $p(b\vert \Dwn, s,a)\coloneqq\int_\Phi p(b\vert s,a,\phi) dp(\phi\vert \Dwn)$. Unlike existing approaches, our posterior density is a function of $\omega$, which correctly accounts for the dependence on $\hat{Q}_\omega$ in our data and the generating distribution $P_B(b\vert s,a,\omega)$. We must therefore introduce a method of learning the correct $Q$-function approximator. As every Bellman operator characterises an MDP, the posterior predictive mean represents a Bayesian estimate of the true MDP by using the posterior to marginalise over all Bellman operators that our model can represent according to our uncertainty in their value:
\begin{align}
\mathcal{B}^\star_{\omega, N}(s,a)\coloneqq &\mathbb{E}_{p(b\vert \mathcal{D}^N_\omega, s,a)}[b] =\mathbb{E}_{p(\phi\vert \mathcal{D}^N_\omega)}\left[ \hat{B}_\phi(s,a)\right].\label{eq:Bayesian_Bellman_Operator}
\end{align}
For this reason, we refer to the predictive mean $\mathcal{B}^\star_{\omega, N}$ as the \emph{Bayesian Bellman operator} and our $Q$-function approximator should satisfy a Bellman equation using $\mathcal{B}^\star_{\omega, N}$. Our objective is therefore to find $\omega^\star$ such that $\hat{Q}_{\omega^\star}=\mathcal{B}^\star_{{\omega^\star}, N}$. A simple approach to learn $\omega^\star$ is to minimise the mean squared Bayesian Bellman error (MSBBE) between the posterior predictive and function approximator:
\begin{align}
\msbbe\coloneqq\left\lVert \Qwdot - \mathcal{B}^\star_{\omega, N}\right\rVert_{\rho,\pi}^2
\label{eq:msbbe}
\end{align}
Here the distribution on the $\ell_2$-norm is $\rho(s)\pi(a\vert s)$ where recall $\rho(s)$ is defined in \cref{ass:ergodic}. Although the MSBBE has a similar form to a mean squared Bellman error with a Bayesian Bellman operator in place of the Bellman operator, our theoretical results in \cref{sec:theory} show its frequentist interpretation is closer to the mean squared projected Bellman operator used by convergent TD algorithms \citep{Sutton09a}. We derive the MSBBE gradient in \cref{app:msbbe_grad}:
\begin{align}
    &\nabla_\omega \textrm{MSBBE}_N(\omega) \\&\quad\quad=\mathbb{E}_{\rho,\pi} \left[\left(\hat{Q}_\omega- \mathbb{E}_{P(\phi\vert \Dwn)} \left[\hat{B}_\phi\right] \right)\left(\nabla_\omega \hat{Q}_\omega-  \mathbb{E}_{P(\phi\vert \Dwn)} \left[\hat{B}_\phi\nabla_\omega \log p(\phi\vert \Dwn)\right] \right)\right].\label{eq:msbbe_grad}
\end{align}
If we can sample from the posterior then unbiased estimates of $\nabla_\omega \textrm{MSBBE}_N(\omega)$ can be obtained, hence minimising the MSBBE via a stochastic gradient descent algorithm is convergent if the standard Robbins-Munro conditions are satisfied \citep{Robbins51}. When existing approaches are used, the posterior has no dependence on $\omega$ and the gradient $\nabla_\omega \log p(\phi\vert \Dwn)$ is not accounted for, leading to gradient terms being dropped in the update. Stochastic gradient descent using these updates does not optimise any objective and so may not converge to any solution. The focus of our analysis in \cref{sec:approximate_inference} is to extend convergent gradient methods for minimising the MSSBE to approximate inference techniques in situations where sampling from the posterior becomes intractable.

Minimising the MSBBE also avoids the double sampling problem encountered in frequentist RL where to minimise the mean squared Bellman error, two independent samples from $P(s'\vert s,a)$ are required to obtain unbiased gradient estimates \citep{Baird95}. In BBO, this issue is avoided by drawing two independent approximate Bellman operators $B_{\phi_1}$ and $B_{\phi_2}$ from the posterior $\phi_1,\phi_2\sim P(\cdot\vert \Dwn)$ instead.

\subsection{Consistency of the Posterior}
\label{sec:theory}
To address Question 2, we develop a set of theoretical results to understand the posterior's relationship to the RL problem. We introduce some mild regularity assumptions on our choice of model:
\begin{assumption}[Regularity of Model]\label{ass:model_regularity}  i)
$\hat{Q}_\omega$ is bounded and $(\Phi,d_\Phi)$ and $(\Omega,d_\Omega)$ are compact metric spaces; ii) $\hat{B}_\phi$ is Lipschitz in $\phi$, $P(b\vert s,a,\phi)$ has finite variance and a density $p(b\vert s,a,\phi)$ which is Lipschitz in $\phi$ and bounded; and iii) $p(\phi)\propto \exp\left(-R(\phi)\right)$ where $R(\phi)$ is bounded and Lipschitz.
\end{assumption}
Our main result is a Bernstein-von-Mises-type theorem \citep{kleijn2012} applied to reinforcement learning. We prove that the posterior asymptotically converges to a Dirac delta distribution centered on the set of parameters that minimise the KL divergence between the true and model distributions:
\begin{align}\vspace{-0.1cm}
    \phi^\star_\omega\coloneqq\argmin_{\phi\in\Phi} \kl{P_B(b, s,a\vert\omega)}{P(b,s,a\vert\phi)}=\argmin_{\phi\in\Phi} \mathbb{E}_{P_B(b, s,a\vert\omega)}\left[ -\log p(b,s,a\vert \phi)\right],\vspace{-0.1cm}
\end{align}
where the expectation is taken with respect to distribution that generates the data: $P_B(b, s,a\vert\omega)=P_B(b\vert s,a,\omega)\pi(a\vert s)\rho(s)$. We make a simplifying assumption that there is a single KL minimising parameter, which eases analysis and exposition of our results. We discuss the more general case where it does not hold in \cref{app:assumptions_theorem_2}.   
\begin{assumption}[Single Minimiser] \vspace{-0.1cm}\label{ass:single_minimiser} The set of minimum KL parameters $\phi^\star_\omega$ exists and is a singleton.
\end{assumption} 
\vspace{-0.1cm}
\begin{theorem}
\label{proof:consistency}
Under Assumptions \ref{ass:ergodic}-\ref{ass:single_minimiser}, in the limit $N\rightarrow\infty$ the posterior concentrates weakly on $\phi^\star$: $i)\ P(\phi\vert \Dwn)\Longrightarrow\delta(\phi=\phi^\star_\omega)$ a.s.; $ii)\  \mathcal{B}^\star_{\omega,N}\xrightarrow{a.s.}\hat{B}_{\phi^\star_\omega}$; and $iii)\ \textrm{MSBBE}_N(\omega)\xrightarrow{a.s.}\lVert \Qwdot - \hat{B}_{\phi^\star_\omega}\rVert_{\rho,\pi}^2$.
\end{theorem}
If our model can sufficiently represent the true conditional distribution then $\kl{P_B(b, s,a\vert\omega)}{P(b,s,a\vert \phi_\omega^\star)}=0\implies P_B(b\vert s,a,\omega)=P(b\vert s,a,\phi_\omega^\star)$. \cref{proof:consistency} proves that the posterior concentrates on $\phi^\star_\omega$ and hence the Bayesian Bellman operator converges to the true Bellman operator: $\hat{B}_{\phi^\star_\omega}(s,a)= \mathbb{E}_{P(b\vert s,a,\phi_\omega^\star)}[b]=\mathbb{E}_{P_B( b\vert s,a,\omega)}[b]=\Bqw$. As every Bellman operator characterises an MDP, any Bayesian RL solution obtained using the BBO posterior such as an optimal policy or value function is consistent with the true RL solution. When the true distribution is not in the model class, $B_{\phi^\star_\omega}$ converges to the closest representation of the true Bellman operator according to the parametrisation that maximises the likelihood $ \mathbb{E}_{P_B(b, s,a\vert\omega)}\left[\log p(b,s,a\vert \phi)\right]$. This is analogous to frequentist convergent TD learning where the function approximator converges to a parametrisation that minimises the projection of the Bellman operator into the model class \citep{Sutton09a,Sutton09b,Maei09}. We now make this relationship precise by considering a Gaussian model. 
\vspace{-0.1cm}
\subsection{Gaussian BBO} \label{sec:Gaussian_bbo} To showcase the power of \cref{proof:consistency} and to provide a direct comparison to existing frequentist approaches, we  consider the nonlinear Gaussian model $P(b\vert s,a,\phi)= \mathcal{N}(\hat{B}_\phi(s,a),\sigma^2)$ that is commonly used for Bayesian regression \citep{Murphy12,Gal2016Uncertainty}. The mean is a nonlinear function approximator that best represents the Bellman operator $B_\phi\approx\Bqwdot$ and the model variance $\sigma^2>0$ represents the aleatoric uncertainty in our samples. Ignoring the $\log$-normalisation constant $c_\textrm{norm}$, the $\log$-posterior is an empirical mean squared error between the empirical Bellman samples and the model mean $\hat{B}_\phi(s_i,a_i)$ with additional regularisation due to the prior (see \cref{app:gaussian_bbo} for a derivation): 
\vspace{-0.1cm}
\begin{align}
    -\log p(\phi \vert \Dwn)=c_\textrm{norm}+\sum_{i=1}^N \frac{(b_i-\hat{B}_\phi(s_i,a_i))^2}{2\sigma^2} +R(\phi), \ \  \phi^\star_\omega\in\argmin_{\phi\in\Phi}\lVert  \hat{B}_\phi-\mathcal{B}[\hat{Q}_\omega]\rVert^2_{\rho,\pi}. \label{eq:gaussian_posterior}
\end{align}
\cref{proof:consistency} proves that in the limit $N\rightarrow \infty$, the effect of the prior diminishes and the Bayesian Bellman operator converges to the parametrisation: $\mathcal{B}^\star_{\omega,N}\xrightarrow{a.s.}\hat{B}_{\phi^\star_\omega}$. As $\phi^\star_\omega$ is the set of parameters that minimise the mean squared error between the true Bellman operator and the approximator, $\hat{B}_{\phi^\star_\omega}$ is a \emph{projection} of the Bellman operator onto the space of functions represented by $\hat{B}_\phi$:
\vspace{-0.1cm}
\begin{align}
    \hat{B}_{\phi^\star_\omega}=\mathcal{P}_{\hat{B}_\phi}\circ\mathcal{B}[\hat{Q}_\omega]\coloneqq\{\hat{B}_\phi: \phi\in \argmin_{\phi\in\Phi}\lVert  \hat{B}_\phi-\mathcal{B}[\hat{Q}_\omega]\rVert^2_{\rho,\pi}\}.\label{eq:projection}
\end{align}
Finally, \cref{proof:consistency} proves that the MSBBE converges to the mean squared projected Bellman error $\textrm{MSBBE}_N(\omega)\xrightarrow{a.s.}\textrm{MSPBE}(\omega)\coloneqq\lVert \Qwdot - \mathcal{P}_{\hat{B}_\phi}\circ\mathcal{B}[\hat{Q}_\omega]\rVert_{\rho,\pi}^2$.
By the definition of the projection operator in \cref{eq:projection}, a solution $\Qwdot=\mathcal{P}_{\hat{B}_\phi}\circ\mathcal{B}[\hat{Q}_\omega]$ is a TD fixed point; hence any asymptotic MSBBE minimiser parametrises a TD fixed point should it exist. To further highlight the relationship between BBO and convergent TD algorithms that minimise the mean squared projected Bellman operator, we explore the linear Gaussian regression model as a case study in \cref{app:linear_bbo}, allowing us to derive a regularised Bayesian TDC/GTD2 algorithm \citep{Sutton09b,Sutton09a}.
\vspace{-0.1cm}

%% file: Sections/04_Approximate_BBO.tex
\section{Approximate BBO}
\vspace{-0.1cm}
\label{sec:approximate_bbo}
We have demonstrated in \cref{eq:msbbe_grad} that if it is tractable to sample from the posterior, a simple convergent stochastic gradient descent algorithm can be used to minimise the MSBBE. We derive the gradient update for the linear Gaussian model as part of our case study in \cref{app:linear_bbo}. Unfortunately, models like linear Gaussians that have analytic posteriors are often too simple to accurately represent the Bellman operator for domains of practical interest in RL. We now extend our analysis to include approximate inference approaches.
\vspace{-0.1cm}
\subsection{Approximate Inference}
\label{sec:approximate_inference}
To allow for more expressive nonlinear function approximators, for which the posterior normalisation is intractable, we introduce a tractable posterior approximation: $q(\phi\vert \Dwn)\approx P(\phi \vert \Dwn)$. In this paper, we use randomised priors (RP) \citep{Osband18} for approximate inference. Randomised priors (PR) inject noise into the maximum a posteriori (MAP) estimate via a noise variable $\epsilon\in\mathcal{E}$ with distribution $P_E(\epsilon)$ where the density $p_E(\epsilon)$ has the same form as the prior. We provide a full exposition of RP for BBO in 
\cref{app:randomised_priors}, including derivations of our objectives. RP in practice uses ensembling: $L$ prior randomisations $\mathcal{E}_L\coloneqq\{\epsilon_l\}_{l=1:L}$ are first drawn from $P_E$. To use RP for BBO, we write the $Q$-function approximator as an ensemble of $L$ parameters $\Omega_L\coloneqq \{\omega_l\}_{l=1:L}$ where $\hat{Q}_\omega=\frac{1}{L}\sum_{l=1}^L \hat{Q}_{\omega_l}$ and require an assumption about the prior and the function spaces used for approximators:
\begin{assumption}[RP Function Spaces]\label{ass:RP_approximators}
     i) $\hat{Q}_{\omega_l}$ and $\hat{B}_{\omega_l}$ share a function space where $\Phi=\Omega\subset \mathbb{R}^n$ is compact, convex with a smooth boundary. ii) $\mathcal{E}\subseteq\mathbb{R}^n$ and $R(\phi-\epsilon)$ is defined for any $\phi\in\Phi, \epsilon\in\mathcal{E}$.
\end{assumption}
For each $l\in\{1:L\}$, a set of solutions to the prior-randomised MAP objective are found: 
\begin{align} \psi_l^\star(\omega_l) \in \argmin_{\phi\in \Phi}\mathcal{L}(\phi;\mathcal{D}_{\omega_l}^N,\epsilon_l)\coloneqq\argmin_{\phi\in \Phi}\frac{1}{N}\left( {R}(\phi-\epsilon_l)-\sum_{i=1}^N\log p(b_i\vert s_i,a_i,\phi)\right).\label{eq:RP_objective}
\end{align}
The RP solution $\psi_l^\star(\omega_l)$ has dependence on $\omega_l$ that mirrors the BBO posterior's dependence on $\omega$. To construct the RP approximate posterior $q(\phi\vert \Dwn)$, we average the set of perturbed MAP estimates over all ensembles: $q(\phi\vert \Dwn)\coloneqq \frac{1}{L}\sum_{l=1}^L \delta (\phi\in \psi_l^\star(\omega_l)) $. To sample from the RP posterior $\phi\sim q(\cdot\vert \Dwn)$, we sample an ensemble uniformly $l\sim \textrm{Unif}(\{1:L\})$ and set $\phi=\psi_l^\star(\omega_l)$. Although BBO is compatible with any approximate inference technique, we justify our choice of RP by proving that it preserves the consistency results developed in \cref{proof:consistency}:
\begin{corollary} \label{proof:consistency_approx} Under Assumptions \ref{ass:ergodic}-\ref{ass:RP_approximators}, results i)-iii) of  \cref{proof:consistency} hold with $P(\phi \vert \Dwn)$ replaced by the RP approximate posterior $q(\phi\vert \Dwn)$ both with or without ensembling.
\end{corollary}
In answer to Question 3), \cref{proof:consistency_approx} shows that the difference between using the RP approximate posterior and the true posterior lies in their characterisation of uncertainty and not their asymptotic behaviour. Existing work shows that RP uncertainty estimates are conservative \citep{Pearce19, Ciosek2020} with strong empirical performance in RL \citep{Osband18,Osband19} for the Gaussian model that we study in this paper. 

The RP approximate posterior $q(\phi\vert \Dwn)$  depends on the ensemble of $Q$-function approximators $\hat{Q}_{\omega_l}$ and like in \cref{sec:bbo} we must learn an ensemble of optimal parametrisations $\omega_l^\star$. We substitute for $q(\phi\vert \Dwn)$ in place of the true posterior in \cref{eq:Bayesian_Bellman_Operator,eq:msbbe} to derive an ensembled RP MSBBE: $\textrm{MSBBE}_\textrm{RP}(\omega_l)\coloneqq \lVert \hat{Q}_{\omega_l}- \hat{B}_{\psi_l^\star(\omega_l)}\rVert_{\rho,\pi}^2$. When a fixed point $\hat{Q}_{\omega_l}=\hat{B}_{\psi_l^\star(\omega_l)}$ exists, minimising $\textrm{MSBBE}_\textrm{RP}(\omega_l)$ is equivalent to finding $\omega_l^\star$ such that $\psi_l^\star(\omega_l^\star)=\omega_l^\star$. To learn $\omega_l^\star$ we can instead minimising the simpler parameter objective $\omega_l^\star\in\argmin_{\omega_l\in\Omega}\mathcal{U}(\omega_l;\psi^\star_l)$:
\begin{align}
\vspace{-0.5cm}
\mathcal{U}(\omega_l;\psi^\star_l)\coloneqq \lVert\omega_l-\psi^\star_l(\omega_l)\rVert_2^2\quad  \textrm{such that}\quad \psi_l^\star(\omega_l) \in \argmin_{\phi\in \Phi} \mathcal{L}(\phi;\mathcal{D}_{\omega_l}^N,\epsilon_l), \label{eq:approx_inference_objective}
\vspace{-1cm}  
\end{align} 
which has the advantage that deterministic gradient updates can be obtained. $\mathcal{U}(\omega_l;\psi^\star_l)$ can still provide an alternative auxilliary objective when a fixed point does not exist as the convergence of algorithms minimising \cref{eq:approx_inference_objective} does not depend on its existence and has the same solution as minimising $\textrm{MSBBE}_\textrm{RP}(\omega_l)$ for sufficiently smooth $B_\phi$. Solving the bi-level optimisation problem in \cref{eq:approx_inference_objective} is NP-hard \citep{Bard91}. To tackle this problem, we introduce an ensemble of parameters $\Psi_L\coloneqq\{\psi_l\}_{1:L}$ to track $\psi_l^\star(\omega_l)$ and propose a two-timescale gradient update for each $l\in\{1:L\}$ on the objectives in \cref{eq:approx_inference_objective} with per-step complexity of $\mathcal{O}(n)$:
\begin{gather}
\psi_l\leftarrow\mathcal{P}_\Omega \left( \psi_l-\alpha_k  \nabla_{\psi_l}\left({R}(\psi_l-\epsilon_l) -\log p(b_i\vert s_i,a_i,\psi_l)\right) \right),\quad \textrm{(fast)} \label{eq:rp_updates_fast}\\
\omega_l\leftarrow\mathcal{P}_\Omega( \omega_l-\beta_k (\omega_l-\psi_l)),\quad \textrm{(slow)\label{eq:rp_updates_slow}}
\end{gather}
where $\alpha_k$ and $\beta_k$ are asymptotically faster and slower stepsizes respectively and $\mathcal{P}_\Omega(\cdot)\coloneqq \argmin_{\omega\in\Omega}\lVert \cdot -\omega\rVert_2^2$ is a projection operator that projects its argument back into $\Omega$ if necessary. From a Bayesian perspective, we are concerned with characterising the uncertainty after a \emph{finite} number of samples $N<\infty$ and hence $(b_i,s_i,a_i)$ should be drawn uniformly from the dataset $\mathcal{D}_{\omega_l}^N$ to form estimates of the summation in \cref{eq:RP_objective}, which becomes intractable with large $N$. When compared to existing RL algorithms, sampling from $\mathcal{D}_{\omega_l}^N$ is analogous to sampling from a replay buffer \citep{mnih2015}. 
 A frequentist analysis of our updates is also possible by considering samples that are drawn online from the underlying data generating distribution $(b_i,s_i,a_i)\sim P_B$ in the limit $N\rightarrow\infty$. We discuss this frequentist interpretation further in \cref{app:frequentist_convergence}.
 
To answer Question 4), we prove convergence of updates~(\ref{eq:rp_updates_fast}) and~(\ref{eq:rp_updates_slow}) using a straightforward application of two-timescale stochastic approximation \citep{Borkar97, Borkar08, Heusel17} to BBO. Intuitively, two timescale analysis proves that the faster timescale update~(\ref{eq:rp_updates_fast}) converges to an element in $\Omega$ using standard martingale arguments, viewing the parameter $\omega_l$ as quasi-static as it behaves like a constant. Since the perturbations are relatively small, the separation of timescales then ensures that $\psi_l$ tracks $\psi^\star_l(\omega_l)$ whenever $\omega_l$ is updated in the slower timescale update~(\ref{eq:rp_updates_slow}), viewing the parameter $\psi_l$ as quasi-equilibrated \citep{Borkar08}. We introduce the standard two-timescale regularity assumptions and derive the limiting ODEs of updates~(\ref{eq:rp_updates_fast}) and~(\ref{eq:rp_updates_slow}) in \cref{app:assumptions_theorem_2}:
\vspace{-0.1cm}
\begin{assumption}[Two-timescale Regularity]\label{ass:two_timescale} i) $\nabla_{\psi_l}{R}(\psi_l-\epsilon_l)$ and $\nabla_{\psi_l}\log p(b_i\vert s_i,a_i,\psi_l)$ are Lipschitz in $\psi_l$ and $(b_i,s_i,a_i)\sim \normalfont{\textrm{Unif}}(\mathcal{D}_{\omega_l}^N)$; ii) $\psi^\circledast (\omega_l)$ and $\omega^\circledast_l$ are local aysmptotically stable attractors of the limiting ODEs of updates~(\ref{eq:rp_updates_fast}) and~(\ref{eq:rp_updates_slow}) respectively and $\psi^\circledast_l(\omega_l)$ is Lipschitz in $\omega_l$; and iii) The stepsizes satisfy: $\lim_{k\rightarrow\infty} \frac{\beta_k}{\alpha_k}=0,\ 
    \sum_{k=1}^\infty \alpha_k =\sum_{k=1}^\infty \beta_k=\infty,\ \sum_{k=1}^\infty\left( \alpha_k^2+ \beta_k^2\right)<\infty$.
\end{assumption} 
\begin{theorem}
\vspace{-0.1cm}
\label{proof:approx_convergence} If Assumptions~\ref{ass:ergodic} to~\ref{ass:two_timescale} hold, $\psi_l$ and $\omega_l$ converge to $\psi^\circledast_l(\omega^\circledast_l)$ and $\omega^\circledast_l$ almost surely.
\end{theorem}
As $\omega_l$ are updated on a slower timescale, they lag the parameters $\psi_l$. When deriving a Bayesian actor-critic algorithm in \cref{sec:bbac}, we demonstrate that these parameters share a similar role to a \emph{lagged critic}. There is no Bayesian explanation for these parameters under existing approaches: when applying approximate inference to $P(Q^\pi\vert s,a,\mathcal{D}^N) $, the RP solution $\psi_l^\star$ has no dependence on $\omega_l$. Hence, minimising  $\mathcal{U}(\omega_l;\psi^\star_l)$ and the approximate MSBBE has an exact solution by setting $\omega_l^\star=\psi_l^\star$. In this case, $\hat{Q}_{\omega_l^\star}=\hat{B}_{\psi_l^\star}$ meaning that existing approaches do not distinguish between the $Q$-function and Bellman operator approximators.
\subsection{Bayesian Bellman Actor-Critic}\label{sec:bbac}
\setlength{\columnsep}{6pt}
\begin{wrapfigure}{r}{0.44\textwidth}
 \centering
   \vspace{-1.4cm}
    \includegraphics[scale=0.55]{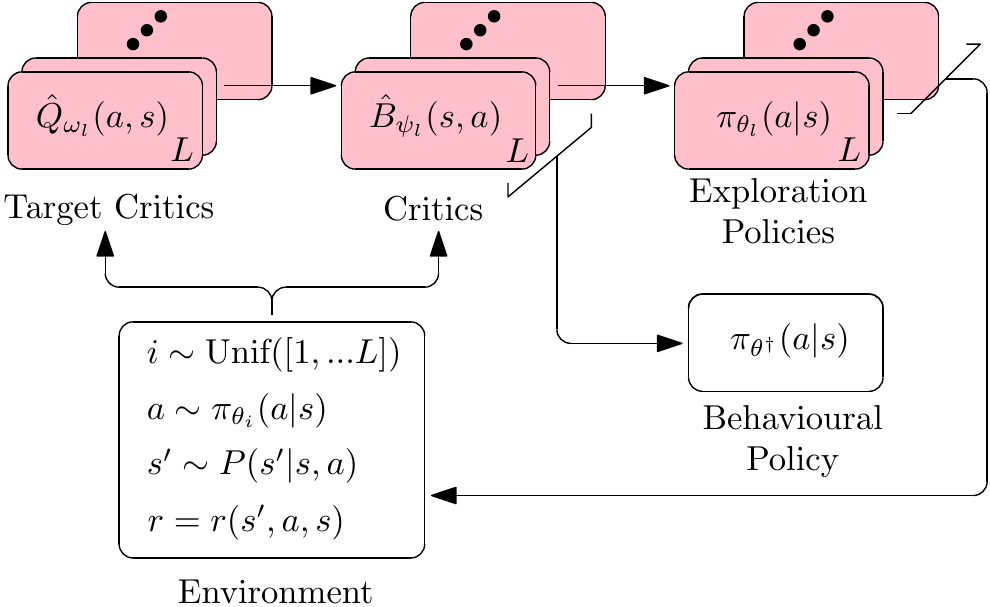}
     \vspace{-0.1cm}
    \caption{Schematic of RP-BBAC.}
    \label{fig:bbac}
    \vspace{-0.5cm}
\end{wrapfigure} 
BootDQN+Prior \citep{Osband18,Osband19} is a state-of-the-art Bayesian model-free algorithm with Thompson sampling \citep{Thompson33} where, in principle, an optimal $Q$-function is drawn from a posterior over optimal $Q$-functions at the start of each episode. As BootDQN+Prior requires bootstrapping, it actually draws a sample from the Gaussian BBO posterior introduced in \cref{sec:Gaussian_bbo} using RP approximate inference with the empirical Bellman function $b_\omega(s',a,s)=r(s',a,s)+\gamma\max_{a'} \hat{Q}_\omega(s',a')$. This empirical Bellman function results from substituting an optimal policy $\pi(a\vert s) =\delta(a\in\argmax_{a'} \hat{Q}_{\omega}(s,a'))$ in \cref{eq:pushforward}. A variable $l$ is drawn uniformly and the optimal exploration policy $\pi_l^\star(a\vert s)=\delta(a\in\argmax_{a'} B_{\phi_l}(s,a')) $ is followed. BootDQN+Prior achieves what \citet{Osband19} call \emph{deep exploration} where exploration not only considers immediate information gain but also the consequences  of an  exploratory action  on  future  learning. Due its use of the $\argmax$ operator,  BootDQN+Prior is not appropriate for continuous action or large discrete action domains as a nonlinear optimisation problem must be solved every time an action is sampled. We instead develop a randomised priors Bayesian Bellman actor-critic (RP-BBAC) to extend BootDQN+Prior to continuous domains. A schematic of RP-BBAC is shown in \cref{fig:bbac} which summarises \cref{alg:BBAC}. Additional details are in 
\cref{app:bbac_algorithm}.  

\paragraph{Comparison to existing actor-critics:} Using a Gaussian model also allows a direct comparison to frequentist actor-critic algorithms \citep{Konda99}: as shown in \cref{fig:bbac}, every ensemble $l\in\{1...L\}$ has its own \emph{exploratory  actor} $\pi_{\theta_l}$, \emph{critic} $B_{\psi_l}$ and \emph{target critic} $\hat{Q}_{\omega_l}$. In BBAC, each critic is the solution to its unique $\epsilon_l$-randomised empirical MSBBE objective from \cref{eq:approx_inference_objective}: $\mathcal{L}_\textrm{critic}(\psi_l)\coloneqq -\frac{1}{\sigma^2}\sum_{i=1}^N (b_i-\hat{B}_{\psi_l}(s_i,a_i))^2+{R}(\psi_l-\epsilon_l)$. The target critic parameters $\omega_l$ for each Bellman sample $b_i=r_i+\gamma\hat{Q}_{\omega_l}(s'_i,a_i')$ are updated on a slower timescale to the critic parameters, which mimics the updating of target critic parameters after a regular interval in frequentist approaches \citep{mnih2015,Haarnoja18}. We introduce an ensemble of parametric exploration policies $\pi_{\theta_l}(a\vert s)$ parametrised by a set of parameters $\Theta_L\coloneqq \{\theta_l\}_{l=1:L}$. Each optimal exploration policy $\pi_l^\star(a\vert s)$ is parametrised by the solution to its own optimisation problem: $\theta_l^\star\in\argmax_{\theta_l\in\Theta}\mathbb{E}_{\rho(s)\pi_{\theta_l}(a\vert s)}[B_{\phi_l}(s,a')]$. Unlike frequentist approaches, an exploratory actor is selected at the start of each episode in accordance with our current uncertainty in the MDP characterised by the approximate RP posterior.\begin{wrapfigure}{r}{0.45\textwidth}
 \centering
   \vspace{-0.5cm}
    \begin{minipage}{0.45\textwidth}
\begin{algorithm}[H]
    \caption{\textsc{RP-BBAC}}
    \label{alg:BBAC}
    \begin{algorithmic}
	    \STATE Initialise $\Theta_L, \Omega_L, \Psi_L,\mathcal{E}_L,\theta^\dagger$ and $\mathcal{D}\leftarrow \varnothing$
	    \STATE Sample initial state $s \sim P_0$
	    \WHILE{\textbf{not} converged}
	    \STATE Sample policy $\theta_l\sim \textrm{Unif}(\Theta_L)$
    	\FOR{$n\in\{1,...N_\textrm{env}\}$}
    	\STATE Sample action $a \sim \pi_{\theta_l}(\cdot \vert s)$
    	\STATE Observe next state $s' \sim P(\cdot \vert s,a)$
    	\STATE Observe reward $r=r(s',a,s)$
    	\STATE $\mathcal{D}\leftarrow \mathcal{D} \cup \{s,a,r,s'\}$
    	\ENDFOR
    	\STATE $\Theta_L, \Omega_L, \Psi_L\leftarrow \textsc{UpdatePosterior}$
    	\STATE $\theta^\dagger \leftarrow \textsc{UpdateBehaviouralPolicy}$
    	\ENDWHILE
    \end{algorithmic}
\end{algorithm}
   \vspace{-1cm}
\end{minipage}
\end{wrapfigure} Exploration is thus both deep and adaptive as actions from an exploration policy are directed towards minimising epistemic uncertainty in the MDP and the posterior variance reduces in accordance with \cref{proof:consistency_approx} as more data is collected. BBAC's explicit specification of lagged target critics is unique to BBO and, as discussed in \cref{sec:approximate_inference}, corrects the theoretical issues raised by applying bootstrapping to existing model-free Bayesian RL theory, which does not account for the posterior's dependence on $\hat{Q}_\omega$. Finally, exploration policies may not perform well at test time, so we learn a behaviour policy $\pi_{\theta^\dagger}(a\vert s)$ parametrised by $\theta^\dagger\in\Theta$ from the data collected by our exploration policies using the ensemble of critics: $\{\hat{B}_{\psi_l}\}_{l=1:L}$. Theoretically, this is the optimal policy for the Bayesian estimate of the true MDP by using the approximate posterior to marginalise over the ensemble of Bellman operators. We augment our behaviour policy objective with entropy regularisation, allowing us to combine the exploratory benefits of Thompson sampling with the faster convergence rates and algorithmic stability of regularised RL \citep{Vieillard2020}. 
\vspace{-0.2cm}

%% file: Sections/05_related_work.tex
 \section{Related Work}
 \vspace{-0.1cm}
Existing model-free Bayesian RL approaches assume either a parametric Gaussian \citep{Gal16a,Osband18,Fortunato18,Lipton18,Osband19,Touati19} or Gaussian process regression model \citep{Engel03,Engel05}. Value-based approaches use the empirical Bellman function $b_\omega(s',a,s) = r(s',a,s)+\gamma\max_{a'}\hat{Q}_\omega(s',a') $ whereas actor-critic approaches use the empirical Bellman function $b_\omega(s',a',s,a) = r(s',a,s)+\gamma\hat{Q}_\omega(s',a') $. In answering Questions 1-4, we have shown existing methods that use bootstrapping inadvertently approximate the posterior predictive over $Q$-functions with the BBO posterior predictive $P(Q^\pi\vert s,a,\mathcal{D}^N)\approx P(b\vert s,a,\mathcal{D}^N_\omega)$. These methods minimise an approximation of the MSBBE where the Bayesian Bellman operator is treated as a supervised target, ignoring its dependence on $\omega$: gradient descent approaches drop gradient terms and fitted approaches iteratively regress the $Q$-function approximator onto the Bayesian Bellman operator $\hat{Q}_{\omega_{k+1}}\leftarrow\mathcal{B}^\star_{\omega_k,N}$. In both cases, the updates may not be a contraction mapping for the same reasons as in non-Bayesian TD \citep{Tsitsiklis97} and so it is not possible to prove general convergence. The additional Bayesian regularisation introduced from the prior can lead to convergence, but only in specific and restrictive cases \citep{Antos07,Antos08,Feng19,Brandfonbrener2019}.

Approximate inference presents an additional problem for existing approaches: many existing methods  na\"ively apply approximate inference to the Bellman error, treating $\mathcal{B}[Q^\pi](s,a)$ and $Q^\pi(s,a)$ as independent variables \citep{Fortunato18,Lipton18,Touati19,Gal16a}. This leads to poor uncertainty estimates as the Bellman error cannot correctly propagate the uncertainty \citep{odonoghue18a,Osband18}. \citet{Osband19} demonstrate that this can cause uncertainty estimates of $Q^\pi(s,a)$ at some $(s,a)$ to be zero and propose BootDQN+Prior as an alternative to achieve deep exploration. BBO does not suffer this issue as the posterior characterises the uncertainty in the Bellman operator directly. In \cref{sec:bbac} we demonstrated that BootDQN+Prior derived from BBO specifies the use of target critics. Despite being essential to performance, there is no Bayesian explanation for target critics under existing model-free Bayesian RL theory, which posits that sampling a critic from $P(Q^\pi\vert s,a,\mathcal{D}^N)$ is sufficient.

\vspace{-0.2cm}

%% file: Sections/06_experiments.tex
\section{Experiments}\vspace{-0.1cm}\begin{wrapfigure}{r}{0.32\textwidth}
\vspace{-1.45cm}
    \centering
    \includegraphics[trim={2mm 2mm 2mm 2mm}, clip, scale=0.7]{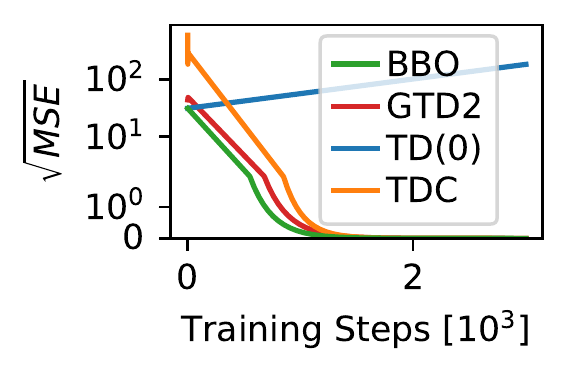}
    \vspace{-2mm}
    \caption{Tsitsiklis counterexample.}
    \label{fig:spiral_result}
    \vspace{-0.4cm}
\end{wrapfigure}
\label{sec:experiments}
\paragraph{Convergent Nonlinear Policy Evaluation}
To confirm our convergence and consistency results under approximation, we evaluate BBO in several nonlinear policy evaluation experiments that are constructed to present a convergence challenge for TD algorithms. We verify the convergence of nonlinear Gaussian BBO in the famous counterexample task of \citet{Tsitsiklis97}, in which the TD(0) algorithm is provably divergent. The results are presented in~\cref{fig:spiral_result}. As expected, TD(0) diverges, while BBO converges to the optimal solution  faster than convergent frequentist nonlinear TDC and GTD2~\cite{Maei09}. We also consider three additional policy evaluation tasks commonly used to test convergence of nonlinear TD using neural network function approximators: 20-Link Pendulum~\citep{dann2014policy}, Puddle World~\citep{Boyan95}, and Mountain Car~\citep{Boyan95}. Results are shown in \cref{fig:non-linear-policy-evaluation-ablation-result-full} of \cref{app:experimental_details_neural_network_function_approximators} from which we conclude that i) by ignoring the posterior's dependence on $\omega$, existing model-free Bayesian approaches are less stable and perform poorly in comparison to the gradient based MSBBE minimisation approach in \cref{eq:msbbe_grad}, ii) regularisation from a prior can improve performance of policy evaluation by aiding the optimisation landscape \citep{Du17}, and iii) better solutions in terms of mean squared error can be found using BBO instead of the local linearisation approach of nonlinear TDC/GTD2\citep{Maei09}.

\paragraph{Exploration for Continuous Control}\setlength{\columnsep}{-3pt}
\begin{wrapfigure}{r}{0.55\textwidth}
 \centering
 \vspace{-0.6cm}
 \begin{subfigure}{0.27\textwidth}
  \centering
    \includegraphics[width=0.93\linewidth, trim={0mm 0mm 0mm 0mm}, clip]{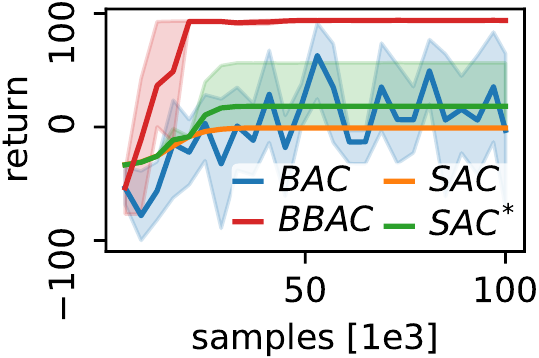}
         \vspace{-1mm}
        \caption{MountainCar}
        \label{fig:bbac-results:mountain-car-v0}
\end{subfigure}
 \begin{subfigure}{0.27\textwidth}
  \centering
         \includegraphics[width=0.93\linewidth, trim={0mm 0mm 0mm 0mm}, clip]{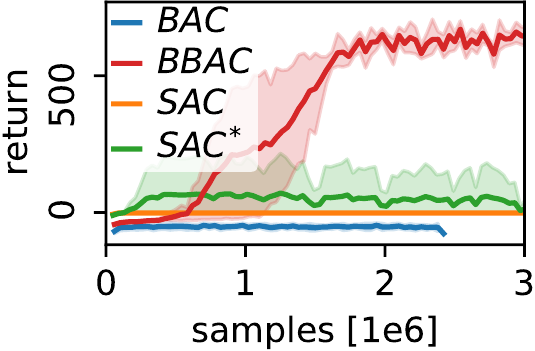}
         \vspace{-1mm}
        \caption{Cartpole}
        \label{fig:bbac-results:cartpole}
\end{subfigure}
 \hspace{-0.8cm}
 \vspace{-0.2cm}
\caption{Continuous control with sparse reward.}
    \label{fig:bbac-results}
\vspace{-0.5cm}
\end{wrapfigure}
In many benchmark tasks for continuous RL, such as the locomotion tasks from MuJoCo Gym suite \citep{brockman2016openai}, the environment reward is shaped to provide a smooth gradient towards a successful task completion and na{\"i}ve Boltzmann dithering exploration strategies from regularised RL can provide a strong inductive bias. In practical real-world scenarios, dense rewards are difficult to specify by hand, especially when the task is learned from raw observations like images. Therefore,  we consider a set of continuous control tasks with sparse rewards as continuous analogues of the discrete experiments used to test BootDQN+Prior \citep{Osband18}: \mbox{\emph{MountainCar-Continuous-v0}} from Gym benchmark suite and a slightly modified version of the \mbox{\emph{cartpole-swingup\_sparse}} from DeepMind Control Suite \citep{tassa2018deepmind}. Both environments have a sparse reward signal and penalize the agent proportional to the magnitude of executed actions. As the agent is always initialised  in the same state, it has to deeply explore costly states in a directed manner for hundreds of steps until it reaches the rewarding region of the state space. We compare RP-BBAC with two variants of the state-of-the-art soft actor-critic: SAC, which is the exact algorithm presented in \citep{Haarnoja18c}; and SAC*, a tailored version which uses a single $Q$-function to avoid \emph{pessimistic underexploration} \citep{ciosek2019better} due to the use of the double-minimum-Q trick (see \cref{app:continuous-control-experiments} for details). To understand the practical implications of our theoretical results, we also compare against BAC which is a variant of RP-BBAC where $\hat{Q}_{\omega_l^\star}=\hat{B}_{\psi_l^\star}$. As we discussed in \cref{sec:approximate_inference}, BAC is the Bayesian actor-critic that results from applying RP approximate inference to the posterior over $Q$-functions used by existing model-free Bayesian approaches with bootstrapping.

\begin{wrapfigure}{r}{0.39\textwidth}
\vspace{-0.4cm}
        \includegraphics[scale=0.27,trim={4mm 4mm 3mm 1mm}]{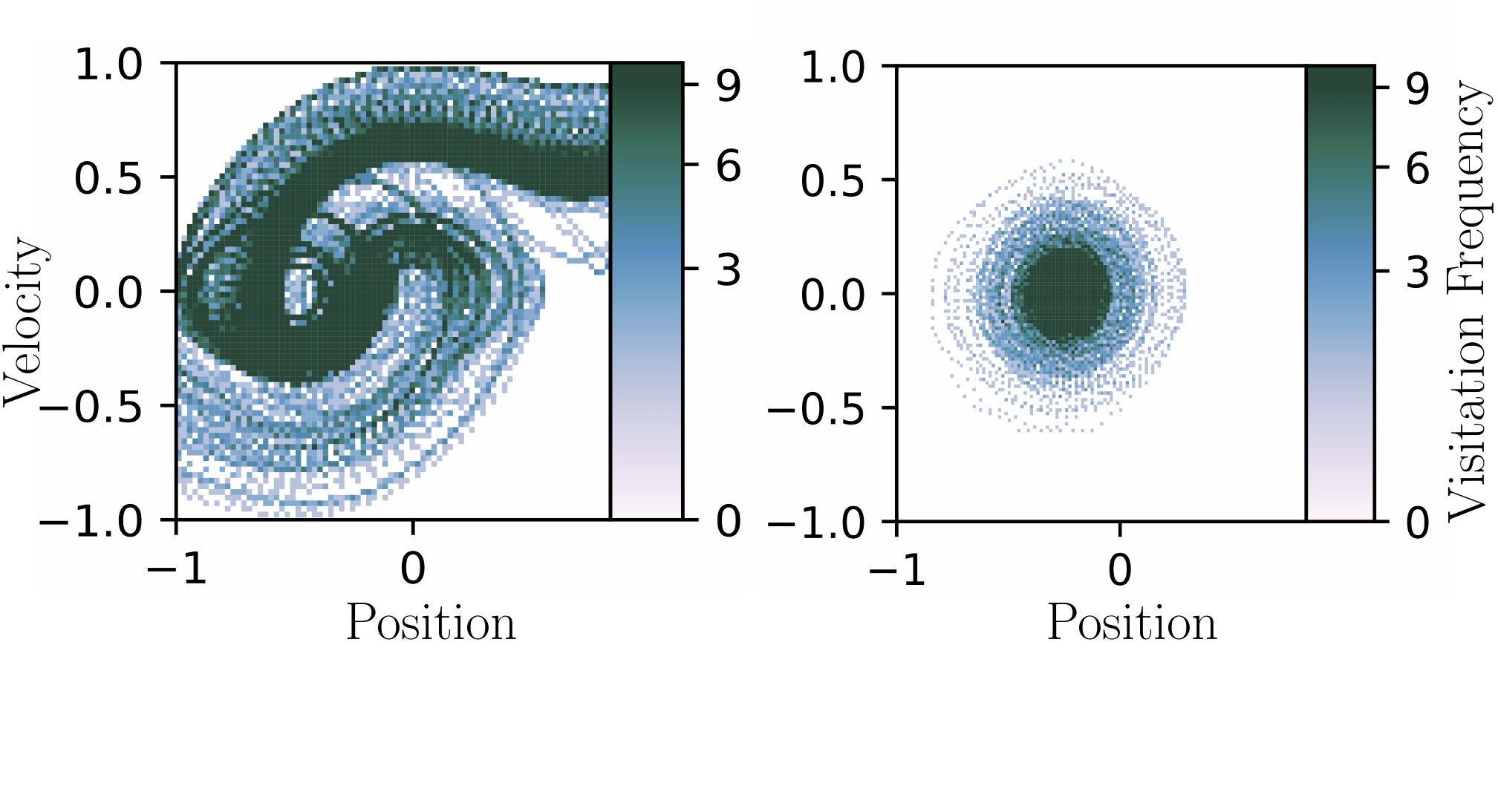}
     \vspace{-0.6cm}
    \caption{State Support for RP-BBAC (left) and SAC  (right) in \mbox{MountainCar-Continuous-v0}.}
    \label{fig:state_support}
    \vspace{-1cm}
\end{wrapfigure}

The results are shown in \cref{fig:bbac-results}. Due to the lack of smooth signal towards the task completion, SAC consistently fails to solve the tasks and converges to always executing the 0-action due to the action cost term, while SAC* achieves the goal in one out of five seeds. RP-BBAC succeeds for all five seeds in both tasks. To understand why, we provide a state support analysis in for \mbox{\emph{MountainCar-Continuous-v0}}  \cref{app:bbac_mountain_car_continuous_v0}. The final plots are shown in \cref{fig:state_support} and confirm that the deep, adaptive exploration carried out by RP-BBAC leads agents to systematically explore regions of the state-action space with high uncertainty. The same analysis for SAC and SAC* confirms the inefficiency of the exploration typical of RL as inference: the agent repeatedly explores actions that lead to poor performance and rarely explores beyond its initial state.  The state support analysis for BAC in \cref{app:bbac_mountain_car_continuous_v0} confirms that by using the posterior over $Q$-functions with bootstrapping, existing model-free Bayesian RL cannot accurately capture the uncertainty in the MDP. Initially, exploration is similar to RP-BBAC but epistemic uncertainty estimates are unstable and cannot concentrate due to the convergence issues highlighted in this paper, preventing adaptive exploration. \setlength{\columnsep}{10pt}
 Our results in \cref{fig:bbac-results} demonstrate that the theoretical issues with existing approaches have negative empirical consequences, verifying that it is essential for Bayesian model-free RL algorithms with bootstrapping sample from the BBO posterior as BAC fails to solve both tasks where sampling from the correct posterior in RP-BBAC succeeds.  In \cref{app:bbac_cartpole}, we also investigate RP-BBAC's sensitivity to randomized prior hyperparameters. The range of working hyperparameters is wide and easy to tune. 
\vspace{-0.2cm}

%% file: Sections/07_conclusion.tex
\section{Conclusion}
\vspace{-0.2cm}
By introducing the BBO framework, we have addressed a major theoretical issue with model-free Bayesian RL by analysing the posterior that is inferred when bootsrapping is used. Our theoretical results proved consistency with frequentist RL and strong convergence properties, even under posterior approximation. We used BBO to extend BootDQN+Prior to continuous domains. Our experiments in environments where rewards are not hand-crafted to aid exploration demonstrate that sampling from the BBO posterior characterises uncertainty correctly and algorithms derived from BBO can succeed where state-of-the-art algorithms fail catastrophically due to their lack of deep exploration. 
\section*{Acknowledgements}
This project has received funding from the European Research Council (ERC) under the European Unions Horizon
2020 research and innovation programme (grant agreement
number 637713). Matthew Fellows and Kristian Hartikainen are funded by the EPSRC. The experiments were made possible by
a generous equipment grant from NVIDIA. We would like to thank Piotr Mi\l{o}\'{s}, whose proof for a similar problem inspired our proof of \cref{proof:posterior_convergence}. 

%% file: Appendix/app_broader_impact.tex
\newpage
\section{Broader Impact}
\label{app:broader_impact}

 Many of the benefits afforded by Bayesian RL methods play a significant role in a wide range of real-world applications, for example, financial applications of deep RL \citep{Deng17} have the potential to destabilise economies, and precise uncertainty quantification can enable safer applications of RL for trading \citep{Jiang17,Spooner18}. Incorporating prior knowledge can have substantial impact on speed and accuracy, for example in medical applications of RL \citep{yu2019reinforcement}. Furthermore, safe exploration and convergence guarantees are crucial in many physical-world domains such as robotics or autonomous vehicles \citep{kiran20}, where undesirable actions and algorithmic divergence could cause direct human harm or even fatalities.

Stronger theoretical guarantees increases confidence in algorithms and can accelerate the implementation of RL for real-world applications. The societal effects of rapidly increased automation are uncertain and disputed \citep{Smith17}. Positive effects include claims of increased job creation, productivity \citep{Acemoglu18} and even techno-utopianism \citep{Srnicek15}. A critical approach to these claims is important, especially given the gulf between the ideals and the reality of what automation has achieved so far \citep{Greenfield18} and the potential automation has for increased inequality \citep{Acemoglu20} and discrimination \citep{Mehrabi19}.

%% file: Appendix/app_proofs.tex
\section{Proofs}
\label{app:proofs}
\setcounter{theorem}{0}
\setcounter{assumption}{0}

\subsection{Assumptions and Preliminaries for \cref{proof:consistency}}
\label{app:assumptions_theorem_1}

\begin{assumption}[State Generating Distribution] \label{ass_app:ergodic} Each state $s_i$ is drawn either i) i.i.d.\ from a distribution $\rho(s)$ with support over $S$ or ii) from an ergodic Markov chain with stationary distribution $\rho(s)$ defined over a $\sigma$-algebra that is countably generated from $S$.
\end{assumption}
Observe that the ergodic Markov chain in \cref{ass_app:ergodic} does not have to be that followed by $\pi(a\vert s)$ and hence our algorithms can also be off-policy as long as the underlying Markov chain is ergodic with stationary distribution: $\rho(s)$. In such a case, the expectation under the evaluation policy $\pi(a\vert s)$ used in the empirical Bellman function may be estimated using importance sampling.

\begin{assumption}[Regularity of Model]\label{ass_app:model_regularity}  i)
$\hat{Q}_\omega$ is bounded and $(\Phi,d_\Phi)$ and $(\Omega,d_\Omega)$ are compact metric spaces; ii) $\hat{B}_\phi$ is Lipschitz in $\phi$, $P(b\vert s,a,\phi)$ has finite variance and a density $p(b\vert s,a,\phi)$ which is Lipschitz in $\phi$ and bounded; and iii) $p(\phi)\propto \exp\left(-R(\phi)\right)$ where $R(\phi)$ is bounded and Lipschitz.
\end{assumption}
 Any $Q$-function is upper and lower bounded as: 
\begin{align}
    \frac{r_{\min} }{1-\gamma } \le Q^\pi \le \frac{r_{\max}}{1-\gamma},
\end{align}
 which can be used as a natural bound when designing the $Q$-function approximator to ensure  \cref{ass_app:model_regularity} i) is satisfied. As the $Q$-function approximator and reward function are bounded, it follows from \cref{eq:bellman_approx} that each $b\sim P_B$ is bounded too, hence $P_B(b\vert s,a,\omega)$ has finite variance. The assumption that the model $P(b\vert s,a,\phi)$ has finite variance under \cref{ass_app:model_regularity} ii) therefore does not affect its capacity to represent $P_B(b\vert s,a,\omega)$. Finally, the prior being bounded under \cref{ass_app:model_regularity} iii) avoids pathological cases where the prior places all its mass on a finite number of parametrisations.
 
\begin{assumption}[Single Minimiser] \label{ass_app:single_minimiser} The set of minimum KL parameters $\phi^\star_\omega$ exists and is a singleton.
\end{assumption} 
 
\cref{ass_app:single_minimiser} is used to simplify analysis and exposition, allowing us to prove convergence to a single Dirac-delta measure in \cref{proof:posterior_convergence}. In the more realistic situation where the KL divergence may have multiple, disjoint minimisers, our analysis holds, however the posterior converges to a weighted sum of Dirac-delta measures centred on each element of the set of minimisers specific to the exact MDP being studied.

Central to our proofs is the empirical regularised log-likelihood (ERLL):
\begin{align}
  \normalfont{\textrm{ERLL}}_N(\phi,\Dwn)\coloneqq\frac{1}{N} \sum_{i=1}^N \log p(b_i\vert s_i,a_i,\phi) - \frac{\sigma^2}{N} R(\phi).\label{empirical_msbe}
\end{align}
Using this notation, we can write the posterior density as:
\begin{align}
    p(\phi\vert \Dwn)=\frac{\exp\left( \frac{N}{\sigma^2}\normalfont{\textrm{ERLL}}_N(\phi,\Dwn)\right)}{\int_\Phi\exp\left( \frac{N}{\sigma^2}\normalfont{\textrm{ERLL}}_N(\phi,\Dwn)\right)d\lambda(\phi)} \label{eq:posterior_alternative_form}
\end{align}
where $\lambda$ is the Lebesgue measure. Our proofs require three separate notions of convergence which we now make precise. 
 \paragraph{$\mathbf{P_\mathcal{D}}$-almost sure convergence:} We denote the distribution of the complete data $\mathcal{D}_\omega\coloneqq \{b_0,s_0,a_0,b_1,s_1,a_1...\}$ as $P_\mathcal{D}$. $\mathcal{D}_\omega^N\subset \mathcal{D}_\omega$ for any finite $N$. A sequence of random variables $Z_N:\mathcal{D}_\omega\rightarrow \mathbb{R} $ converges $P_\mathcal{D}$-almost surely to $Z$ if  $P_\mathcal{D}(\mathcal{D}:\lim_{N\rightarrow \infty} Z_N(\mathcal{D})=Z(\mathcal{D}))=1$ for all $\mathcal{D}\in\mathcal{D}_\omega$. We denote $P_\mathcal{D}$-almost sure convergence of $Z_N$ as $Z_N\xrightarrow[]{P_\mathcal{D}-a.s.} Z$. 
 
 \paragraph{Weak $\mathbf{P_\mathcal{D}}$-almost sure convergence:} Our theorems analyse the behaviour of the posterior $P(\phi\vert \Dwn)$ which depends on the data. We must therefore extend the usual notion of weak convergence to weak $P_\mathcal{D}$-almost sure convergence to account for this dependence by characterising the $P_\mathcal{D}$-almost sure convergence of the random variable $\int_\Phi f(\phi) dP(\phi\vert \mathcal{D}_\omega^N)$:
 \begin{definition}[Weak $P_\mathcal{D}$-almost sure convergence:] A distribution $P_N(\phi\vert \mathcal{D})$ converges weakly to $P^\star(\phi)$ $P_\mathcal{D}$-almost surely if for any continuous, bounded $f:\Phi\rightarrow\mathbb{R}$:
 \begin{align}
     P_\mathcal{D}\left(\mathcal{D}:\lim_{N\rightarrow \infty} \int_\Phi f(\phi) dP_N(\phi\vert \mathcal{D})= \int_\Phi f(\phi) dP^\star(\phi)\right)=1,
 \end{align}
 for all $\mathcal{D}\in\mathcal{D}_\omega $.
 \end{definition}
 We denote weak $P_\mathcal{D}$-almost sure convergence of $P_N(\phi\vert \mathcal{D})$ as $P_N(\phi\vert \mathcal{D})\xLongrightarrow[]{P_\mathcal{D}-a.s.} P^\star(\phi)$
 
 \paragraph{Uniform and Uniform $\mathbf{P_\mathcal{D}}$-almost sure convergence} Informally, uniform convergence strengthens the notion of pointwise convergence, ensuring that for every scalar $\varepsilon>0$, a sequence of functions remain uniformly bounded within a margin $\varepsilon$ of their limiting value after a fixed number in the sequence. Uniform convergence is formally defined as:
\begin{definition}[Uniform Convergence]\label{def:uniform}
Let $f_N :X\rightarrow \mathbb{R}$ be a sequence of real-valued functions. The sequence $(f_N)$ converges uniformly to $f :X\rightarrow \mathbb{R}$ on $X$ if for every $\varepsilon>0$ there exists some natural number $K$ such that for all $x\in X$ and $N\ge K$,
\begin{align}
    \lvert f_N(x)-f(x) \rvert < \varepsilon. \label{eq:uniform_bound_def}
\end{align}
An equivalent definition of uniform converge is:
\begin{align}
   \lim_{N\rightarrow\infty} \sup_{x\in X} \lvert f_N(x)-f(x) \rvert = 0. \label{eq:uniform_sup_def}
\end{align}
\end{definition}
We will denote uniform convergence of $f_N(x)$ as $f_N\xlongrightarrow[]{\textrm{unif}} f$ .

As we will be proving uniform convergence of the ERLL, which is a sequence of random variables, we must extend the notion of uniform convergence to uniform $P_\mathcal{D}$-almost sure convergence by replacing the pointwise convergence condition used to define almost sure convergence with the uniform convergence condition:

\begin{definition}[Uniform $P_\mathcal{D}$-almost sure convergence]\label{def:uniform_as}
Let $Z_N :\mathcal{D}_\omega\rightarrow \mathbb{R}$ be a sequence of random variables. The sequence $(Z_N) $ converges uniformly to $Z :\mathcal{D}_\omega\rightarrow \mathbb{R}$ on $P_\mathcal{D}$-almost surely if for every $\varepsilon>0$ there exists some natural number $K$ such that for all $\mathcal{D}\in \mathcal{D}_\omega$ and $N\ge K$,
\begin{align}
    \lvert Z_N(\mathcal{D})-Z(\mathcal{D}) \rvert < \varepsilon, \label{eq:uniform_bound_def}
\end{align}
except possibly a some subset $\mathcal{D}'\subset \mathcal{D}_\omega$ such that $P_\mathcal{D}(\mathcal{D}')=0$. An equivalent definition of uniform $P_\mathcal{D}$-almost sure converge is:
\begin{align}
   P_\mathcal{D}\left(\mathcal{D}': \lim_{N\rightarrow\infty} \sup_{\mathcal{D}\in \mathcal{D}'} \lvert Z_N(\mathcal{D})-Z(\mathcal{D}) \rvert=0\right) = 1. \label{eq:uniform_sup_def}
\end{align}
\end{definition}
We denote uniform $P_\mathcal{D}$-almost sure convergence of the sequence $(f_N) $ as $f_N\xlongrightarrow[]{\textrm{unif}-P_\mathcal{D}} f$ .
We now start with a proposition that establishes a few useful facts about functions that depend on $\phi$ from our assumptions.

\begin{proposition}[Useful Facts About Functions of $\phi$]\label{proof:useful_facts}
Under \cref{ass:model_regularity}, $i)\ \hat{B}_\phi$ is bounded; and $ii)\ \mathbb{E}_{P_B(b, s,a,\vert\omega)}\left[ \vert\log p(b\vert s,a,\phi) \rvert \right]<\infty\ \forall \ \phi\in\Phi$.
\begin{proof}
To prove $i)$, we recall that $\hat{B}_\phi$ is defined as the model's mean: $\hat{B}_\phi(s,a)\coloneqq \mathbb{E}_{P(b\vert s,a,\phi)}[b]$.  If $ \hat{B}_\phi(s,a)$ is not bounded for all $(s,a)$, the variance of $P(b\vert s,a,\phi)$ would also not be bounded, which would lead to a contradiction as $P(b\vert s,a,\phi)$ has finite variance under \cref{ass:model_regularity}. To prove $ii)$, we note that as $p(b\vert s,a,\phi)$ is bounded under \cref{ass:model_regularity}, it must be $P_B$-integrable and hence $\mathbb{E}_{P_B(b, s,a,\vert\omega)}\left[ \vert\log p(b\vert s,a,\phi) \rvert \right]<\infty\ \forall \ \phi\in\Phi$.
\end{proof}
\end{proposition}

\subsection{Proof of \cref{proof:consistency}}
\label{app:proof_consistency}
Our first proof establishes uniform almost sure convergence of the ERLL (see \cref{app:assumptions_theorem_1} for a detailed definition of the ERLL and uniform almost sure convergence) under the assumptions of \cref{proof:consistency}. 

\begin{lemma}[Uniform Almost Sure Convergence of ERLL]
\label{proof:uscon}
Under Assumptions~\ref{ass:ergodic} and~\ref{ass:model_regularity}, $\normalfont{\textrm{ERLL}}_N(\phi,\Dwn)\xlongrightarrow[]{\textrm{unif}-P_\mathcal{D}} \mathbb{E}_{P_B(b\vert s,a,\omega)}\left[\log p(b\vert s,a,\phi)\right]$.
\begin{proof}
Applying the triangle inequality to \cref{def:uniform_as} yields:
\begin{align}
    &\sup_{\phi\in\Phi}\left\lvert  \normalfont{\textrm{ERLL}}_N(\phi,\Dwn)- \mathbb{E}_{P_B(b\vert s,a,\omega)}\left[\log p(b\vert s,a,\phi)\right]\right\rvert\\
    &=\sup_{\phi\in\Phi}\bigg\lvert \frac{1}{N} \sum_{i=1}^N \log p(b_i\vert s_i,a_i,\phi)-\mathbb{E}_{P_B(b,s,a\vert\omega)}\left[ \log p(b\vert s,a,\phi)\right] - \frac{1}{N}R(\phi)\bigg\rvert ,\\
      &\le\sup_{\phi\in\Phi}\bigg\lvert \frac{1}{N} \sum_{i=1}^N  \log p(b_i\vert s_i,a_i,\phi)-\mathbb{E}_{P_B(b,s,a\vert\omega)}\left[ \log p(b\vert s,a,\phi)\right] \bigg\rvert+ \frac{1}{N}\sup_{\phi\in\Phi}\bigg\lvert R(\phi)\bigg\rvert.
\end{align}
As $R(\phi)$ is bounded by \cref{ass:model_regularity}, $\lim_{N\rightarrow\infty}\frac{1}{N}\sup_{\phi\in\Phi}\bigg\lvert R(\phi)\bigg\rvert=0$, hence:
\begin{align}
    &\lim_{N\rightarrow \infty} \sup_{\phi\in\Phi}\left\lvert  \normalfont{\textrm{ERLL}}_N(\phi,\Dwn)- \mathbb{E}_{P_B(b,s,a\vert\omega)}\left[ \log p(b\vert s,a,\phi)\right]\right\rvert\\
      &\quad\le \lim_{N\rightarrow \infty}\sup_{\phi\in\Phi}\bigg\lvert \frac{1}{N} \sum_{i=1}^N  \log p(b_i\vert s_i,a_i,\phi)-\mathbb{E}_{P_B(b,s,a\vert\omega)}\left[ \log p(b\vert s,a,\phi)\right] \bigg\rvert.
\end{align}
We are therefore left to prove:
\begin{align}
     \frac{1}{N} \sum_{i=1}^N  \log p(b_i\vert s_i,a_i,\phi)\xlongrightarrow[]{\textrm{unif}-P_\mathcal{D}}\mathbb{E}_{P_B(b,s,a\vert\omega)}\left[ \log p(b\vert s,a,\phi)\right]\label{eq:uniform_convergence}
\end{align}
Theorem 3 of \citet{Andrews92} states that~(\ref{eq:uniform_convergence}) holds if i) $\Phi$ is bounded, ii) $ \log p(b\vert s,a,\phi)$ is Lipschitz in $\phi$ and iii) the empirical mean converges pointwise almost surely, that is:
\begin{align}
     \frac{1}{N} \sum_{i=1}^N  \log p(b_i\vert s_i,a_i,\phi)\xlongrightarrow[]{P_\mathcal{D}-a.s.}\mathbb{E}_{P_B(b,s,a\vert\omega)}\left[ \log p(b\vert s,a,\phi)\right]
\end{align}
Conditions i) and ii) are satisfied by \cref{ass:model_regularity}. To prove pointwise strong convergence for condition iii), we use the strong law of large numbers (SLLN) under the two sampling options presented in \cref{ass:ergodic}. As $\mathbb{E}_{P_B(b, s,a\vert\omega)}[\lvert  \log p(b\vert s,a,\phi)\rvert ]<\infty$ from \cref{proof:useful_facts}, the pointwise SLLN holds for i.i.d. samples by (for example) \citet{Williams91} Theorem 14.5 or for sampling from an ergodic Markov chain under the conditions of \cref{ass:ergodic} by Theorem 4.3 of \citet{gilks96}. Conditions i)-iii) are satisfied, hence~(\ref{eq:uniform_convergence}) holds, completing our proof.
\end{proof}
\end{lemma}

An important consequence of uniform convergence is that any sequence of supremums and infimums of the sequence of continuous, bounded functions also converges:
\begin{figure}[H]
\vspace{-0.2cm}
        \centering
        \includegraphics[scale=0.8]{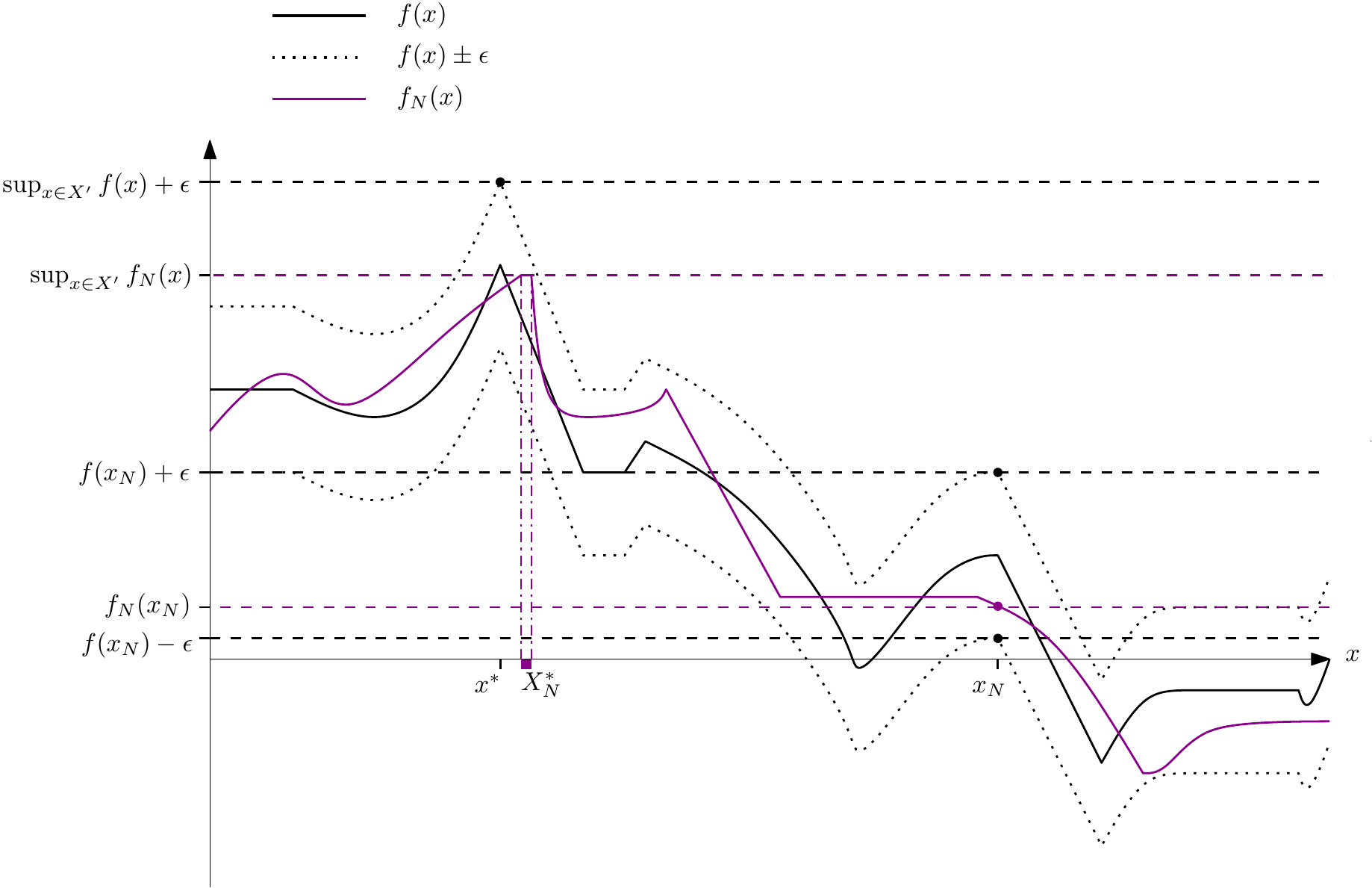}
        \vspace{-0.5cm}
        \caption{Sketch of \cref{proof:limsup_exchange}.}
        \label{fig:limsup_exchange}
\end{figure}

\begin{lemma}[Continuity of $\sup$ and $\inf$]
\label{proof:limsup_exchange}
If a sequence of continuous, bounded functions $f_N :X'\rightarrow\mathbb{R}$ converges uniformly to $f $ on $X\subseteq X'$, then 
\begin{align}
i) \lim_{N\rightarrow \infty} \sup_{x\in X} f_N(x) =\sup_{x\in X} f(x),\\
ii) \lim_{N\rightarrow \infty} \inf_{x\in X} f_N(x) =\inf_{x\in X} f(x).
\end{align}
\begin{proof}
We begin by proving i). To aid the reader's understanding, a sketch of our proof is given in \cref{fig:limsup_exchange}. The bounded sequence $ (\sup_{x\in X} f_N(x)) $ converges to $\sup_{x\in X} f(x)$ if and only if 
\begin{align}
    \limsup_{N\rightarrow\infty} \sup_{x\in X}f_N(x) = \liminf_{N\rightarrow\infty} \sup_{x\in X}f_N(x) = \sup_{x\in X} f(x).\label{eq:liminfsup_convergence}
\end{align}
As
\begin{align}
    \limsup_{N\rightarrow\infty} \sup_{x\in X}f_N(x)\ge\liminf_{N\rightarrow\infty} \sup_{x\in X} f_N(x),
\end{align}it suffices to show:
\begin{align}
    \limsup_{N\rightarrow\infty} \sup_{x\in X} f_N(x)\le \sup_{x\in X} f(x), \label{eq:lim_lower_bound}\\
    \liminf_{N\rightarrow\infty} \sup_{x\in X} f_N(x)\ge \sup_{x\in X} f(x) \label{eq:lim_upper_bound}.
\end{align}
We begin by proving the lower bound in~(\ref{eq:lim_lower_bound}). Denote the preimage  of $\sup_{x\in X}f_N(x)$ as:
\begin{align}
    X^*_N\coloneqq \left\{x:f_N(x)=\sup_{x\in X}f_N(x)\right\}.
\end{align}
By the definition of uniform convergence (\cref{def:uniform}), for every $\varepsilon>0$ there exists a $K\in\mathbb{N}$ such that for all $N\ge K$ and $x_N^*\in X_N^*$:
\begin{align}
    f_N(x_N^*) &< f(x_N^*)+\varepsilon\le \sup_{x\in X} f(x)+\varepsilon,\\
    \implies \sup_{x\in X} f_N(x) &<\sup_{x\in X} f(x)+\varepsilon. \label{eq:lower_bound_sup}
\end{align}
As $\varepsilon$ is arbitrary, it follows that:
\begin{align}
    \limsup_{N\rightarrow\infty}\sup_{x\in X} f_N(x) \le \sup_{x\in X} f(x),
\end{align}
or else it would be possible to find a $\varepsilon'>0$ and $K$ such that $\sup_{x\in X} f_N(x) \ge\sup_{x\in X} f(x)+\varepsilon'$ for all $N\ge K$, which contradicts \cref{eq:lower_bound_sup}. 

Now to show the upper bound in~(\ref{eq:lim_upper_bound}) holds using a similar analysis, consider any sequence $x_N\rightarrow x^*$ where $f(x^*)=\sup_{x\in X} f(x)$. By the definition of uniform convergence (\cref{def:uniform}), for every $\varepsilon>0$ there exists a $K\in\mathbb{N}$ such that for all $N\ge K$:
\begin{align}
    \sup_{x\in X} f_N(x)&\ge f_N(x_N) > f(x_N)-\varepsilon,\\
    \implies \sup_{x\in X} f_N(x) &> f(x_N)-\varepsilon. \label{eq:upper_bound_sup}
\end{align}
As $\varepsilon$ is arbitrary, it follows that:
\begin{align}
    \liminf_{N\rightarrow\infty}\sup_{x\in X} f_N(x) \ge \liminf_{N\rightarrow\infty} f(x_N),
\end{align}
or else it would be possible to find a $\varepsilon'>0$ and $K$ such that $\sup_{x\in X} f_N(x) \le f(x_N)-\varepsilon'$ for all $N\ge K$, which contradicts \cref{eq:upper_bound_sup}. By the definition of continuity of $f $:
\begin{align}
    \liminf_{N\rightarrow\infty}f(x_N)=\lim_{N\rightarrow\infty} f(x_N)=f(x^*)=\sup_{x\in X} f(x).
\end{align}
which proves that the upper bound holds:
\begin{align}
    \liminf_{N\rightarrow\infty} \sup_{x\in X} f_N(x)\ge \sup_{x\in X} f(x).
\end{align}
As~(\ref{eq:lim_lower_bound}) and~(\ref{eq:lim_upper_bound}) hold then i) follows immediately from \cref{eq:liminfsup_convergence}. 

To prove ii), we note that as $f_N $ and $f $ are bounded, $\inf_{x\in X} f(x)=-\sup_{x\in X}( -f(x))$ and $\inf_{x\in X} f_N(x)=-\sup_{x\in X}( -f_N(x))$. Using $-f $ and $-f_N $ in place of $f $ and $f_N $, the inequalities~(\ref{eq:lim_lower_bound}) and~(\ref{eq:lim_upper_bound}) still hold, hence: 
\begin{align}
     \limsup_{N\rightarrow\infty} \sup_{x\in X}(-f_N(x)) &= \liminf_{N\rightarrow\infty} \sup_{x\in X}(-f_N(x))=\sup_{x\in X}(- f(x)),\\
     \implies \limsup_{N\rightarrow\infty} \inf_{x\in X}f_N(x) &= \liminf_{N\rightarrow\infty} \inf_{x\in X}f_N(x)=\inf_{x\in X} f(x),\\
     \implies \lim_{N\rightarrow \infty} \inf_{x\in X} f_N(x) &=\inf_{x\in X} f(x),
\end{align}
as required.
\end{proof}
\end{lemma}
In the context of BBO, \cref{proof:limsup_exchange} implies that any sequence of minimisers/maximisers of the ERLL converges pointwise, which we now use in \cref{proof:posterior_convergence} to prove that our posterior concentrates on the KL-minimising parameters: 
\begin{figure}[H]
        \centering
        \includegraphics[scale=0.82]{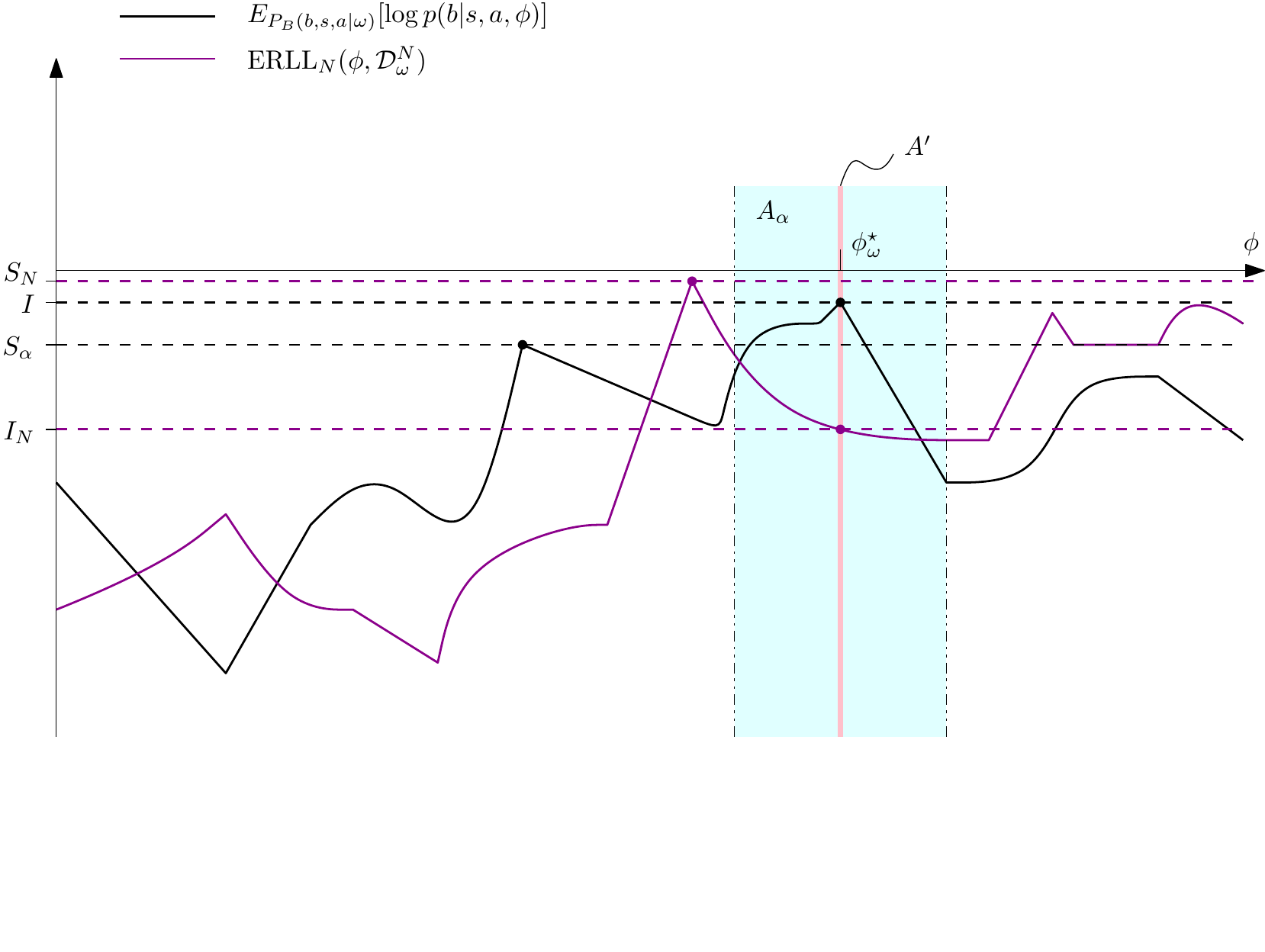}
        \vspace{-2cm}
        \caption{Sketch of \cref{proof:posterior_convergence}.}
        \label{fig:proof_posterior}
\end{figure}
\begin{lemma}[Posterior Concentration]
\label{proof:posterior_convergence}
Under Assumptions~\ref{ass:ergodic}-\ref{ass:single_minimiser}, the posterior converges weak $P_\mathcal{D}$-almost surely to a Dirac delta distribution centered on the parameters that minimises the KL divergence:
\begin{align}
    P(\phi\vert \Dwn)\xLongrightarrow[]{P_\mathcal{D}-a.s.} \delta(\phi=\phi^\star_\omega)
\end{align}
where
\begin{align}
    \phi_{\omega}^\star=\argmax_{\phi\in\Phi}\mathbb{E}_{P_B(b,s,a\vert \omega)}\left[\log p(b\vert s,a,\phi)\right]
\end{align}
\begin{proof} Consider an open ball of radius $\alpha$ centered on  $\phi_{\omega}^\star$:
\begin{align}
    A_\alpha \coloneqq\{\phi: d_\Phi(\phi,\phi_{\omega}^\star)< \alpha  \}
\end{align}
 for $\alpha>0$. From the definition of weak convergence of measures \citep{Billing09}, it suffices to show that $\lim_{N\rightarrow \infty} P(A_\alpha \vert \Dwn) =1\ P_\mathcal{D}-a.s.\ \forall\ \alpha>0$ or equivalently, 
\begin{align}
    \lim_{N\rightarrow\infty}  P(\Phi\setminus A_\alpha \vert \Dwn)=0\quad  P_\mathcal{D}-a.s. \ \forall\ \alpha>0.
\end{align}
From Kolomogrov's first axiom $ P(\Phi\setminus A_\alpha \vert \Dwn)\ge 0\ \forall\ N$, 
hence we are left to prove:
\begin{align}
    \lim_{N\rightarrow\infty}\int_{\Phi\setminus A_\alpha} dP(\phi \vert \Dwn)\le 0\quad  P_\mathcal{D}-a.s. \ \forall\ \alpha>0.
\end{align}
To aid the reader's understanding of our proof, we provide a sketch in \cref{fig:proof_posterior}. Let
\begin{align}
    S_\alpha \coloneqq \sup_{\phi\in\Phi\setminus A_\alpha}\mathbb{E}_{P_B(b,s,a\vert \omega)}\left[\log p(b\vert s,a,\phi)\right]<\mathbb{E}_{P_B(b,s,a\vert \omega)}\left[\log p(b\vert s,a,\phi_\omega^\star)\right].
\end{align}
From \cref{ass:single_minimiser}, $\phi_{\omega}^\star$ is a unique maximiser of $\mathbb{E}_{P_B(b,s,a\vert \omega)}\left[\log p(b\vert s,a,\phi)\right]$,  hence by continuity there exists some open subset $A'\subset A_\alpha $ such that
\begin{align}
    I\coloneqq & \inf_{\phi\in A' }\mathbb{E}_{P_B(b,s,a\vert \omega)}\left[\log p(b\vert s,a,\phi)\right]>S_\alpha\ \forall\ \alpha>0.
\end{align}
Writing the posterior measure as an integral over its density using \cref{eq:posterior_alternative_form} yields:
\begin{align}
P(\Phi\setminus A_\alpha \vert \Dwn)=\frac{ \int_{\Phi\setminus A_\alpha} \exp\left(\frac{N}{\sigma^2}\normalfont{\textrm{ERLL}}_N(\phi,\Dwn)\right)d\lambda(\phi)}{ \int_{\Phi} \exp\left(\frac{N}{\sigma^2}\normalfont{\textrm{ERLL}}_N(\phi,\Dwn)\right)d\lambda(\phi)},
\label{eq:sequence_posterior}
\end{align}
where $\lambda$ is the Lebesgue measure. Let $I_N \coloneqq \inf_{\phi\in A'}\normalfont{\textrm{ERLL}}_N(\phi,\Dwn)$ and $S_N \coloneqq \sup_{\phi\in \Phi\setminus A_\alpha}\normalfont{\textrm{ERLL}}_N(\phi,\Dwn)$. Consider the numerator of \cref{eq:sequence_posterior}, which we can upper bound using $S_N$:
\begin{align}
    \int_{\Phi\setminus A_\alpha} \exp\left(\frac{N}{\sigma^2}\normalfont{\textrm{ERLL}}_N(\phi,\Dwn)\right)d\lambda(\phi)&\le \int_{\Phi\setminus A_\alpha} \exp\left( \frac{N}{\sigma^2}S_N \right)d\lambda(\phi),\\
    &=\lambda( \Phi\setminus A_\alpha)\exp\left( \frac{N}{\sigma^2}S_N \right),\label{eq:upper_bound}
\end{align}
 Similarly, we can lower bound the denominator of  \cref{eq:sequence_posterior} using $I_N$ and, since the integrand is positive over $\Phi$, by changing the integration over $\Phi$ to over $A'$ :
\begin{align}
\int_{\Phi} \exp\left(\frac{N}{\sigma^2}\normalfont{\textrm{ERLL}}_N(\phi,\Dwn)\right)d\lambda(\phi)&\ge 
\int_{A'} \exp\left(\frac{N}{\sigma^2}\normalfont{\textrm{ERLL}}_N(\phi,\Dwn)\right)d\lambda(\phi),\\
&\ge \int_{A'} \exp\left(\frac{N}{\sigma^2}I_N \right)d\lambda(\phi),\\
&=\lambda(A') \exp\left(\frac{N}{\sigma^2}I_N \right). \label{eq:lower_bound}
\end{align}
Together, \cref{eq:lower_bound,eq:upper_bound} allow us to upper bound the sequence of integrals in \cref{eq:sequence_posterior}:
\begin{align}
\int_{\Phi\setminus A_\alpha} dP(\phi \vert \Dwn)&\le \frac{ \lambda(\Phi\setminus A_\alpha)\exp\left( \frac{N}{\sigma^2}S_N \right)}{\lambda(A') \exp\left(\frac{N}{\sigma^2}I_N N\right)},\\
&=\frac{ \lambda(\Phi\setminus A_\alpha)}{\lambda(A')}\exp\left( (S_N-I_N) \frac{N}{\sigma^2}\right).
\end{align}
Using \cref{proof:limsup_exchange} and \cref{proof:uscon}, we can take the limit of $S_N-I_N$ as \begin{align}
   \lim_{N\rightarrow\infty} (S_N-I_N)=S_\alpha-I,\quad \quad  P_\mathcal{D}-a.s.
\end{align}
Since $I>S_\alpha$ and hence $S_\alpha-I<0$, we can take limits to obtain the desired bound:
\begin{align}
    \lim_{N\rightarrow\infty} \int_{\Phi\setminus A_\alpha} dP(\phi \vert \Dwn)&\le \frac{ \lambda(\Phi\setminus A_\alpha)}{\lambda(A')}\lim_{N\rightarrow\infty}\exp\left( (S_N-I_N) \frac{N}{\sigma^2}\right),\\
     &\quad= \frac{ \lambda(\Phi\setminus A_\alpha)}{\lambda(A')}\exp\left( (S_\alpha-I) \frac{\lim_{N\rightarrow\infty} N}{\sigma^2}\right)\ P_\mathcal{D}-a.s.,\\
     &\quad =0 \quad  P_\mathcal{D}-a.s.\ \forall\ \alpha>0.
\end{align}
\end{proof}
\end{lemma}

\begin{theorem}
\label{proof:consistency_app}
Under Assumptions \ref{ass:ergodic}-\ref{ass:single_minimiser}, in the limit $N\rightarrow\infty$ the posterior concentrates weakly on $\phi^\star$: $i)\ P(\phi\vert \Dwn)\xLongrightarrow[]{P_\mathcal{D}-a.s.}\delta(\phi=\phi^\star_\omega)$ with $ii)\  \mathcal{B}^\star_{\omega,N}\xrightarrow{P_\mathcal{D}-a.s.}\hat{B}_{\phi^\star_\omega}$, $iii)\ \textrm{MSBBE}_N(\omega)\xrightarrow{P_\mathcal{D}-a.s.}\lVert \Qwdot - \hat{B}_{\phi^\star_\omega}\rVert_{\rho,\pi}^2$.
\begin{proof} 
\begin{align} &\textbf{Claim\ } \bm{i):}\quad P(\phi\vert \Dwn)\xLongrightarrow[]{P_\mathcal{D}-a.s.}\delta(\phi=\phi^\star_\omega).\\ 
\intertext{Claim $i)$ follows immediately from \cref{proof:posterior_convergence}. }
&\textbf{Claim\ } \bm{ii):}\quad \mathcal{B}^\star_{\omega,N}\xrightarrow{P_\mathcal{D}-a.s.}\hat{B}_{\phi^\star_\omega}.
\end{align}
To prove $ii)$ we analyse the convergence of the Bayesian Bellman operator, writing it as an expectation:
\begin{align}
    \lim_{N\rightarrow\infty}\mathcal{B}^\star_{\omega,N} =& \lim_{N\rightarrow\infty}\int_{\Phi} \Bpdot dP(\phi\vert \Dwn).
\end{align}
From $i)$, the posterior converges weakly to a Dirac delta distribution centered on $\phi_{\omega}^\star$ $P_\mathcal{D}$-almost surely. 
From \cref{proof:useful_facts} $\hat{B}_\phi$ is bounded and from \cref{ass:model_regularity} is Lipschitz in $\phi$, so we can apply the portmanteau theorem for weak convergence of measures \citep{Billing09} to find the limit: 
\begin{align}
\lim_{N\rightarrow\infty}\mathcal{B}^\star_{\omega,N} = \lim_{N\rightarrow\infty}\int_{\Phi} \Bpdot dP(\phi\vert \Dwn)=\int_{\Phi} \Bpdot d\delta(\phi=\phi^\star_\omega)=\hat{B}_{\phi^\star_\omega}\quad P_\mathcal{D}-a.s., \label{eq:convergence_projection}
\end{align}
as required. 

\begin{align} \textbf{Claim\ } \bm{iii):}\quad \textrm{MSBBE}_N(\omega)\xrightarrow{P_\mathcal{D}-a.s.}\lVert \Qwdot - \hat{B}_{\phi^\star_\omega}\rVert_{\rho,\pi}^2.
\end{align}

To prove $iii)$, we start with the definition of the MSBBE in the limit $N\rightarrow\infty$ :
\begin{align}
   \lim_{N\rightarrow\infty} \textrm{MSBBE}_N(\omega)&=\lim_{N\rightarrow\infty}\left\lVert \Qwdot - \mathcal{B}^\star_{\omega,N}\right\rVert_{\rho,\pi}^2\\
   &=\lim_{N\rightarrow\infty}\frac{1}{2}\int_{\mathcal{S}\times\mathcal{A}} \left(\Qw- \mathcal{B}^\star_{\omega,N}(s,a)\right)^2 d(\rho(s)\pi(a\vert s
   )).\label{eq:lim_msbbe}
\end{align}
To apply the dominated convergence theorem to \cref{eq:lim_msbbe}, we must show the integrand is dominated and convergences pointwise \citep{Bass13}. Firstly, from \cref{eq:convergence_projection}, we have $\mathcal{B}^\star_{\omega,N}\xrightarrow{P_\mathcal{D}-a.s.} \hat{B}_{\phi^\star_\omega}$, which is bounded by \cref{proof:useful_facts}. Now consider the Bayesian Bellman operator for finite $N$: $\mathcal{B}^\star_{\omega,N}=\int_\Phi \hat{B}_\phi dP(\phi\vert \Dwn)$. As $\hat{B}_\phi$ is bounded, $\Phi$ is compact and $P(\phi\vert \Dwn)$ is absolutely continuous with respect to the Lebesgue measure, it follows that $\hat{B}_\phi$ is $P(\phi\vert \Dwn)$-integrable hence $\mathcal{B}^\star_{\omega,N}$ is bounded for all $N$ by some positive constant $\tilde{C}$. 

Consider now the integrand in \cref{eq:lim_msbbe}. Applying the triangle inequality, we have:
\begin{align}
 &\lvert\Qw-\mathcal{B}^\star_{\omega,N}(s,a)\rvert^2\le(\lvert\Qw\rvert+\lvert\mathcal{B}^\star_{\omega,N}(s,a)\rvert)^2\le(\lvert\Qw\rvert+\tilde{C})^2.
\end{align}
From \cref{ass:model_regularity} $\Qwdot$ is bounded and so the integrand is dominated by some finite constant. The dominated convergence theorem therefore applies for \cref{eq:lim_msbbe}, hence
\begin{align}
   &\lim_{N\rightarrow\infty} \textrm{MSBBE}_N(\omega)=\left\lVert \Qwdot- \lim_{N\rightarrow\infty}\mathcal{B}^\star_{\omega,N}\right\rVert_{\rho,\pi}^2=\left\lVert \Qwdot-\hat{B}_{\phi^\star_\omega}\right\rVert_{\rho,\pi}^2\quad P_\mathcal{D}-a.s,
\end{align}
as required. 
\end{proof}
\end{theorem}
Finally, we prove our corollary, which establishes the results of \cref{proof:consistency_app} when RP approximate posterior is used in place of the true posterior.

\begin{corollary} \label{proof:consistency_approx_app}Under Assumptions \ref{ass:ergodic}-\ref{ass:RP_approximators}, results i)-iii) of  \cref{proof:consistency} hold with $P(\phi \vert \Dwn)$ replaced by the RP approximate posterior $q(\phi\vert \Dwn)$ both with or without ensembling.
\begin{proof} As results $ii)$ and $iii)$ of \cref{proof:consistency} follow directly from $i)$ under Assumptions \ref{ass:ergodic}-\ref{ass:single_minimiser}, we only need to prove $i)$ still holds with the approximate posterior, i.e. that
\begin{align}
    q(\phi\vert \Dwn)\xLongrightarrow[]{P_\mathcal{D}-a.s.}\delta(\phi=\phi^\star_\omega).
\end{align}
We consider the case of using the exact RP posterior (without ensembling) $q(\phi\vert \Dwn)\coloneqq \int_\mathcal{E} \delta(\phi\in\psi(\Dwn,\epsilon)) dP_E(\epsilon)$. From the Portmanteau Theorem for the weak convergence of measures \citep{Billing09}, it suffices to prove:
\begin{align}
    \int_\Phi f(\phi) dq(\phi\vert \Dwn)\xrightarrow[]{P_\mathcal{D}-a.s.} \int_\Phi f(\phi) d\delta(\phi=\phi^\star_\omega)=f(\phi^\star_\omega). \label{eq:approx_convergence_condition}
\end{align}
for any Lipschitz, bounded function $f:\Phi\rightarrow \mathbb{R}$. Substituting for the definition of the RP approximate posterior $q(\phi\vert \Dwn)\coloneqq \int_\mathcal{E} \delta(\phi\in\psi(\Dwn,\epsilon)) dP_E(\epsilon)$, proving ~(\ref{eq:approx_convergence_condition}) holds is equivalent to proving: 
\begin{align}
    \int_\mathcal{E} f\circ \psi_N(\epsilon) dP_E(\epsilon)\xrightarrow[]{P_\mathcal{D}-a.s.} f(\phi^\star_\omega),\label{eq:approx_convergence_condition_2}
\end{align}
for any sequence $(\psi_N(\epsilon)) $ where $\psi_N(\epsilon)\in \psi(\Dwn,\epsilon)$. Under the definition of RP (see \cref{app:randomised_priors}), it is implicitly assumed that $f\circ \psi_N:\mathcal{E}\rightarrow \mathbb{R}$ is $P_E$-integrable for any bounded, Lipschitz $f: \Phi\rightarrow \mathbb{R}$. Hence, we can apply the dominated convergence theorem to~(\ref{eq:approx_convergence_condition_2}) to derive an equivalent condition to prove that~(\ref{eq:approx_convergence_condition}) holds:
\begin{align}
    f\circ \psi_N(\epsilon) \xrightarrow[]{P_\mathcal{D}-a.s.}f(\phi^\star_\omega)\quad \forall\  \epsilon\in\mathcal{E},
\end{align}
which, from the continuity of $f$, is equivalent to proving that a sequence of minimisers of $\mathcal{L}(\phi;\Dwn,\epsilon)$ converge almost surely to the maximisers of $\mathbb{E}_{P_B(b\vert s,a,\omega)}\left[\log p(b\vert s,a,\phi)\right]$, that is:
\begin{align}
    \psi_N(\epsilon)\xlongrightarrow[]{P_\mathcal{D}-a.s.} \phi^\star_\omega\quad\forall\  \epsilon\in\mathcal{E}. \label{eq:min_max_convergence}
\end{align}
Theorem 7.33 of \citet{Rockafellar98} states that~(\ref{eq:min_max_convergence}) holds if for all $\epsilon\in\mathcal{E}$: $a)$ $\mathcal{L}(\phi;\Dwn,\epsilon)$ and $-\mathbb{E}_{P_B(b\vert s,a,\omega)}\left[\log p(b\vert s,a,\phi)\right]$ are proper, lower semi-continuous functions of $\phi$; $b)$ the sequence  $(\mathcal{L}(\phi;\Dwn,\epsilon)) $ is eventually level-bounded; and $c)$ $\mathcal{L}(\phi;\Dwn,\epsilon)$ epi-converges to $-\mathbb{E}_{P_B(b\vert s,a,\omega)}\left[\log p(b\vert s,a,\phi)\right]$. 

Condition $a)$ is trivially satisfied by the assumption of $p(b\vert s,a,\phi)$ being bounded and Lipschitz in $\phi$ in \cref{ass:model_regularity}, and as a continuous function, is lower semi-continuous as any continuous mapping between any two metric spaces is proper. Recall the definition of $\mathcal{L}(\phi;\Dwn,\epsilon)$:
\begin{align}
    \mathcal{L}(\phi;\Dwn,\epsilon)\coloneqq  -\frac{1}{N}\sum_{i=1}^N\log p(b_i\vert s_i,a_i,\phi)+\frac{1}{N}{R}(\phi-\epsilon).
\end{align}
As $p(b\vert s,a,\phi)$ is bounded and ${R}(\phi-\epsilon)$ is bounded by some $R_\textrm{max}>0$ under \cref{ass:model_regularity}, there exists a $K>0$ such that $\lvert \log p(b_i\vert s_i,a_i,\phi)\rvert\le K$ which we use to bound $\mathcal{L}(\phi;\Dwn,\epsilon)$:
\begin{align}
   \lvert \mathcal{L}(\phi;\Dwn,\epsilon)\rvert&\le \frac{1}{N}\sum_{i=1}^N\lvert \log p(b_i\vert s_i,a_i,\phi)\rvert+\frac{1}{N}\lvert {R}(\phi-\epsilon)\rvert,\\
   &\le \frac{1}{N}\sum_{i=1}^N K+\frac{R_\textrm{max}}{N},\\
   &= K +\frac{R_\textrm{max}}{N}.
\end{align}
As $\mathcal{L}(\phi;\Dwn,\epsilon)$ is bounded for all $N\in\mathbb{N}$ and $\epsilon\in\mathcal{E}$, the sequence is trivially eventually level-bounded, hence $b)$ is satisfied.

To establish epi-convergence, we first establish uniform convergence. As $\lim_{N\rightarrow\infty}\frac{1}{N}\sup_{\phi\in\Phi}\lvert {R}(\phi-\epsilon)\rvert=0$ for all $\phi\in\Phi$ and $\epsilon\in\mathcal{E}$, we can use \cref{proof:uscon} with $\textrm{ERLL}_N(\phi,\Dwn)$ replaced by $\mathcal{L}(\phi;\Dwn,\epsilon)$ to prove uniform almost sure convergence:
\begin{align}
    \mathcal{L}(\phi;\Dwn,\epsilon)\xlongrightarrow[]{\textrm{unif}-P_\mathcal{D}}-\mathbb{E}_{P_B(b\vert s,a,\omega)}\left[\log p(b\vert s,a,\phi)\right]\quad\forall\  \epsilon\in\mathcal{E}.\label{eq:unif_approx}
\end{align}
As we have already proved that each $\mathcal{L}(\phi;\Dwn,\epsilon)$ is lower semi-continuous, Proposition 7.15 a) of \citet{Rockafellar98} applies, which strengthens uniform convergence of~(\ref{eq:unif_approx}) to epi-convergence $P_\mathcal{D}$-almost surely. Condition $c)$ is satisfied and hence our desired result hold. All arguments above also hold using the ensembled approximation $q(\phi\vert \Dwn)\approx  \frac{1}{L}\sum_{l=1}^N  \delta(\phi\in\psi(\Dwn,\epsilon_l))$ if we replace expectations of any function $f(\epsilon)$ under $P_E(\epsilon)$: $\int_\mathcal{E} f(\epsilon) dP_E(\epsilon)$ with expectations over ensembles: $\frac{1}{L}\sum_{l=1}^N f(\epsilon_l)$.
\end{proof}
\end{corollary}

\subsection{Assumptions and Preliminaries for \cref{proof:approx_convergence}}
\label{app:assumptions_theorem_2}
As we are concerned with finite $N$ in our Bayesian analysis, we analyse the RP objective ignoring the factor of $\tfrac{1}{N}$ as it leaves the solution unchanged: 
\begin{align}
    \mathcal{L}(\phi;\mathcal{D}_{\omega_l}^N,\epsilon_l)\coloneqq {R}(\phi-\epsilon_l)-\sum_{i=1}^N\log p(b_i\vert s_i,a_i,\phi).
\end{align}

For convenience, we repeat the sequence of updates from \cref{sec:approximate_bbo} here for convenience:
\begin{gather}
\psi_l\leftarrow \mathcal{P}_\Omega \left(\psi_l-\alpha_k  \nabla_{\psi_l}\left({R}(\psi_l-\epsilon_l) -\log p(b_i\vert s_i,a_i,\psi_l)\right) \right),\quad \textrm{(fast)} \label{eq:app_rp_updates_fast}\\
\omega_l\leftarrow \mathcal{P}_\Omega(\omega_l-\beta_k (\omega_l-\psi_l)) .\quad \textrm{(slow)\label{eq:app_rp_updates_slow}}
\end{gather}

To analyse the limiting ODE of this sequence of updates we require an assumption regarding the parameter space:

\begin{assumption}[RP Function Spaces]\label{ass_app:RP_approximators}
     i) $\hat{Q}_{\omega_l}$ and $\hat{B}_{\omega_l}$ share a function space where $\Phi=\Omega\subset \mathbb{R}^n$ is compact, convex with a smooth boundary. ii) $\mathcal{E}\subseteq\mathbb{R}^n$ and $R(\phi-\epsilon)$ is defined for any $\phi\in\Phi, \epsilon\in\mathcal{E}$.
\end{assumption}
\cref{ass_app:RP_approximators} i) is easily satisfied by defining $\Omega$ to be a closed ball of arbitrary radius, which should be applicable to the majority of cases where parametric function approximators are used. Provided $\Omega$ is large enough, there is diminishing probability that our update ever leaves $\Omega$, especially with regularisation. Hence we do not expect to require projection in practice unless the environment is particularly ill posed. In the unlikely eventuality that projection is required, using a ball makes projection simple as the operator projects back along the line connecting a point to the ball's origin, which defines a normal. In order to derive the limiting ODEs of our updates, we characterise the directional derivative of the projection operator in \cref{proof:directional_derivative}:
\begin{proposition}[\citet{Shapiro88}]
\label{proof:directional_derivative}
The directional derivative of the projection operator at $\omega\in\Omega$ in the direction $y\in \mathbb{R}^{n}$ given by:
\begin{align}
    \Gamma_\omega \left (y\right)\coloneqq \lim_{\epsilon\downarrow 0}\left(\frac{\mathcal{P}_\Omega\left(\omega+\epsilon y \right)-\omega}{\epsilon}\right)
\end{align}
which always exists under \cref{ass_app:RP_approximators} and is equivalent to the projection of $y$ onto the tangent cone $T_\Omega(\omega)$:
\begin{align}
    \Gamma_\omega \left (y\right)=\mathcal{P}_{T_\Omega(\omega)}(y)\coloneqq \argmin_{t\in T_\Omega(\omega)}\lVert y - t\rVert_2^2
\end{align}
where $T_\Omega(\omega)\coloneqq \normalfont{\textrm{Closure}}\left(\bigcup_{\varepsilon>0}\bigcup_{\omega'\in\Omega} \frac{1}{\varepsilon} (\omega'-\omega) \right)$

\begin{proof} See \citet{Shapiro88}.
\end{proof}
\end{proposition}

\begin{wrapfigure}{r}{0.4\textwidth}
  \begin{center}
  \vspace{-0.8cm}
    \includegraphics[width=0.38\textwidth]{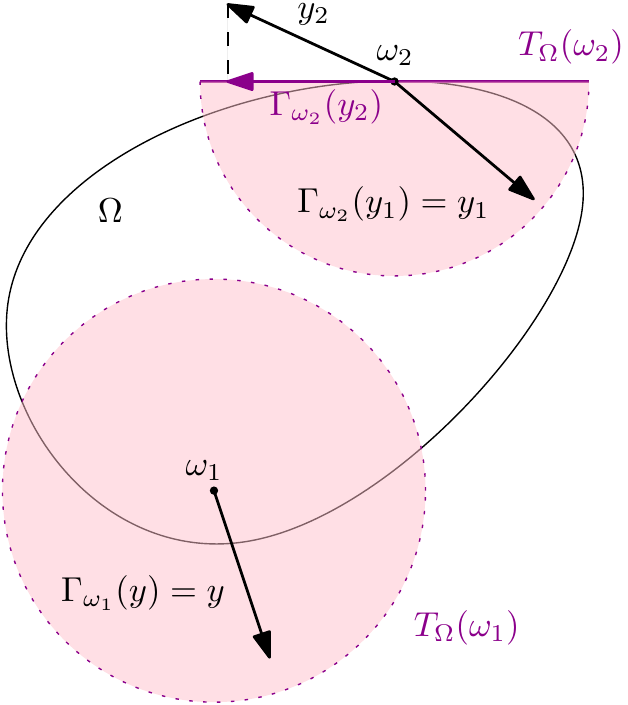}
  \end{center}
  \caption{Sketch of Tangent Cones in $\Omega$}
  \label{fig:tangent_sketch}
  \vspace{-0.5cm}
\end{wrapfigure}
The directional derivative of the projection operator has been well studied \citep{ZARANTONELLO71,Shapiro87,Shapiro88,Borkar08} and we provide an informal sketch of its interpretation in \cref{fig:tangent_sketch} for the reader's intuition. We see that for any point $\omega_1\in\Omega$ in the interior of $\Omega$, the projection operator is simply the identity function: $\Gamma_{\omega_1} \left (y\right)=y$ as $y\in T_\Omega(\omega_1)$. For any point $\omega_2$ on the boundary of $\Omega$, the tangent cone is the closure of the cone formed by all half-lines emanating from $\omega_2$ intersecting $\Omega$ in at least one point distinct from $\omega_2$. There are two cases to consider for boundary points. Firstly, when the directional vector $y_1$ defines a half-line from $\omega_2$ that intersects $\Omega$, the projection operator is the identity $\Gamma_{\omega_2} \left (y\right)=y$. In the second case when the directional vector $y_2$ defines a half-line from $\omega_2$ that leaves $\Omega$, $\Gamma_{\omega_2} \left (y_2\right)$ returns the nearest element along the boundary of $T_\Omega(\omega_2)$ (solid purple) according to the projection $\mathcal{P}_{T_\Omega(\omega_2)}(y_2)$.

Using the directional derivative of the projection operator, we can derive the limiting ODE of   update~(\ref{eq:app_rp_updates_fast}):
\begin{align}
\partial_t \psi_l(t)&=-\Gamma_{\psi_l(t)
}(\nabla_{\psi_l}\mathcal{L}(\psi_l(t);\mathcal{D}_{\omega_l}^N,\epsilon_l))\label{eq:fast_ode}
\end{align}
Under standard ODE analysis, any equilibria of \cref{eq:fast_ode} satisfy $-\Gamma_{\psi_l(t)
}(\nabla_{\psi_l}\mathcal{L}(\psi_l(t);\mathcal{D}_{\omega_l}^N,\epsilon_l))=0$ and we denote an asymptotically stable local equilibrium as $\psi_l^\circledast(\omega_l)$. Of course, there may be several or even infinite stable local equilibrium for the ODE, but as our assumptions only require the existence of at least one stable attractor within the domain of attraction defined by the initial parametrisations, we lose no generality by considering $\psi_l^\circledast(\omega_l)$ in isolation.

As the fast update converges asymptotically quicker than the slower update, we consider $\psi_l$ to be equilibrated at $\psi_l^\circledast(\omega_l)$ when analysing  the limiting ODE of the slow update. We make this argument rigorously in \cref{proof:approx_convergence_app}, from which we derive the limiting ODE for the update~(\ref{eq:app_rp_updates_slow}) as:
\begin{align}
    \partial_t \omega_l(t)&=-\Gamma_{\omega_l(t)
}(\omega_l(t)-\psi_l^\circledast(\omega_l(t))).\label{eq:slow_ode}
\end{align}
Crucially, this provides reassurance that our updates and any equilibria satisfying $\Gamma_{\omega_l(t)
}(\omega_l(t)-\psi_l^\circledast(\omega_l))=0$ preserve the dependence of $\psi_l$ on $\omega_l$. We denote an asymptotically stable local equilibrium of \cref{eq:slow_ode} as $\omega_l^\circledast$.

\begin{assumption}[Two-timescale Regularity]\label{ass_app:two_timescale}  i) $\nabla_{\psi_l}{R}(\psi_l-\epsilon_l)$, $\nabla_{\psi_l}\log p(b_i\vert s_i,a_i,\psi_l)$ and $\Gamma_{\psi_l}\left[ \nabla_{\psi_l}\mathcal{L}(\psi_l;\mathcal{D}_{\omega_l}^N,\epsilon) \right]$ are Lipschitz in $\psi_l$, $\Gamma_{\omega_l}\left[- (\omega_l-\psi_l)\right]$ is Lipschitz in $\omega_l$ and $(b_i,s_i,a_i)\sim \normalfont{\textrm{Unif}}(\mathcal{D}_{\omega_l}^N)$; ii) $\psi^\circledast (\omega_l)$ and $\omega^\circledast_l$ are local aysmptotically stable attractors of the limiting ODEs of updates~(\ref{eq:rp_updates_fast}) and~(\ref{eq:rp_updates_slow}) respectively and $\psi^\circledast_l(\omega_l)$ is Lipschitz in $\omega_l$; and iii) The stepsizes satisfy: $\lim_{k\rightarrow\infty} \frac{\beta_k}{\alpha_k}=0,\ 
    \sum_{k=1}^\infty \alpha_k =\sum_{k=1}^\infty \beta_k=\infty,\ \sum_{k=1}^\infty\left( \alpha_k^2+ \beta_k^2\right)<\infty$.
\end{assumption} 
Compared the presentation in the main body of our paper, we have introduce the additional requirement in $i)$ that  $\Gamma_{\psi_l}\left[ \nabla_{\psi_l}\mathcal{L}(\psi_l;\mathcal{D}_{\omega_k}^N,\epsilon) \right]$ and $\Gamma_{\omega_l}\left[- (\omega_l-\psi_l)\right]$ are Lispchtiz. We justify this exclusion from the main body of our text for two reasons: firstly, using the same arguments as \cref{proof:bounded_facts} below, it is easy to establish that $\nabla_{\psi_l}\mathcal{L}(\psi_l;\mathcal{D}_{\omega_l}^N,\epsilon)$ is Lipschitz and by inspection, $(\omega_l-\psi_l)$ is Lipschitz and hence Lipschitzness of $\Gamma_{\psi_k}\left[ \nabla_{\psi_l}\mathcal{L}(\psi_l;\mathcal{D}_{\omega_l}^N,\epsilon) \right]$ and $\Gamma_{\omega_l}\left[- (\omega_l-\psi_l)\right]$ can be established provided $\Omega$ is sufficiently large to contain the entire limiting flow of the ODE. For this reason we are essentially encouraged to consider the non-projected form \citep{Chung19}. Secondly, this subtlety is often ignored altogether by existing papers considering two-timescale analysis \citep{Maei09} as it carries a heavy expositional burden but does not effect the algorithm in practice. 

As noted by \citet{Heusel17}, the assumption of locally asymptotically stable ODEs in \cref{ass_app:two_timescale} ii) can be ensured by an additional weight decay term in the loss function which increases the eigenvalues of the Hessian. This fits naturally in a Bayesian setting where prior regularisation introduces weight decay into our objectives. Finally, \cref{ass_app:two_timescale} iii) extends the classic Robbins-Munro stepsize conditions \citep{Robbins51} to the two-timescale case. The additional assumption $\lim_{k\rightarrow\infty} \frac{\beta_k}{\alpha_k}=0$ ensures that the faster timescale update converges asymptotically faster than the slower update. We now use \cref{ass_app:two_timescale} to establish the boundedness of several quantities that will be essential for our main proof:
\begin{proposition}\label{proof:bounded_facts}
Under Assumptions~\ref{ass_app:RP_approximators} and~\ref{ass_app:two_timescale}, the quantities i) $\nabla_{\psi_l}{R}(\psi_l-\epsilon_l)$ and $\nabla_{\psi_l}\log p(b_i\vert s_i,a_i,\psi_l)$; ii) $\sum_{j\ne i}^N \nabla_{\psi_l} \log p(b_j\vert s_j,a_j,\psi_l)$; and iii)  $\nabla_{\psi_l}\mathcal{L}(\psi_l;\mathcal{D}_{\omega_l}^N,\epsilon_l)$ are all bounded.
\begin{proof}
We establish $i)$ by noting that from  \cref{ass_app:two_timescale}, $\nabla_{\psi_l}{R}(\psi_l-\epsilon_l)$ and $\nabla_{\psi_l}\log p(b_i\vert s_i,a_i,\psi_l)$ are Lipschitz in $\psi_l$, and from \cref{ass_app:RP_approximators}, $\Phi$ is compact, hence as any Lipschitz function defined on a compact set must be bounded, it follows that $\nabla_{\psi_l}{R}(\psi_l-\epsilon_l)$ and $\nabla_{\psi_l}\log p(b_i\vert s_i,a_i,\psi_l)$ are bounded. To prove $ii)$, we bound $\sum_{j\ne i}^N \nabla_{\psi_l} \log p(b_j\vert s_j,a_j,\psi_l)$ using the triangle inequality:
\begin{align}
    \left\lvert \sum_{j\ne i}^N \nabla_{\psi_l}\log p(b_j\vert s_j,a_j,\psi_l)\right\rvert\le & \sum_{j\ne i}^N\left\lvert \nabla_{\psi_l}\log p(b_j\vert s_j,a_j,\psi_l)\right\rvert,\\
    <&N\sup_{j\in\{1:N\}}\left\lvert \nabla_{\psi_l}\log p(b_j\vert s_j,a_j,\psi_l)\right\rvert,
\end{align}
which is bounded from $i)$ and the fact that $N$ is finite in our Bayesian regime. To prove $iii)$, we establish a similar bound:
\begin{align}
    \left\lvert\nabla_{\psi_l}\mathcal{L}(\psi_l;\mathcal{D}_{\omega_l}^N,\epsilon_l)\right\rvert&= \left\lvert\nabla_{\psi_l}\left( {R}(\psi_l-\epsilon_l)-\sum_{i=1}^N\log p(b_i\vert s_i,a_i,\psi_l)\right)\right\rvert,\\
    &\le \left\lvert \nabla_{\psi_l}{R}(\psi_l-\epsilon_l)\right\rvert+\sum_{i=1}^N\left\lvert \nabla_{\psi_l}\log p(b_i\vert s_i,a_i,\psi_l)\right\rvert,\\
    &\le \left\lvert \nabla_{\psi_l}{R}(\psi_l-\epsilon_l)\right\rvert+N\sup_{i\in\{1:N\}}\left\lvert \nabla_{\psi_l}\log p(b_i\vert s_i,a_i,\psi_l)\right\rvert,
\end{align}
which is bounded from $i)$ and $N$ being finite.
\end{proof}
\end{proposition}

\subsection{Proof of \cref{proof:approx_convergence}}
\label{app:proof_approx_convergence}
To ease the notational burden of our proof, we drop dependence on the ensemble index $l\in\{1:L\}$ as the convergence proof is the same for all ensembles. We formalise updates~(\ref{eq:app_rp_updates_fast}) and~(\ref{eq:app_rp_updates_slow}) by analysing the recursive sequence:
\begin{gather}
\psi_{k+1}= \mathcal{P}_\Omega \left(\psi_k-\alpha_k  \nabla_{\psi}\left({R}(\psi_k-\epsilon) -\log p(b_i\vert s_i,a_i,\psi_k)\right) \right),\quad \textrm{(fast)} \label{eq:rp_fast_recursion}\\
\omega_{k+1}= \mathcal{P}_\Omega(\omega_k-\beta_k (\omega_k-\psi_k)) .\quad \textrm{(slow)\label{eq:rp_slow_recursion}}
\end{gather}
where $i\sim \textrm{Unif}(\{1:N\})$.
\begin{theorem}
\label{proof:approx_convergence_app}
 If Assumptions~\ref{ass:ergodic} to~\ref{ass:two_timescale} hold, $\psi_l$ and $\omega_l$ converge to $\psi^\circledast_l(\omega^\circledast_l)$ and $\omega^\circledast_l$ almost surely.
\begin{proof} Firstly, we define the martingale:
\begin{align}
      \mathcal{M}^i_k\coloneqq&-\nabla_\psi\sum_{j\ne i}^N \log p(b_j\vert s_j,a_j,\psi_k).
\end{align}
Using the identity:
\begin{align}
    \nabla_{\psi}\left({R}(\psi_k-\epsilon) -\log p(b_i\vert s_i,a_i,\psi_k)\right) =\nabla_{\psi}\mathcal{L}(\psi_k;\mathcal{D}_{\omega_k}^N,\epsilon) -\mathcal{M}^i_k\label{eq:martingale_indentity}
\end{align}
we re-write the recursive update in \cref{eq:rp_fast_recursion} as a martingale difference sequence:
\begin{align}
    \psi_{k+1}= \mathcal{P}_\Omega \left(\psi_k-\alpha_k( \nabla_{\psi}\mathcal{L}(\psi_k;\mathcal{D}_{\omega_k}^N,\epsilon) -\mathcal{M}^i_k)\right).
\end{align}
As $\mathcal{P}_\Omega $ has a well-defined directional derivative under \cref{proof:directional_derivative}, we can re-write the fast update using a series expansion about $\psi_k$:
\begin{align}
    &\mathcal{P}_\Omega\left(\psi_k-\alpha_k( \nabla_{\psi}\mathcal{L}(\psi_k;\mathcal{D}_{\omega_k}^N,\epsilon) -\mathcal{M}^i_k)\right)\\
    &\quad=\mathcal{P}_\Omega(\psi_k)-\Gamma_{\psi_k}\left[\alpha_k( \nabla_{\psi}\mathcal{L}(\psi_k;\mathcal{D}_{\omega_k}^N,\epsilon) -\mathcal{M}^i_k)\right]+o(\alpha_n^2),\\
   &\quad=\psi_k+\alpha_k\Gamma_{\psi_k}\left[\mathcal{M}^i_k- \nabla_{\psi}\mathcal{L}(\psi_k;\mathcal{D}_{\omega_k}^N,\epsilon) \right]+o(\alpha_n^2),\\
     &\quad=\psi_k+\alpha_k\left(\Gamma_{\psi_k}\left[\mathcal{M}^i_k\right]-\Gamma_{\psi_k}\left[ \nabla_{\psi}\mathcal{L}(\psi_k;\mathcal{D}_{\omega_k}^N,\epsilon) \right]+o(\alpha_n)\right),
\end{align}
where we have used the identity $\mathcal{P}_\Omega(\psi_k)=\psi_k$ in deriving the second equality and the linearity of the derivative for the final equality. Consider now the slow update, which we can write using a similar expansion:
\begin{align}
     \mathcal{P}_\Omega(\omega_k-\beta_k (\omega_k-\psi_k))&=\mathcal{P}_\Omega(\omega_k)-\Gamma_{\psi_k}\left[\beta_k (\omega_k-\psi_k)\right]+o(\beta_k^2),\\
    &=\omega_k+\beta_k\left(\Gamma_{\psi_k}\left[- (\omega_k-\psi_k)\right]+o(\beta_k)\right).
\end{align}

We write the two updates together here for clarity:
\begin{align}
    &\psi_{n+1} =\psi_k+\alpha_k\left(\Gamma_{\psi_k}\left[\mathcal{M}^i_k\right]-\Gamma_{\psi_k}\left[ \nabla_{\psi}\mathcal{L}(\psi_k;\mathcal{D}_{\omega_k}^N,\epsilon) \right]+o(\alpha_n)\right),\label{eq:slow_update}\\
     &\omega_{n+1}=\omega_k+\beta_k\left(\Gamma_{\psi_k}\left[- (\omega_k-\psi_k)\right]+o(\beta_k)\right).\label{eq:fast_update}
\end{align}
Our proof adapts Theorem 1 of \citet{Heusel17}, which is a generalisation of \citet{Borkar08} using the proofs of \citet{Karmakar18}. There is one subtle difference between the updates of \citet{Heusel17} and the updates for BBO in \cref{eq:fast_update,eq:slow_update}: in \cref{eq:fast_update,eq:slow_update}, there is an additional term $o(\alpha_k)$ and $o(\beta_k)$ such that by \cref{ass_app:two_timescale}, $\lim_{k\rightarrow\infty} o(\alpha_k)=o(\beta_k)=0$. $o(\beta_k)$ will be absorbed in the error term $\epsilon_n$ in the slow update in Eq. 14 and $o(\alpha_k)$ will be absorbed in the error term $\epsilon_n'$ in the fast update in Eq. 17 of \citet{Karmakar18}. As $\lim_{n\rightarrow\infty}\epsilon_n=\epsilon_n'=0$, these additional terms $o(\alpha_n)$ and $o(\beta_n)$ leave the proof of Theorem 5 of \citet{Karmakar18} unaffected. Theorem 1 of \citet{Heusel17} therefore remains unchanged, which uses Theorem 5 of \citet{Karmakar18} to establish that the sequences of \cref{eq:slow_update,eq:fast_update} converge almost surely to a set of local attractors of the underlying ODE:
\begin{align}
    \{\omega_k,\psi_k\} \xrightarrow{k\rightarrow\infty}   \{\omega^\circledast,\psi^\circledast(\omega^\circledast)\}\ a.s.
\end{align} 
if
\begin{enumerate}
    \item $\Gamma_{\psi_k}\left[- \nabla_{\psi}\mathcal{L}(\psi_k;\mathcal{D}_{\omega_k}^N,\epsilon) \right]$ is Lipschitz in $\psi_k$ and $\Gamma_{\omega_k}\left[- (\omega_k-\psi_k)\right]$ is Lipschitz in $\omega_k$ under \cref{ass_app:two_timescale},
    \item Stepsizes $\alpha_k$ and $\beta_k$ satisfy \cref{ass_app:two_timescale}
    \item For all $k$, $(\Gamma_{\psi_k}\left[\mathcal{M}^i_k\right])$ is a martingale difference sequence with respect to the filtration of increasing $\sigma$-algebras:
\begin{align}
    \mathscr{F}_k\coloneqq\sigma\left(\psi_j,\Gamma_{\psi_j}\left[\mathcal{M}^i_j\right]:j=1,...k\right),
\end{align}
with
\begin{align}
    \mathbb{E}\left[\left\lVert \Gamma_{\psi_k}\left[\mathcal{M}^i_k\right]\right\rVert^2\vert \mathscr{F}_k\right]&<\infty,
\end{align}
    \item The limiting ODEs $\partial_t \psi(t)=-\Gamma_{\psi(t)
}(\nabla_{\psi}\mathcal{L}(\psi(t);\mathcal{D}_{\omega}^N,\epsilon))$ and $\partial_t \omega(t)=-\Gamma_{\omega(t)
}(\omega_l(t)-\psi_l^\circledast(\omega_l(t)))$  satisfy \cref{ass_app:two_timescale},
    \item $\sup_n\lVert\psi_n\rVert < \infty$ and $\sup_n\lVert\omega_n\rVert < \infty$ a.s.
\end{enumerate}
As 5) is satisfied by virtue of the projection operators, we are left to prove 3). 

As $\Gamma_{\psi_k}\left[\cdot\right]$ is a linear operator with $\Gamma_{\psi_k}\left[0\right]=0$,  $(\Gamma_{\psi_k}\left[\mathcal{M}^i_k\right])$ is a Martingale difference sequence if $( \mathcal{M}^i_k)$ is a Martingale difference sequence. To prove 3) holds, it therefore suffices to show that $(\mathcal{M}^i_k)$ is a Martingale difference sequences with respect to the filtration of increasing $\sigma$-algebras:
\begin{align}
    \mathscr{F}_k'\coloneqq\sigma\left(\psi_j,\mathcal{M}^i_j:j=1,...k\right),
\end{align}
To prove this, we observe that: i) $\mathcal{M}^i_k$ is $\mathscr{F}_k'$-measurable for all $k$ by construction, from \cref{proof:bounded_facts} $\mathcal{M}^i_k$ is bounded, hence ii) $\mathbb{E}\left[\mathcal{M}^i_k\right]<\infty$ and from \cref{eq:martingale_indentity} iii) the conditional expectation of $\mathcal{M}^i_k$ satisfies for all $k$:
\begin{align}
    \mathbb{E}\left[\mathcal{M}^i_k\vert \mathscr{F}_k'\right] &=
    \nabla_{\psi}\mathcal{L}(\psi_k;\mathcal{D}_{\omega_k}^N,\epsilon)-\mathbb{E}\left[\nabla_{\psi}\left({R}(\psi_k-\epsilon) -\log p(b_i\vert s_i,a_i,\psi_k)\right)\Bigg\vert \mathscr{F}_k'\right],\\
    &=\nabla_{\psi}\mathcal{L}(\psi_k;\mathcal{D}_{\omega_k}^N,\epsilon)-\nabla_{\psi}{R}(\psi_k-\epsilon) +\mathbb{E}\left[\nabla_{\psi}\log p(b_i\vert s_i,a_i,\psi_k)\vert \mathscr{F}_k'\right],\\
    &=\nabla_{\psi}\mathcal{L}(\psi_k;\mathcal{D}_{\omega_k}^N,\epsilon)-\nabla_{\psi}{R}(\psi_k-\epsilon) +\sum_{i=1}^N\nabla_{\psi}\log p(b_i\vert s_i,a_i,\psi_k),\\
     &=0.
\end{align}
Together i) to iii) form the definition of a martingale difference sequence with respect to the filtration $(\mathscr{F}_k')$  \citep{Williams91}. As we sample from a discrete uniform distribution and from \cref{proof:bounded_facts} $\nabla_{\psi_l}{R}(\psi_l-\epsilon_l)$, $\nabla_{\psi_l}\log p(b_i\vert s_i,a_i,\psi_k)$ and $\nabla_{\psi}\mathcal{L}(\psi_k;\mathcal{D}_{\omega_k}^N,\epsilon)$ are bounded, the variance of the updates is finite: 
\begin{align}
    &\mathbb{E}\left[\lVert\mathcal{M}^i_k\rVert^2\vert \mathscr{F}_k'\right]\\ &\quad=\mathbb{E}\left[\left( \nabla_{\psi}\mathcal{L}(\psi_k;\mathcal{D}_{\omega_k}^N,\epsilon)-\nabla_{\psi}\left({R}(\psi_k-\epsilon) -\log p(b_i\vert s_i,a_i,\psi_k)\right)\right)^2\Bigg\vert \mathscr{F}_k'\right] < \infty,
\end{align}
and hence 3) is satisfied as required. 
\end{proof}
\end{theorem}
\subsection{A Frequentist Analysis of \cref{proof:approx_convergence}}
\label{app:frequentist_convergence}
In the frequentist regime, we are concerned with the objective in the limit $N\rightarrow\infty$ and we consider data to be arriving online. Our convergence analysis holds if we replace the assumption of uniform sampling from the dataset $\Dwn$ with (assuming samples have finite variance) sampling i.i.d. from the underlying data distribution $(b_i,s_i,a_i)\sim P_B$. The condition of i.i.d. unbiased estimates leaves our analysis unchanged, however our proof can be extended to cases where the updates are sampled from a Markov chain. In particular, if assumptions (A1), (A5) and (A6)' of \citet{Karmakar18} are satisfied, \cref{proof:approx_convergence_app} can be trivially extended to the non-i.i.d. case. We do not discuss these assumptions further as they have been extensively discussed (with a case study) in \citet{Heusel17} which sufficiently covers extending BBO to sampling from a Markov chain in the frequentist regime. Note that the ergodicity of the Markov chain in \cref{ass_app:ergodic} greatly simplifies their verification as the set of ergodic occupancy measures is a singleton.

%% file: Appendix/app_derivations.tex
\section{Derivations}
\emph{To ease the notational burden of our derivations, we abuse notation slightly by writing the density and the distribution of the policy as $\pi(a\vert s)$ and the density and distribution of the state sampling distribution as $\rho(s)$.}
\subsection{Posterior Density Derivation}
\label{app:posterior_derivation}

We now derive the posterior density in \cref{eq:posterior}, which is identical for the two sampling regimes outlined in \cref{ass_app:ergodic}. We start by deriving the likelihood of the data $\Dwn$ for the i.i.d. case:
\begin{align}
    p(\Dwn\vert \phi) = \prod_{i=1}^N \rho(s_i)\pi(a_i\vert s_i)p(b_i\vert s_i,a_i,\phi).
\end{align}
Using our prior $p(\phi)$ and Bayes' rule, we can infer the posterior as:
\begin{align}
    p(\phi\vert \Dwn) &= \frac{p(\Dwn\vert \phi)p(\phi)}{\int_\Phi p(\Dwn\vert \phi)dP(\phi) },\\
    &=\frac{\prod_{i=1}^N\left( \rho(s_i)\pi(a_i\vert s_i)p(b_i\vert s_i,a_i,\phi)\right)p(\phi)}{\int_\phi \prod_{i=1}^N \left(\rho(s_i)\pi(a_i\vert s_i)p(b_i\vert s_i,a_i,\phi)\right)dP(\phi) },\\
    &=\frac{\prod_{j=1}^N (\rho(s_j)\pi(a_j\vert s_j)) \prod_{i=1}^N p(b_i\vert s_i,a_i,\phi)p(\phi)}{ \int_\phi \prod_{j=1}^N (\rho(s_j)\pi(a_j\vert s_j)) \prod_{i=1}^N  p(b_i\vert s_i,a_i,\phi)dP(\phi) },\\
     &=\frac{\prod_{j=1}^N (\rho(s_j)\pi(a_j\vert s_j)) \prod_{i=1}^N p(b_i\vert s_i,a_i,\phi)p(\phi)}{ \prod_{j=1}^N (\rho(s_j)\pi(a_j\vert s_j)) \int_\phi \prod_{i=1}^N  p(b_i\vert s_i,a_i,\phi)dP(\phi) },\\
    &=\frac{\prod_{i=1}^N p(b_i\vert s_i,a_i,\phi)p(\phi)}{ \int_\phi \prod_{i=1}^N  p(b_i\vert s_i,a_i,\phi)dP(\phi) }. \label{eq:iid_posterior}
\end{align}
For sampling from an ergodic Markov chain, we denote the initial state-action density as $ p_0(s,a)$ and the transition density as $p(s',a'\vert s,a)$. For sampling on-policy these distributions are defined as:
\begin{align}
    p_0(s,a)&\coloneqq p_0(s)\pi(a\vert s)\\
    p(s',a'\vert s,a)&\coloneqq p(s'\vert s,a)\pi(a'\vert s').
\end{align}
We write our likelihood as:
\begin{align}
    p(\Dwn\vert \phi) &=p_0(s_1,a_1)p(b_1\vert s_1,a_1,\phi)\prod_{i=2}^N (p(s_i\,a_i\vert s_{i-1},a_{i-1})p(b_i\vert s_i,a_i,\phi)),\\
    &=p_0(s_1,a_1)\prod_{j=2}^N p(s_j\,a_j\vert s_{j-1},a_{j-1})\prod_{i=1}^N p(b_i\vert s_i,a_i,\phi),\\
    &=p(S_N,A_N) \prod_{i=1}^N p(b_i\vert s_i,a_i,\phi),
\end{align}
where $S_N\coloneqq\{s_1,...s_N\} $, $A_N\coloneqq\{a_1,...a_N\}$ and 
\begin{align}
    p(S_N,A_N)=p_0(s_1,a_1)\prod_{j=2}^N p(s_j\,a_j\vert s_{j-1},a_{j-1}).
\end{align}
We now infer the posterior using Bayes' rule:
\begin{align}
p(\phi\vert \Dwn) &= \frac{p(\Dwn\vert \phi)p(\phi)}{\int_\Phi p(\Dwn\vert \phi)dP(\phi) },\\
&=\frac{ p(S_N,A_N)p(b_i\vert s_i,a_i,\phi)p(\phi)}{ p(S_N,A_N)\int_\phi \prod_{i=1}^N  p(b_i\vert s_i,a_i,\phi)dP(\phi) },\\
    &=\frac{\prod_{i=1}^N p(b_i\vert s_i,a_i,\phi)p(\phi)}{ \int_\phi \prod_{i=1}^N  p(b_i\vert s_i,a_i,\phi)dP(\phi) },
\end{align}
which has the same form as the i.i.d. case in \cref{eq:iid_posterior}. \subsection{Gaussian BBO Derivation}
\label{app:gaussian_bbo}
Now, using a Gaussian model:
\begin{align}
    P(b\vert s,a,\phi)\coloneqq \mathcal{N}(\Bp,\sigma^2),
\end{align}
and defining the log-normalisation constant as  $c_\textrm{norm}\coloneqq \log\int_\phi \prod_{i=1}^N  p(b_i\vert s_i,a_i,\phi)dP(\phi)$, we can derive the exact form of the $\log$-posterior given in \cref{eq:gaussian_posterior}:
\begin{align}
    p(\phi\vert \Dwn)&= \exp(-c_\textrm{norm})\prod_{i=1}^N p(b_i\vert s_i,a_i,\phi)p(\phi),\\
    &=\exp(-c_\textrm{norm})\prod_{i=1}^N \exp\left(-\frac{1}{2\sigma^2}(b_i-\hat{B}_\phi(s_i,a_i))^2\right)\exp\left(-R(\phi)\right),\\
    &=\exp\left(-c_\textrm{norm}-\frac{1}{2\sigma^2}\sum_{i=1}^N(b_i-\hat{B}_\phi(s_i,a_i))^2-R(\phi)\right),\\
    \implies -\log p(\phi\vert \Dwn)&=c_\textrm{norm}+\sum_{i=1}^N \frac{(b_i-\hat{B}_\phi(s_i,a_i))^2}{2\sigma^2} +R(\phi),
\end{align}
as required. We now derive the minimising objective for the set of KL minimising parameters in \cref{eq:gaussian_posterior} by substituting for the definition of the Gaussian model and ignoring terms independent of $\phi$:
\begin{align}
    \phi^\star_\omega&\coloneqq \argmin_{\phi\in\Phi} \kl{P_B(b, s,a\vert\omega)}{P(b,s,a\vert\phi)}\\
    &=\argmin_{\phi\in\Phi} \mathbb{E}_{P_B(b, s,a\vert\omega)}\left[ -\log p(b,s,a\vert \phi)\right],\\
    &=\argmin_{\phi\in\Phi} \mathbb{E}_{P_B(b, s,a\vert\omega)}\left[ -\log p(b\vert s,a, \phi)\rho(s)\pi(a\vert s)\right],\\
     &=\argmin_{\phi\in\Phi} \mathbb{E}_{P_B(b, s,a\vert\omega)}\left[ -\log p(b\vert s,a, \phi)\right],\\
     &=\argmin_{\phi\in\Phi} \mathbb{E}_{P_B(b, s,a\vert\omega)}\left[ (b-\hat{B}_\phi(s,a))^2\right].\label{eq:msbbe_min}
\end{align}
Now, we consider the inner expectation with respect to $P_B(b\vert \cdot)$ which we denote as $ \mathbb{E}_B[\cdot]$ for convenience:
\begin{align}
     \mathbb{E}_B\left[ (b-\hat{B}_\phi)^2\right]&=\mathbb{E}_B\left[ b^2\right]-2\mathbb{E}_B\left[ b\right]\hat{B}_\phi+\hat{B}_\phi^2,\\
     &=\mathbb{E}_B\left[ b^2\right]-2\mathcal{B}[\hat{Q}_\omega]\hat{B}_\phi+\hat{B}_\phi^2,
\end{align}
Denoting the conditional variance of $b$ as $\mathbb{V}\left[b\right]$, we substitute for $\mathbb{E}_B\left[ b^2\right]=\mathbb{V}\left[b\right]+\mathcal{B}[\hat{Q}_\omega]^2$:
\begin{align}
     \mathbb{E}_B\left[ (b-\hat{B}_\phi)^2\right]&=\mathbb{V}\left[b\right]+\mathcal{B}[\hat{Q}_\omega]^2-2\mathcal{B}[\hat{Q}_\omega]\hat{B}_\phi+\hat{B}_\phi^2,\\
     &=\mathbb{V}\left[b\right]+(\mathcal{B}[\hat{Q}_\omega]-\hat{B}_\phi)^2.
\end{align}
As $\mathbb{V}\left[b\right]$ has no dependence on $\phi$ and we are finding $\argmin_{\phi\in\Phi}$, we can ignore it from our derivation. Taking expectations with respect to $\rho$ and $\pi$, scaling by $\tfrac{1}{2}$ and substituting into \cref{eq:msbbe_min} yields our desired result:
\begin{align}
     \phi^\star_\omega&= \argmin_{\phi\in\Phi}\lVert\mathcal{B}[\hat{Q}_\omega]-\hat{B}_\phi \rVert_{\rho,\pi}^2.
\end{align}
\subsection{MSBBE Gradient Derivation}
\label{app:msbbe_grad}
We now derive an analytic form for the derivative of the MSBBE in \cref{eq:msbbe_grad}. Starting from the definition of the MSBBE:
\begin{align}
\nabla_\omega\msbbe&=\nabla_\omega\left\lVert \Qwdot - \mathcal{B}^\star_{\omega, N}\right\rVert_{\rho,\pi}^2,\\
&=\frac{1}{2}\nabla_\omega\mathbb{E}_{\rho,\pi}\left[(\Qwdot - \mathcal{B}^\star_{\omega, N})^2\right],\\
&=\mathbb{E}_{\rho,\pi}\left[(\Qwdot - \mathcal{B}^\star_{\omega, N})(\nabla_\omega\Qwdot-\nabla_\omega\mathcal{B}^\star_{\omega, N})\right],
\end{align}
Substituting for the definition of the Bayesian Bellman operator from \cref{eq:Bayesian_Bellman_Operator} yields our desired result:
\begin{align}
\nabla_\omega\msbbe&=\mathbb{E}_{\rho,\pi}\left[\left(\Qwdot - \mathbb{E}_{P(\phi\vert \Dwn)} \left[\hat{B}_\phi\right]\right)\left(\nabla_\omega\Qwdot-\mathbb{E}_{P(\phi\vert \Dwn)} \left[\hat{B}_\phi\right]\right)\right],\\
&=\mathbb{E}_{\rho,\pi}\left[\left(\Qwdot - \mathbb{E}_{P(\phi\vert \Dwn)} \left[\hat{B}_\phi\right]\right)\left(\nabla_\omega\Qwdot-\mathbb{E}_{P(\phi\vert \Dwn)} \left[\hat{B}_\phi\nabla_\omega\log p(\phi\vert \Dwn) \right]\right)\right].
\end{align}

%% file: Appendix/app_linear_bbo.tex
\section{Linear BBO}
\label{app:linear_bbo}
We now consider a simple Gaussian linear regression model as a case study where the Bellman operator and Q-function approximators take the forms $\Bp=v(s,a)^\top\phi $ and $\Qw=v(s,a)^\top\omega $. $v(s,a)$ is an $n$-dimensional feature vector and $\phi$ and $\omega$ are $n$-dimensional parameter vectors. We first derive the MSBBE and its derivative, showing that an analytic solution exists. For the sake of analysis we use a Gaussian conjugate prior over model parameters $\phi$:
\begin{align}
    p(\phi\vert \phi_0, \Sigma_0)=\mathcal{N}(\phi_0,\Sigma_0).
\end{align}
For mathematical convenience, we form an $N\times n$ matrix of features $V_N\coloneqq (v_1\cdots v_N)^\top$ from our data where $v_i\coloneqq v(s_i,a_i)$ and an $N$-dimensional vector of datapoints $\beta_\omega^N\coloneqq \{b_1,b_2,...b_N\}$. Our Bayesian Gaussian linear regression model has been well studied (see \citet{Murphy12}) and the posterior can be shown to be a Gaussian:
\begin{align}
    p(\phi \vert \Dwn)=\mathcal{N}(\phi_\omega^N,\Sigma_N),
\end{align}
where
\begin{align}
    \Sigma_N&=\left(\Sigma_0^{-1}+\frac{1}{\sigma^2}V_N^\top V_N\right)^{-1},\label{eq:linear_covariance_matrix}\\
    \phi_\omega^N&=\Sigma_N\left( \Sigma_0^{-1}\phi_0+\frac{1}{\sigma^2}V_N^\top\beta_\omega^N\right). \label{eq:linear_mean_vector}
\end{align}
We can also derive the posterior predictive:
\begin{align}
p(b \vert s,a, \Dwn)= \mathcal{N}(v(s,a)^\top \phi_\omega^N,\sigma^2+v(s,a)^\top \Sigma_N v(s,a)).
\end{align}
 By definition, the predictive mean is the Bayesian Bellman operator  $\mathcal{B}^\star_{\omega, N}(s,a)=v(s,a)^\top \phi_\omega^N$ from which we derive the MSBBE for linear BBO:
\begin{align}
    \msbbe&\coloneqq\left\lVert \Qwdot - \mathcal{B}^\star_{\omega, N}\right\rVert_{\rho,\pi}^2,\\
    &=\left\lVert  v(s,a)^\top\omega- v(s,a)^\top \phi_\omega^N\right\rVert_{\rho,\pi}^2,\\
    &=\left\lVert  v(s,a)^\top(\omega-  \phi_\omega^N)\right\rVert_{\rho,\pi}^2,\\
    &=\frac{1}{2}(\omega-  \phi_\omega^N)^\top\mathbb{E}_{\rho,\pi}\left[  v(s,a)v(s,a)^\top\right](\omega-  \phi_\omega^N).\label{eq:linear_msbbe}
\end{align}
The variance of the posterior predictive is composed of two terms: the first term, $\sigma^2$, is the aleatoric uncertainty of the data due to observation noise; the second term, $v(s,a)^\top\Sigma_N v(s,a)$, characterises the epistemic uncertainty, and grows whenever the test point $v(s,a)$ is far from observed data. It is this epistemic uncertainty we are concerned with for exploration as it provides a measure of how well explored regions of the state-action space are. Alongside the predictive mean, an agent can gauge how worthwhile it is to explore a region of high epistemic uncertainty depending upon its potential for high returns. Areas of low epistemic uncertainty will be `crossed-off' by the agent and are likely not to be explored again, only exploited if their predictive returns are high enough.  

We now take derivatives of \cref{eq:linear_msbbe} directly to derive the MSBBE gradient required by our algorithms:
\begin{align}
    \nabla_\omega\msbbe&=(\nabla_\omega\omega-  \nabla_\omega\phi_\omega^N)^\top\mathbb{E}_{\rho,\pi}\left[  v(s,a)v(s,a)^\top\right](\omega-  \phi_\omega^N),\\
    &=(I-  \nabla_\omega(\phi_\omega^N)^\top)\mathbb{E}_{\rho,\pi}\left[  v(s,a)v(s,a)^\top\right](\omega-  \phi_\omega^N),\label{eq:grad_msbbe_part}
\end{align}
To proceed, we must find a expression for the derivative of $(\phi_\omega^N)^\top$: 
\begin{align}
    \nabla_\omega (\phi_\omega^N)^\top&=\nabla_\omega (\phi_\omega^N)^\top,\\
    &=\nabla_\omega \left(\Sigma_N\left( \Sigma_0^{-1}\phi_0+\frac{1}{\sigma^2}V_N^\top\beta_\omega^N\right)\right)^\top ,\\
    &=\nabla_\omega\frac{1}{\sigma^2}(\beta_\omega^N)^\top V_N\Sigma_N^\top .
\end{align}
Now, for each $b_i$ forming the vector $\beta_\omega^N$, we have $\nabla_\omega b_i=\nabla_\omega(r+\gamma \omega^\top v'_i)=\gamma v'_i$. The derivative of the matrix-vector product $(\beta_\omega^N)^\top V_N$ can thus be found as:
\begin{align}
    \nabla_\omega (\beta_\omega^N)^\top V_N= \gamma\sum_{i=1}^N v'_i v_i^\top.
\end{align}
We substitute $\nabla_\omega (\beta_\omega^N)^\top V_N= \gamma\sum_{i=1}^N v'_i v_i^\top$ into \cref{eq:grad_msbbe_part} to obtain the gradient:
\begin{align}
\nabla_\omega\msbbe&
=\left(I-\frac{\gamma}{\sigma^2}\sum_{i=1}^N {v'_i}v_i^\top\right)\Sigma_N^\top\mathbb{E}_{\rho,\pi}\left[ v(s,a) v(s,a)^\top\right](\omega-  \phi_\omega^N). \label{eq:msbbe_grad_linear_exact}
\end{align}

\subsection{Deriving the LSTD algorithm}
Although we have derived the exact gradient of the MSBBE, a simple approach for inferring the posterior predictive mean is to solve the equation $\omega= \phi_\omega^N$: indeed, as a sanity check, we see from \cref{eq:msbbe_grad_linear_exact} that any $\omega^\star$ such that $\omega^\star=\phi_{\omega^\star}^N$ trivially results in an MSBBE gradient of 0 and parametrises a global minimiser. Expanding our equation using the definition of $\phi_\omega^N$ from \cref{eq:linear_mean_vector}, we obtain:
\begin{align}
    \omega^\star&= \phi_{\omega^\star}^N,\\
    &=\Sigma_N\left( \Sigma_0^{-1}\phi_0+\frac{1}{\sigma^2}V_N^\top\beta_{\omega^\star}^N\right),\\
    &=\Sigma_N\Sigma_0^{-1}\phi_0+\frac{1}{\sigma^2}\Sigma_NV_N^\top\beta_{\omega^\star}^N,\\
    &=\Sigma_N\Sigma_0^{-1}\phi_0+\frac{1}{\sigma^2}\Sigma_N\sum_{i=1}^N v_i b_i.\label{eq:linear_equation}
\end{align}
Introducing the shorthand $v_i'\coloneqq v'_i$, we substitute for the definition of the empirical Bellman equation from \cref{eq:bellman_approx} using the linear function approximator, $b_i=r_i+\gamma {v'_i}^\top \omega$. This allows us to factorise \cref{eq:linear_equation}:
\begin{align}
    \omega^\star=&\Sigma_N\Sigma_0^{-1}\phi_0+\frac{1}{\sigma^2}\Sigma_N\sum_{i=1}^N v_i (r_i+\gamma {v'_i}^\top \omega^\star),\\
    \Sigma_N^{-1}\omega^\star=&\Sigma_0^{-1}\phi_0+\frac{1}{\sigma^2}\sum_{i=1}^N v_i r_i+\frac{\gamma}{\sigma^2} \left(\sum_{i=1}^N v_i {v'_i}^\top\right) \omega^\star,\\
    \Bigg(\Sigma_N^{-1}-&\frac{\gamma}{\sigma^2} \sum_{i=1}^N v_i {v'_i}^\top\Bigg)\omega^\star=\Sigma_0^{-1}\phi_0+\frac{1}{\sigma^2}\sum_{i=1}^N v_i r_i,
\end{align}
Substituting for the definition of the predictive covariance from \cref{eq:linear_covariance_matrix}:
\begin{align}
     \Sigma_N^{-1}&=\Sigma_0^{-1}+\frac{1}{\sigma^2}V_N^\top V_N=\Sigma_0^{-1}+\frac{1}{\sigma^2} \sum_{i=1}^N v_i v_i^\top,
\end{align}
yields:
\begin{align}
        \Bigg(\Sigma_0^{-1}+&\frac{1}{\sigma^2}\sum_{i=1}^N v_i\left(v_i- \gamma {v'_i}\right)^\top\Bigg)\omega^\star=\Sigma_0^{-1}\phi_0+\frac{1}{\sigma^2}\sum_{i=1}^N v_i r_i,
\end{align}
Define the matrix:
\begin{align}
    D_N\coloneqq\Sigma_0^{-1}+\frac{1}{\sigma^2}\sum_{i=1}^N v_i\left(v_i- \gamma {v'_i}\right)^\top,
\end{align}\begin{wrapfigure}{r}{0.5\textwidth}
\begin{minipage}{0.5\textwidth}
\vspace{-0.3cm}
\begin{algorithm}[H]
    \caption{Calculate $\omega^\star$}
	\label{alg:exact_omega}
	\begin{algorithmic}
	    \FOR{$i\in\{1,...N\}$} 
		\IF{$i==1$}
		\STATE ${D}^{-1}\leftarrow\left( \sigma^2\Sigma_0^{-1}\right)^{-1}$
		\STATE $\chi\leftarrow\sigma^2\Sigma_0^{-1}\phi_0$
		\ENDIF
		\STATE $\Delta\leftarrow w_i (v_i-\gamma v_i')$
		\STATE${D}^{-1}\leftarrow{D}^{-1}-\frac{{D}^{-1}v_i\Delta^\top{D}^{-1}}{1+\Delta^\top{D}^{-1} v_i}$
		\STATE $\chi\leftarrow \chi + w_i v_i r_i$
		\ENDFOR
		\STATE $\omega^\star \leftarrow D^{-1}\chi $
		\RETURN $\omega^\star$
	\end{algorithmic}
\end{algorithm}
\vspace{-0.7cm}
\end{minipage}
\end{wrapfigure}
and the vector:
\begin{align}
    \chi_N\coloneqq \Sigma_0^{-1}\phi_0+\frac{1}{\sigma^2}\sum_{i=1}^N v_i r_i
\end{align}
with which we obtain a solution for $\omega^\star$:
\begin{align}
    \omega^\star=D_N^{-1}\chi_N 
\end{align}

\label{app:algoritmic_details}

If the problem is tractable enough to store $D_N^{-1}$, then the exact solution $\omega^\star$ can be obtained using Sherman-Morrison updates as outlined in \cref{alg:exact_omega}. We use an importance weight $w_i\coloneqq \frac{\pi(a_i\vert s_i)}{\pi_e(a_i\vert s_i)}$ for the off-policy case when an exploratory policy $\pi_e$ that is difference from the evaluation policy $\pi$ is used to gather data. When the on-policy sampling is used, $w_i=1$. If the prior takes a form that can be inverted with complexity $\mathcal{O}(n^2)$ or less, the overall complexity of \cref{alg:exact_omega} is $\mathcal{O}(Nn^2)$. 
\begin{wrapfigure}{r}{0.5\textwidth}
\begin{minipage}{0.5\textwidth}
\vspace{-0.7cm}
\begin{algorithm}[H]
    \caption{Calculate $\omega^\star$ with Frequentist prior}
	\label{alg:exact_omega_uniformative }
	\begin{algorithmic}
	    \FOR{$i\in\{1,... N\}$}
		\IF{i==0}
		\STATE ${D}^{-1}\leftarrow\epsilon I$
		\STATE $\chi\leftarrow 0$
		\ENDIF
		\STATE $\Delta\leftarrow w_i(v_i-\gamma v_i')$
		\STATE${D}^{-1}\leftarrow{D}^{-1}-\frac{{D}^{-1}v_i\Delta^\top{D}^{-1}}{1+\Delta^\top{D}^{-1} v_i}$
		\STATE $\chi\leftarrow \chi + w_iv_i r_i$
		\IF{$i \le n$}
		\STATE$D^{-1}\leftarrow D^{-1}\frac{\epsilon D^{-1}_i}{1+\epsilon\left[D^{-1}\right]_{i,i}}$
		\ENDIF
		\ENDFOR
		\STATE $\omega^\star \leftarrow -D^{-1}\chi $
		\RETURN $\omega^\star$
	\end{algorithmic}
\end{algorithm}
\vspace{-1.5cm}
\end{minipage}
\end{wrapfigure}

To obtain a frequentist equivalent of our algorithm, we use an uninformative prior which can be obtained by choosing $\phi_0=0$ and $\Sigma_0=\frac{1}{\sigma_0^2}I$ with $\sigma_0^2\rightarrow 0$. To obtain an algorithm with complexity $\mathcal{O}(Nn^2)$ that uses this prior, we first choose $({\sigma^2\Sigma_0})^{-1}=\epsilon I$ for some $\epsilon>>1$. For each datapoint $i\le n$, we can then add the outer product $\epsilon 1_i 1_i^\top $ using the Sherman-Morrison formula, where $1_i$ is a vector of zeros except for the $i$th entry which is a $1$. After $ n$ datapoints, we have then 'removed' the original prior, obtaining exactly $\omega^\star$ inferred using an uniformative prior. We outline this procedure in \cref{alg:exact_omega_uniformative } where $D^{-1}_i\coloneqq D^{-1}1_i 1_i^\top D^{-1} $ is the outer product of the $i$th column and row of $D^{-1}$ and $[D^{-1}]_{i,i}$ is the $i$th diagonal element of $D^{-1}$.
\subsection{Recovering TDC/GTD2}
\label{app:tdc_gtd2}
To derive TDC/GTD2 from our Gaussian linear regression model, we observe that the Bayesian Bellman operator can be interpreted as an $N$-sample Monte-Carlo estimate of the  projection operator with additional bias due to the prior:
\begin{align}
    \mathcal{B}^\star_{\omega, N}&=v^\top \left(\sigma^2\Sigma_0^{-1}+V_N^\top V_N\right)^{-1}\left(\sigma^2\Sigma_0^{-1}\phi_0+V_N^\top\beta_\omega^N\right),\\
    &=v^\top \left(\frac{1}{N}\left(\sigma^2\Sigma_0^{-1}+\sum_{i=1}^N v_i v_i^\top\right) \right)^{-1}\frac{1}{N}\left(\sigma^2\Sigma_0^{-1}\phi_0+\sum_{i=1}^N v_ib_i\right)\label{eq:n-sample-linear}.
\end{align}
As the TDC/GTD2 is a frequentist algorithm, we must derive the Bayesian Bellman operator in the limit $N\rightarrow\infty$. Taking the limit $N\rightarrow\infty$ of \cref{eq:n-sample-linear} using the strong law of large numbers, we see that the effect of the prior will diminish (observe that it does not scale with increasing $N$) and the Bayesian Bellman operator converges to the projection operator:
\begin{align}
    \lim_{N\rightarrow\infty} \mathcal{B}^\star_{\omega, N}&=\lim_{N\rightarrow\infty} v^\top \left(\frac{1}{N}\left(\sigma^2\Sigma_0^{-1}+\sum_{i=1}^N v_i v_i^\top\right) \right)^{-1}\frac{1}{N}\left(\sigma^2\Sigma_0^{-1}\phi_0+\sum_{i=1}^N v_ib_i\right),\\
    &=\lim_{N\rightarrow\infty} v^\top \left(\frac{1}{N}\sum_{i=1}^N v_i v_i^\top \right)^{-1}\frac{1}{N}\sum_{i=1}^N v_ib_i,\\
    &=v^\top\mathbb{E}_{\rho,\pi} \left[vv^\top\right]^{-1}\mathbb{E}_{\rho,\pi} \left[v\Bqwdot\right],\\
    &=\mathcal{P}_{\hat{B}_\phi}\circ\mathcal{B}[\hat{Q}_\omega].
\end{align}
This confirms the consistency results that we established between the mean squared projected Bellman error (MSPBE) and MSBBE in the limit $N\rightarrow\infty$ under \cref{proof:consistency_app} in \cref{sec:Gaussian_bbo} for Gaussian models. The MSPBE has been well studied for linear function approximators \citep{Sutton09a,Sutton09b,Maei09} and can be shown to take the form:
\begin{align}
   &\textrm{MSPBE}(\omega)= \mathbb{E}_{\rho,\pi}\left[v^\top (\Bqwdot-\Qwdot)\right]\mathbb{E}_{\rho,\pi} \left[vv^\top\right]^{-1}\mathbb{E}_{\rho,\pi} \left[v(\Bqwdot-\Qwdot)\right].\label{eq:mspbe}
\end{align}
Taking gradients to minimise the MSPBE via stochastic gradient descent leads to the TDC/GTD2 algorithms, which are derived in full in \citet{Sutton09a,Sutton09b} from the above objective. To avoid the costly matrix inversion in \cref{eq:mspbe}, a set of weights to approximate the term $\zeta\approx\mathbb{E}_{\rho,\pi}\left[vv^\top\right]^{-1}\mathbb{E}_{\rho,\pi}\left[v(\Bqwdot-\Qwdot)\right]$ is learnt, which are updated on a slower timescale:
\begin{align}
    \zeta_{k+1}\leftarrow \zeta_k+\alpha_k^\zeta (r+\gamma\hat{Q}_{\omega_k}'-\hat{Q}_{\omega_k} -v_k^\top\zeta_k)v_k.
\end{align}
This approximation is then used to update the function approximator. When a TD estimate is used, this results in two algorithms :
\begin{gather}
    \omega_{k+1}\leftarrow \omega_k+\alpha_k^\omega \left(v_k-\gamma v'_k \right)v_k^\top\zeta_k\quad \textrm{(GTD2)},\\
    \omega_{k+1}\leftarrow \omega_k+\alpha_k^\omega v_k(r+\gamma\hat{Q}_{\omega_k}'-\hat{Q}_{\omega_k})-\alpha_k^\omega\gamma v'_k v_k^\top\zeta_k\quad \textrm{(TDC)}.
\end{gather}
As the TDC/GTD2 algorithm is derived from a frequentist perspective, it does not characterise uncertainty in the MDP. Conversely, our framework allows us to estimate the predictive variance due to state visitation after $N$ updates of GDT2/TDC, which is essential for achieving deep exploration.

%% file: Appendix/app_randomised_priors.tex
\section{Randomised Priors for BBO}
\label{app:randomised_priors}
Randomised priors is a method for obtaining an approximation of an intractable posterior by `randomising' the maximum a posteriori (MAP) point estimate. Here, a noise variable $\epsilon\in\mathcal{E}$ with distribution $P_E(\epsilon)$ where the density $p_E(\epsilon)$ has the same form as the prior is introduced, which defines a continuum of $\epsilon$-randomised MAP estimates:
\begin{align}
   \psi^\star(\epsilon;\Dwn)= \argmax_{\phi\in\Phi}\mathcal{L}(\phi;\Dwn,\epsilon),\quad \mathcal{L}(\phi;\Dwn,\epsilon) \coloneqq\frac{1}{N}\left(R(\phi-\epsilon)-\sum_{i=1}^N\log p(b_i\vert s_i,a_i,\phi)\right).
\end{align}
For simplicity, it is implicitly assumed when using RP that the prior-randomised MAP estimate $\psi^\star(\epsilon;\Dwn)$ is $P_E$-integrable and that each $\argmax_{\phi\in\Phi}\mathcal{L}(\phi;\Dwn,\epsilon)$ is a singleton. We also assumed in \cref{ass_app:RP_approximators} that $R(\phi-\epsilon)$ is well-defined for any $\epsilon\in\mathcal{E}$ and $\phi\in\Phi$. The RP approximate posterior can then be constructed by averaging over all MAP estimates using $P_E$:
\begin{align}
   q(\phi\vert \Dwn) =\int_\mathcal{E} \delta(\phi=\psi^*(\epsilon;\Dwn)) dP_E(\epsilon).
\end{align}
The RP approximate posterior is motivated by the fact that sampling from $q(\phi\vert \Dwn)$ is equivalent to sampling from the true posterior when a linear Gaussian model is used \citep{Osband18,Osband19}. When a nonlinear model is used, we still expect the RP posterior to provide a good approximation to the true posterior. \citet{Pearce19} confirm that this holds where nonlinear neural networks are used as function approximations. 

A na{\"i}ve way to sample from the approximate posterior is to first sample $\epsilon\sim P_E$ and solve the optimisation problem $\psi^\star(\epsilon;\Dwn)= \argmax_{\phi\in\Phi}\mathcal{L}(\phi;\Dwn,\epsilon)$. This approach is not tractable as solving the optimisation problem is NP-hard, so instead an ensembling approach is used.  As outlined in \cref{sec:approximate_bbo}, $L$ prior randomisations $\mathcal{E}_L\coloneqq\{\epsilon_l\}_{l=1:L}$ are drawn from $P_E$. For each $l\in\{1:L\}$, a set of solutions to the prior-randomised MAP objective are found: 
\begin{align} \psi_l^\star(\omega_l) \in \argmin_{\phi\in \Phi}\mathcal{L}(\phi;\mathcal{D}_{\omega_l}^N,\epsilon_l)\coloneqq\argmin_{\phi\in \Phi}\frac{1}{N}\left( {R}(\phi-\epsilon_l)-\sum_{i=1}^N\log p(b_i\vert s_i,a_i,\phi)\right).
\end{align}

When using RP with ensembling for reinforcement learning, we write the $Q$-function approximator as an ensemble of $L$ parameters $\Omega_L\coloneqq \{\omega_l\}_{l=1:L}$ where $\hat{Q}_\omega=\frac{1}{L}\sum_{l=1}^L \hat{Q}_{\omega_l}$ and treat learning each $\omega_l$ as a separate problem for each ensemble, leading to the randomised priors ensembled MSBBE:
\begin{align}
\textrm{MSBBE}_\textrm{RP}(\omega_l)\coloneqq \lVert \hat{Q}_{\omega_l}- \hat{B}_{\psi_l^\star(\omega_l)}\rVert_{\rho,\pi}^2
\end{align}

%% file: Appendix/app_algorithms.tex
\section{BBAC Algorithm}
\label{app:bbac_algorithm}
For BBAC, we use the Gaussian BBO model introduced in \cref{sec:Gaussian_bbo} with a Gaussian prior: $R(\phi)=\frac{1}{\sigma_0^2}\lVert\phi-\phi_0\rVert_2^2$. The $\log$-posterior is thus:
\begin{align}
    -\log p(\phi \vert \Dwn)=c_\textrm{norm}+\sum_{i=1}^N \frac{(b_i-\hat{B}_\phi(s_i,a_i))^2}{2\sigma^2} +\frac{1}{\sigma_0^2}\lVert\phi-\phi_0\rVert_2^2
\end{align}
We choose a prior parameterisation of $\phi_0=0$, which implies a corresponding Gaussian noise distribution: $P_E=\mathcal{N}(0,\sigma^2_0I)$ \citep{Osband18}. The RP (critic) objective can be derived from the $\log$-posterior as:
\begin{align}
    \mathcal{L}(\psi_l;\mathcal{D}_{\omega_l}^N,\epsilon_l)=\sum_{i=1}^N \frac{(b_i-\hat{B}_\phi(s_i,a_i))^2}{2\sigma^2} +\frac{1}{\sigma_0^2}\lVert\psi_l-\epsilon_l\rVert_2^2,\label{eq:Gaussian_RP_objective}
\end{align}
where we have ignored $\tfrac{1}{N}$ as $N$ is finite and so does not contribute to the objective's solution. 
The two-timescale updates in~(\ref{eq:rp_updates_fast}) and~(\ref{eq:rp_updates_slow}) are thus:
\begin{gather}
\psi_l\leftarrow\mathcal{P}_\Omega \left( \psi_l-\alpha_k  \nabla_{\psi_l}\hat{\mathcal{L}}_\textrm{BBAC}^i(\psi_l) \right),\quad \textrm{(critic)} \label{eq:bbac_update_fast}\\
\omega_l\leftarrow\mathcal{P}_\Omega( \omega_l-\beta_k (\omega_l-\psi_l)),\quad \textrm{(target critic)\label{eq:bbac_updates_slow}}
\end{gather}
where:
\begin{align}
    &\hat{\mathcal{L}}_\textrm{BBAC}^i(\psi_l)\coloneqq\frac{1}{2\sigma^2}\left(r_i+\gamma \hat{Q}_{\omega_l}(s'_i,a'_i)- \hat{B}_{\psi_l}(s_i,a_i)\right)^2  +\frac{1}{\sigma_0^2}\lVert\psi_l-\epsilon_l\rVert_2^2.
\end{align}

For our ensembled actors, we choose a Gaussian policy: $\pi_{\theta_l}=\mathcal{N}(\mu_{\theta_l},\Sigma_{\theta_l})$. Using the reparametrisation trick for actor-critic \citep{kingma2014,Heess15,fellows18}, we can  derive low variance policy gradient updates by introducing a variable $\upsilon\sim P_\upsilon(\cdot)$ where $P_\upsilon=\mathcal{N}(0,I)$. Defining the transformation of variables $t^{-1}_{\theta_l}(\upsilon,s)=\Sigma^{\frac{1}{2}}_{\theta_l}(s)(\mu_{\theta_l}(s)-\upsilon)$, we can write our actor objective as an expectation under $P_\upsilon$: 
\begin{align}
    \mathbb{E}_{\rho(s)\pi_{\theta_l}(a\vert s)}[B_{\phi_l}(s,a)]=\mathbb{E}_{\rho(s)P_\upsilon(\upsilon)}[B_{\phi_l}(s,a=t^{-1}_{\theta_l}(\upsilon,s))].\label{eq:bbac_actor_objective}
\end{align}

\begin{minipage}{\textwidth}
\vspace{-0.25cm}
\begin{algorithm}[H]
    \caption{$\textsc{UpdatePosterior}(\Theta_L, \Omega_L, \Psi_L,\mathcal{E}_L, \mathcal{H})$}
    \label{alg:update_posterior}
    \begin{algorithmic}
        \STATE $k\leftarrow 0$
        \WHILE{not converged}
    	\STATE Sample mini-batch of transitions $T\sim \mathcal{D}$
    	\FOR{$l\in\{1,...L\}$} 
    	\FOR{$\{s_i,a_i,r_i,s_i'\}\in T$}
    	\STATE $a'_i\sim \pi_{\theta_l}(\cdot\vert s'_i)$
        \STATE $\psi_l\leftarrow\mathcal{P}_\Omega \left( \psi_l-\frac{\alpha_k}{\lvert T\rvert}  \nabla_{\psi_l}\hat{\mathcal{L}}_\textrm{BBAC}^i(\psi_l) \right)$
        \ENDFOR
        \STATE $\omega_l\leftarrow\mathcal{P}_\Omega( \omega_l-\beta_k (\omega_l-\psi_l)) $
        \STATE Sample batch $B_l$ of states from $s\sim d(\cdot)$
        \FOR{$s\in B_l$}
        \STATE $\upsilon\sim P_\upsilon(\cdot)$
        \STATE $\theta_l \leftarrow \theta_l  + \frac{\zeta_k}{\lvert B_l\rvert} \nabla_{\theta_l}B_{\phi_l}(s,a=t^{-1}_{\theta_l}(\upsilon,s)) $
        \ENDFOR
        \ENDFOR
        \STATE $k \leftarrow k+1$
        \ENDWHILE
    \end{algorithmic}
\end{algorithm}
\end{minipage}

To minimising the objective in \cref{eq:bbac_actor_objective}, we propose a stochastic gradient descent update:
\begin{align}
    \theta_l \leftarrow \theta_l  +\zeta_k  \nabla_{\theta_l}B_{\phi_l}(s,a=t^{-1}_{\theta_l}(\upsilon,s)),\quad \textrm{(actor)}
\end{align}
where $\zeta_k $ is the actor learning rate. The  pseudocode for updating the posterior using these objectives is shown in \cref{alg:update_posterior} using batch updates.

Proving full convergence of our actor-critic is beyond the scope of this paper, however we leverage insights from our two-timescale analysis in \cref{sec:approximate_inference} and ensure that $\alpha_k>\beta_k>\zeta_k$ to stabilise learning. Finally, we include the pseudocode used to learn our Gaussian behavioural policy in \cref{alg:update_policy}. Like in soft actor-critic, the algorithm randomly samples two critics from the ensemble and updates the actor using an entropy regularised objective with the minimum of the samples. 
\begin{algorithm}[H]
    \caption{$\textsc{UpdateBehaviouralPolicy}(\theta^\dagger, \Psi_L, \mathcal{H})$}
    \label{alg:update_policy}
    \begin{algorithmic}
        \STATE $k\leftarrow 0$
        \WHILE{not converged}
    	\STATE Sample mini-batch of states $B\sim \mathcal{H}$
    	\STATE Sample $\psi_1,\psi_2\sim \textrm{Unif}(\Psi_L)$
    	\FOR{$s\in B:$}
        \STATE $\upsilon\sim p_\upsilon(\cdot)$
        \STATE $\theta^\dagger \leftarrow \theta^\dagger + \frac{\alpha_k}{\lvert B\rvert} \nabla_{\theta^\dagger}\hat{J}(\theta^\dagger;s,\upsilon,\psi_1,\psi_2)$ 
        \ENDFOR
        \STATE $k \leftarrow k+1$
        \ENDWHILE
    \end{algorithmic}
\end{algorithm}
where 
\begin{align}
    \hat{J}(\theta^\dagger;s,\upsilon,\psi_1,\psi_2)\coloneqq&\min_{i\in\{1,2\}}\left(\hat{B}_{\psi_i}(s,a=t^{-1}_{\theta^\dagger}(\upsilon,s))\right)-\alpha \log \pi_{\theta^\dagger}(a=t^{-1}_{{\theta^\dagger}}(\upsilon,s)\vert s).
\end{align}

%% file: Appendix/app_experiments.tex
\section{Policy Evaluation Experiments}
\label{app:policy-evaluation-experiments}
In this section, we empirically study the properties of BBO in the policy evaluation regime. We start by presenting linear BBO, showing that it performs on-par with well-studied existing linear policy evaluation algorithms. We then consider nonlinear regime, first in a simplistic counter example, which shows the convergence of nonlinear BBO in an easily-understandable practical setting, and further in more complex domains with neural network function approximators, which demonstrate convergence and consistency of BBO.

\subsection{Linear Policy Evaluation}
\label{app:experimental_details_linear_policy_evaluation}

We first evaluate the linear BBO from \cref{alg:exact_omega} in a suite of 9 policy evaluation tasks, comparing it with 6 other methods: TD, TDC, GTD2, BRM, and two LSTD variants, which we refer to as LSTD and LSTD+. LSTD corresponds to vanilla version of the algorithm and thus can be seen as the non-regularized frequentist version of BBO, whereas LSTD+ corresponds to LSTD with additional improvements, such as improved off-policy reweighting, eligibility traces, and regularization, as presented in \citet{dann2014policy} (Section 3.4). We note that LSTD+ is not directly comparable with BBO, but is rather presented here as a strong baseline. All the environments, evaluated policies, and other training configurations follow exactly those presented by \citet{dann2014policy} (Section 3.1). We omit three tasks included in \citet{dann2014policy}: the Baird counterexample due to its solution corresponding to 0 weights and thus being unfairly trivial for least-squares methods like BBO and LSTD to solve, and the on- and off-policy Cart-Pole Swingup tasks due to our lack of access to the software required to run the policies.

\subsubsection{Hyperparameters}

We use the hyperparameters provided by \cite{dann2014policy} (Section 3.1.7) for all the baseline algorithms (LSTD, TD, TDC, GTD2, BRM). For BBO, we test the following values for the prior variance: ${\{1\mathrm{e}{-1}, 3\mathrm{e}{-1}, 1\mathrm{e}{0}, 3\mathrm{e}{0}, 1\mathrm{e}{1}\}}$. In the results reported, we use $1\mathrm{e}{-1}$ for Cart-Pole with impoverished features and  $1\mathrm{e}{1}$ for the rest of the tasks.


\subsubsection{Results}

\begin{figure*}[h]
    \centering

    \includegraphics[width=\linewidth, trim={2mm 2mm 2mm 1mm}, clip]{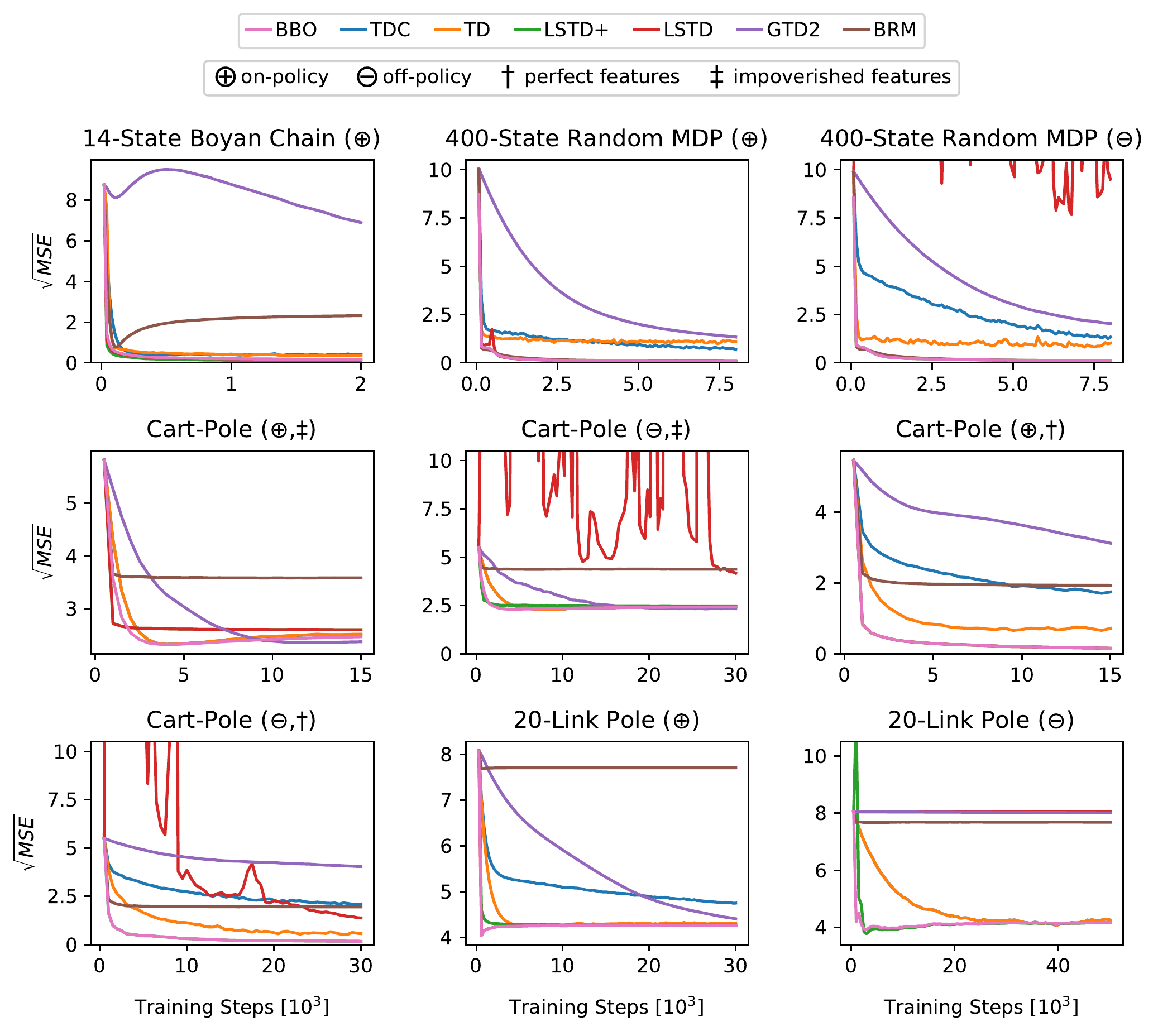}
    \caption{Linear policy evaluation results.}
    \label{fig:linear_policy_evaluation_result_full}
\end{figure*}

The full training curves are presented in \cref{fig:linear_policy_evaluation_result_full} and summarized in tables \cref{tab:linear_policy_evaluation_final_MSE} and \cref{tab:linear_policy_evaluation_sum_RMSE}. \cref{tab:linear_policy_evaluation_final_MSE} presents the final mean squared error for each algorithm in each task. BBO achieves top final mean squared error in 6 out of 9 tasks, and performs close to the best method in all the rest. In terms of cumulative mean squared error, as shown in \cref{tab:linear_policy_evaluation_sum_RMSE}, BBO outperforms other methods in 4 out of 9 tasks, again performing similarly to top-performing methods across all tasks. This demonstrates the benefits of Bayesian methods and incorporating priors in policy evaluation models.

\begin{table*}[t]
\centering
\caption{MSE of final predictions. The values for all methods except for BBO are obtained with code provided by~\cite{dann2014policy}. $\oplus$=on-policy, $\ominus$=off-policy, $\dag$=perfect features, $\ddag$=impoverished features.}
\label{tab:linear_policy_evaluation_final_MSE}
\begin{tabular}{llllllll}
\toprule
{} &            BBO &           GTD2 &    TD &   TDC &          LSTD+ &           LSTD &            BRM \\
\midrule
14-State Boyan Chain ($\oplus$)  &           0.16 &           6.89 &  0.36 &  0.40 &  \textbf{0.10} &           0.16 &           2.32 \\
400-State Random MDP ($\oplus$)  &  \textbf{0.07} &           1.34 &  1.09 &  0.69 &  \textbf{0.07} &  \textbf{0.07} &           0.08 \\
400-State Random MDP ($\ominus$) &  \textbf{0.11} &           2.03 &  1.02 &  1.33 &  \textbf{0.11} &           9.50 &  \textbf{0.11} \\
Cart-Pole ($\oplus$,$\ddag$)     &           2.46 &  \textbf{2.37} &  2.51 &  2.51 &           2.60 &           2.60 &           3.58 \\
Cart-Pole ($\ominus$,$\ddag$)    &           2.42 &  \textbf{2.33} &  2.44 &  2.44 &           2.47 &           4.17 &           4.37 \\
Cart-Pole ($\oplus$, $\dag$)     &  \textbf{0.15} &           3.13 &  0.72 &  1.75 &  \textbf{0.15} &  \textbf{0.15} &           1.93 \\
Cart-Pole ($\ominus$, $\dag$)    &  \textbf{0.17} &           4.04 &  0.57 &  2.10 &  \textbf{0.17} &           1.38 &           1.95 \\
20-Link Pole ($\oplus$)          &  \textbf{4.26} &           4.41 &  4.31 &  4.75 &           4.27 &  \textbf{4.26} &           7.71 \\
20-Link Pole ($\ominus$)         &  \textbf{4.17} &           8.01 &  4.25 &  4.25 &  \textbf{4.17} &           8.04 &           7.68 \\
\bottomrule
\end{tabular}
\end{table*}

\begin{table*}[t]
\centering
\caption{Sum of square root MSE over all timesteps. The values for all methods except for BBO are obtained with code provided by~\cite{dann2014policy}. $\oplus$=on-policy, $\ominus$=off-policy, $\dag$=perfect features, $\ddag$=impoverished features.}
\label{tab:linear_policy_evaluation_sum_RMSE}
\begin{tabular}{llllllll}
\toprule
{} &              BBO &    GTD2 &      TD &     TDC &           LSTD+ &             LSTD &     BRM \\
\midrule
14-State Boyan Chain ($\oplus$)  &            33.46 &  841.45 &   61.33 &   61.55 &  \textbf{25.06} &            32.21 &  214.35 \\
400-State Random MDP ($\oplus$)  &   \textbf{24.74} &  342.56 &  122.51 &  119.84 &  \textbf{24.74} &            27.75 &   27.87 \\
400-State Random MDP ($\ominus$) &   \textbf{29.65} &  442.96 &  113.30 &  266.36 &  \textbf{29.65} &       $> 10^{3}$ &   32.13 \\
Cart-Pole ($\oplus$,$\ddag$)     &   \textbf{76.99} &   89.96 &   79.65 &   79.61 &           81.56 &            81.56 &  109.88 \\
Cart-Pole ($\ominus$,$\ddag$)    &  \textbf{243.86} &  291.02 &  253.96 &  253.88 &          253.53 &       $> 10^{3}$ &  438.94 \\
Cart-Pole ($\oplus$, $\dag$)     &   \textbf{13.51} &  116.98 &   31.85 &   68.75 &  \textbf{13.51} &   \textbf{13.51} &   62.78 \\
Cart-Pole ($\ominus$, $\dag$)    &            24.58 &  267.92 &   67.91 &  159.23 &  \textbf{24.57} &           681.19 &  122.12 \\
20-Link Pole ($\oplus$)          &           428.98 &  555.56 &  441.97 &  508.58 &          431.56 &  \textbf{428.97} &  770.76 \\
20-Link Pole ($\ominus$)         &  \textbf{415.37} &  802.78 &  470.12 &  470.20 &          421.74 &           804.45 &  768.59 \\
\bottomrule
\end{tabular}
\end{table*}

\subsection{Tsitsiklis' Triangle Counterexample}
\label{app:experimental_details_tsitsiklis_triangle_counter_example}

\begin{wrapfigure}{r}{0.2\textwidth}
\vspace{-15mm}
    \begin{center}
        \includegraphics[width=\linewidth, trim={0mm 0mm 0mm 0mm}, clip]{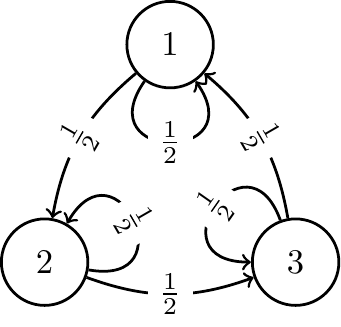}
    \vspace{-7mm}
    \caption{Tsitsiklis' \\ Triangle MDP~\cite{Tsitsiklis97}.}
    \label{fig:spiral-mdp}
    \end{center}
\vspace{-5mm}
\end{wrapfigure}

We then consider the three-state Tsitsiklis' Triangle MDP~\cite{Tsitsiklis97} designed to prove divergence of TD methods with nonlinear function approximators. The purpose of this experiment is to empirically validate the convergence properties of nonlinear BBO algorithms.

\subsubsection{Environment and Value Function}

The environment, illustrated in \cref{fig:spiral-mdp}, consists of the state space $S=\{1,2,3\}$ and the action-independent transition kernel

\begin{align}
    P\coloneqq [p(s'=j\vert \cdot, s=i)]_{i,j}=\begin{bmatrix}
\frac{1}{2} & 0 & \frac{1}{2}\\[6pt]
\frac{1}{2}& \frac{1}{2} & 0\\[6pt]
0& \frac{1}{2}& \frac{1}{2}
\end{bmatrix}.
\end{align}

Let ${\hat{V}(\omega)\coloneqq [ \hat{V}_\omega(s=1),\hat{V}_\omega(s=2),\hat{V}_\omega(s=3)]^\top }$ be the value function vector. It can be shown that any $\hat{V}(\omega)$ parametrised by $\omega\in\mathbb{R}$ and satisfying the linear dynamical system

\begin{align}
    \frac{d \hat{V}(\omega)}{d \omega}=\left(Q+\epsilon I\right)\hat{V}(\omega), \label{eq:linear_system}
\end{align}


with the condition that $\hat{V}(0)^\top 1=0$ where $\epsilon>0$ is a small constant and:

\begin{align}
    Q\coloneqq \begin{bmatrix}
1 & \frac{1}{2} & \frac{3}{2} \\[6pt]
\frac{3}{2} & 1 & \frac{1}{2}\\[6pt]
\frac{1}{2} & \frac{3}{2} & 1
\end{bmatrix},
\end{align}

diverges when updated using the TD algorithm \citep{Tsitsiklis97}. To be consistent with \citet{Maei09} we choose ${\hat{V}(\omega) =\exp(\epsilon\omega)(a\cos(\lambda\omega)-b\sin(\lambda\omega))}$ with ${a = (-14.9996, -35.0002, 50.0004)}$, ${b = (-49.0753, 37.5278, 11.5469)}$, ${\lambda = \sqrt{3}/2}$, and ${\epsilon = 10^{-2}}$
as a solution to \cref{eq:linear_system} (using the values that we received from the authors' implementation), we use the discount $\gamma=0.9$ and normalize the gradient steps to stabilize the updates. Each update batch includes all 6 environment transitions.

\subsubsection{Hyperparameters}
We search over the learning rates for all the algorithms. For TDC, and GTD2, we begin with a coarse grid search with values of ${\{ 1\mathrm{e}{-3}, 1\mathrm{e}{-2}, 1\mathrm{e}{-1}, 1\mathrm{e}{0} \}}$, followed by finer search of ${\{ 1\mathrm{e}{-1}, 2\mathrm{e}{-1}, ..., 9\mathrm{e}{-1}, 1\mathrm{e}{0} \}}$. For TD(0), we manually search around the learning rate used in \citet{Maei09}. We do not search over BBO's prior loss weight and set to 1.0. The final hyperparameters are presented in \cref{tab:tsitsiklis-triangle-hyperparameters}.

\begin{table}[H]
\centering
\caption{Hyperparameters for reported Tsitsiklis Triangle experiments.}

\label{tab:tsitsiklis-triangle-hyperparameters}
\renewcommand{\arraystretch}{1.2}
\begin{tabular}{ l l c } 
\toprule

Method & Parameter & Value \\

\midrule
\multirow{2}{2.2em}{BBO} & Lower-level learning rate & 8e-1 \\
                         & Upper-level learning rate & 1e-1 \\
\hline
\multirow{1}{2.2em}{TD(0)} & Learning rate & 2e-3 \\
\hline
\multirow{2}{2.2em}{TDC} & Fast timescale learning rate & 1e0 \\
                         & Slow timescale learning rate & 1e-1 \\
\hline
\multirow{2}{2.2em}{GTD2} & Fast timescale learning rate & 8e-1 \\
                          & Slow timescale learning rate & 1e-1 \\
\bottomrule
\end{tabular}
\end{table}

\subsubsection{Results}
As can be seen in~\cref{fig:spiral_result}, BBO converges to optimal solution similarly to prior convergent nonlinear methods, TDC and GTD2~\cite{Maei09}, while TD(0), as expected, diverges. These results verify the convergence properties of the proposed nonlinear BBO algorithms.

\subsection{Neural Network Function Approximators}
\label{app:experimental_details_neural_network_function_approximators}

Most interesting real-world tasks demand use of expressive function approximators such as neural networks. Despite their lack of theoretical convergence guarantees, neural networks have been successfully used in practice for estimating value functions in a wide range of recent reinforcement learning applications. Our proposed gradient BBO algorithms provide provably convergent method for policy evaluation with runtime complexity that is linear in the dimension of the parameter space. In this experiment, we evaluate these properties empirically by applying BBO to a nonlinear regime with neural network function approximators. We use a MAP approximate posterior $p(\phi\vert \Dwn)\approx \delta(\phi = \phi^*_{\Dwn})$ where:
\begin{align}
    \phi^*_{\Dwn} \in \argmin_{\phi\in\Phi}\left(\sum_{i=1}^N \frac{(b_i-\hat{B}_\phi(s_i,a_i))^2}{2\sigma^2} +\frac{1}{\sigma_0^2}\lVert\phi-\phi_0\rVert_2^2\right),
\end{align}
The MAP objective is recommended for policy evaluation as there is no need for the agent to explore, hence learning the posterior uncertainty is inappropriate when a point estimate will suffice. Under a similar derivation as in \cref{app:randomised_priors}, we can minimise the MSBBE with the MAP posterior estimate using the two timescale updates:
\begin{gather}
\phi\leftarrow\mathcal{P}_\Omega \left( \phi-\alpha_k  \nabla_{\phi}\hat{\mathcal{L}}_\textrm{MAP}^i(\phi) \right),\quad \textrm{(fast)}\\
\omega\leftarrow\mathcal{P}_\Omega( \omega-\beta_k (\omega-\phi)),\quad \textrm{(slow)}
\end{gather}
where:
\begin{align}
    &\hat{\mathcal{L}}_\textrm{MAP}^i(\phi)\coloneqq\frac{1}{2}\left(r_i+\gamma \hat{Q}_{\omega}(s'_i,a'_i)- \hat{B}_{\phi}(s_i,a_i)\right)^2  +\frac{1}{\sigma_0^2}\lVert\phi-\phi_0\rVert_2^2.
\end{align}
We refer to algorithms that minimise the MSBBE in this way as \emph{gradient} BBO as they take the posterior's dependence of $\hat{Q}_\omega$ into account. We also test a version where we ignore this dependence that we call \emph{direct} BBO, ignoring the slow update and setting $\hat{B}_\phi=\hat{Q}_\omega$.

We investigate the performance between gradient vs. direct and Bayesian vs. frequentist variants of Gaussian BBO and compare them to prior nonlinear TD(0) and TDC algorithms. We set $\phi_0$ to our initial estimate of the value function parameters, which is initialised using a Glorot uniform initialisation. Our experiments are designed to (1) verify BBO's convergence and consistency properties in nonlinear regime, especially in cases where not all theoretical assumptions are fulfilled exactly (2) investigate the effect of bi-level optimization (i.e. gradient vs. direct methods), and (3) investigate the effect of additional regularization in BBO due to the prior.

\subsubsection{Environments}
We consider three environments commonly used in policy evaluation literature: 20-Link Pendulum~\citep{dann2014policy} with 40D continuous observation space, Puddle World~\citep{Boyan95} with 2D continuous observation space, and the continuous variant of Mountain Car~\citep{Boyan95}\footnote{We use the \href{https://github.com/openai/gym/blob/074bc269b5405c22e95856920e43a067a14302b1/gym/envs/classic_control/continuous_mountain_car.py}{MountainCar-Continuous-v0 implementation} from the OpenAI Gym suite~\citep{brockman2016openai}} with 2D continuous observation space.

\subsubsection{Datasets}

The datasets used for evaluating the policies consist of 20000, 20000, and 30000 on-policy transitions for Puddle World, Mountain Car, and 20-Link Pendulum, respectively. For Puddle World and Mountain Car, each transition is sampled independently by resetting the state uniformly at random in the state space after each transition. Puddle World and Mountain Car observations are normalized to range $[-1, 1]$.

\paragraph{Policies.} For Puddle World and Mountain Car experiments, we run the policy evaluation using a simple, non-optimal policies. In Puddle World, the policy selects either up or down action uniformly at random. In Mountain Car, the right action is chosen when velocity > 0, and otherwise the left action. The 20-Link Pendulum experiments, on the other hand, use an optimal policy obtained with dynamic programming, similar to the policy used in the corresponding linear experiments. See \cite{dann2014policy} for details.

\paragraph{Ground-truth Value Functions.} For Puddle World and Mountain Car, the ground-truth value function, for which the mean squared errors are computed, is obtained for a set of 625 evenly-spaced states (${25\times25}$ grid) in the 2D state space. We reset the agent to each of the 625 states 1000 times, rolling it out and computing the cumulative sum of rewards for up to 1000 steps, finally averaging over the 1000 resets. The 20-Link Pendulum environment is a linear-quadratic MDP, for which we obtain the exact values for 5000 states, as is done by \citet{dann2014policy}. All the value functions are computed with the same discount factor that is used for training the approximate value functions.

\subsubsection{Network Structure}
All the experiments use a single-layer feedforward network with hidden layer size of 256. TDC uses a $tanh$ activation whereas other algorithms use $relu$.

\subsubsection{Training Details}
The discount factor is set to 0.98 for all tasks. The optimization is carried out with Adam optimizer \citep{kingma2014adam} using randomly sampled mini-batches of size 512. The hyperparameter searches are done using 3 datasets and the final results are reported are averaged over 24 separate datasets and seeds.  Each full trial (100k steps) run takes about 15 minutes for BBO variants, 7 minutes for TD(0), and 85 minutes for TDC on standard desktop machine.

\subsubsection{Hyperparameters}

We use the same grid search for the hyperparameters across all environments. The hyperparameters with their evaluated values for each algorithm are presented in \cref{tab:bbo-hyperparameter-grid} for BBO, \cref{tab:td0-hyperparameter-grid} for TD(0), and \cref{tab:tdc-hyperparameter-grid} for TDC. For BBO, instead of searching over the full grid at once, we first perform a coarse grid search for learning rates without using a prior, then do another denser search around the best values (with values still given in the \cref{tab:bbo-hyperparameter-grid}), and finally perform a grid search over the lower-level gradient steps per training steps and prior values (when applicable). For TDC and TD(0), we perform a full grid search over the listed values.
\begin{table*}[h]
\centering
\caption{BBO hyperparameter grid}
\label{tab:bbo-hyperparameter-grid}

     \begin{threeparttable}
\begin{tabular}{l l}
    \toprule
    Hyperparameter &  Grid values \\
    \midrule
      Learning rates & \{ $1\mathrm{e}{-6}$, $3\mathrm{e}{-6}$, $1\mathrm{e}{-5}$, \dots, $1\mathrm{e}{-1}$, $3\mathrm{e}{-1}$, $1\mathrm{e}{0}$ \} \\
      Weight for the prior loss ($1/\sigma_{0}^{2}$)\tnote{*} & \{ $1\mathrm{e}{-4}$, $3\mathrm{e}{-4}$, $1\mathrm{e}{-3}$, \dots, $1\mathrm{e}{-1}$, $3\mathrm{e}{-1}$, $1\mathrm{e}{0}$ \} \\
      Lower-level steps / training step & $\{ 1, 5, 10, 20 \}$ \\
    \bottomrule
\end{tabular}

\begin{tablenotes}
    \item[*] For 20-Link Pendulum, we also include \{ $2.5\mathrm{e}{-1}$, $5\mathrm{e}{-1}$, $7.5\mathrm{e}{-1}$ \}.
\end{tablenotes}
\end{threeparttable}
\end{table*}

\begin{table*}[h]
\centering
\caption{TD(0) hyperparameter grid}
\label{tab:td0-hyperparameter-grid}

\begin{threeparttable}
\begin{tabular}{l l}
    \toprule
    Hyperparameter &  Grid values \\
    \midrule
      Learning rate\tnote{*} & \{ $1\mathrm{e}{-6}$, $3\mathrm{e}{-6}$, $1\mathrm{e}{-5}$, \dots, $1\mathrm{e}{-1}$, $3\mathrm{e}{-1}$, $1\mathrm{e}{0}$ \} \\
    \bottomrule
\end{tabular}

\begin{tablenotes}
    \item[*] For 20-Link Pendulum, we also include \{ $1\mathrm{e}{-7}$, $3\mathrm{e}{-7}$ \}.
\end{tablenotes}
\end{threeparttable}
\end{table*}

\begin{table*}[h]
\centering
\caption{TDC hyperparameter grid}
\label{tab:tdc-hyperparameter-grid}

\begin{tabular}{l l}
    \toprule
    Hyperparameter &  Grid values \\
    \midrule
      Fast timescale learning rate & \{ $1\mathrm{e}{-6}$, $3\mathrm{e}{-6}$, $1\mathrm{e}{-5}$, \dots, $1\mathrm{e}{-1}$, $3\mathrm{e}{-1}$, $1\mathrm{e}{0}$ \} \\
      Slow timescale learning rate & \{ $1\mathrm{e}{-6}$, $3\mathrm{e}{-6}$, $1\mathrm{e}{-5}$, \dots, $1\mathrm{e}{-1}$, $3\mathrm{e}{-1}$, $1\mathrm{e}{0}$ \} \\
    \bottomrule
\end{tabular}
\end{table*}
\begin{table*}[h]
\centering
\caption{Hyperparameters for reported nonlinear policy evaluation results.}
\label{tab:non-linear-policy-evaluation-hyperparameters}

\setlength{\tabcolsep}{4.5pt}
\renewcommand{\arraystretch}{1.3}

\begin{tabular}{ l l c c c } 
\toprule

\multirow{3}{*}{Method}&\multirow{3}{*}{Parameter}&\multicolumn{3}{c}{Environment} \\ \cline{3-5} 
& & 20-Link Pendulum & Puddle World & Mountain Car \\

\midrule
\multirow{5}{4em}{Gradient BBO w/ prior} 
& Upper-level learning rate & 1e-2 & 3e-2 & 1e-2 \\
& Lower-level learning rate & 1e-3 & 1e-2 & 3e-3 \\
& Prior loss weight & 5e-1 & 3e-4 & 1e-1 \\
& Lower-level steps / training step & 20 & 10 & 10 \\
\hline
\multirow{3}{4em}{Gradient BBO w/o prior} 
& Upper-level learning rate & 3e-4 & 1e-3 & 1e-2 \\
& Lower-level learning rate & 1e-3 & 1e-2 & 3e-3 \\
& Lower-level steps / training step & 20 & 10 & 10 \\
\hline
\multirow{3}{4em}{Direct BBO w/ prior} 
& Learning rate & 3e-7 & 1e-3 & 3e-4 \\
& Prior loss weight & 1.0 & 3e-4 & 1e-1 \\
\\
\hline
\multirow{1}{4em}{TD(0)}
& Learning rate & 3e-7 & 3e-3 & 3e-4 \\
\hline
\multirow{2}{4em}{TDC}
& Fast timescale learning rate & 3e-3 & 1e-3 & 1e-2 \\
& Slow timescale learning rate & 1e-4 & 1e-3 & 1e-5 \\
\bottomrule
\end{tabular}
\end{table*}

The final hyperparameters for each method and task are presented in \cref{tab:non-linear-policy-evaluation-hyperparameters}. The values are chosen by manually picking the best-performing set from the hyperparameter search based on the final MSE, and the final results are reported for 24 holdout datasets.

\subsubsection{Results}
\begin{figure}[H]
    \centering
    \includegraphics[width=\linewidth, trim={0mm 0mm 0mm 0mm}, clip]{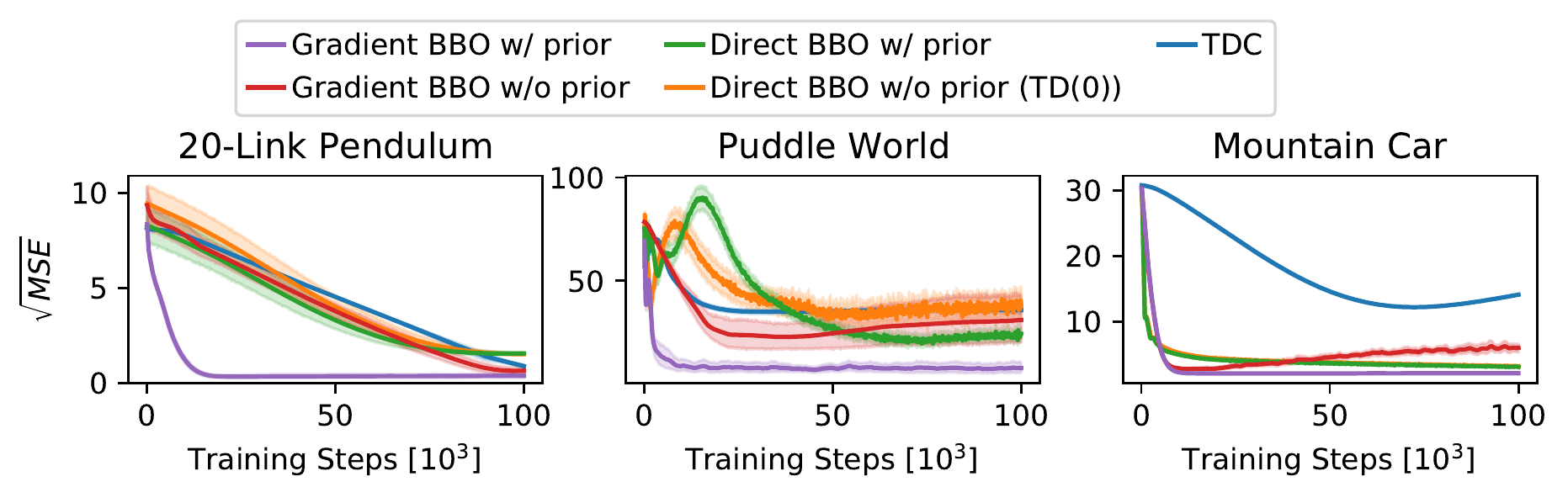}
    \caption{Nonlinear policy evaluation results.}
    \label{fig:non-linear-policy-evaluation-ablation-result-full}
\end{figure}
The results are presented in \cref{fig:non-linear-policy-evaluation-ablation-result-full}. Gradient BBO with a prior quickly converges to a good solution in all tasks, outperforming the non-Bayesian and direct BBO variants as well as other nonlinear methods (note that direct BBO without prior corresponds to TD(0)) across all tasks. Given that both the bi-level optimization and the prior alone perform worse than Bayesian gradient BBO, the speed and quality of the solution can be attributed to their combination. Furthermore, while not a direct measure, the convergence to near-zero MSE provides empirical evidence of the algorithm's consistency.

\section{Continuous Control Experiments}
\label{app:continuous-control-experiments}

This section extends the continuous control experiments in \cref{sec:experiments} and provides further details and analysis of the BBAC algorithm. We refer to $\lambda\coloneqq\frac{1}{\sigma_0^2}$ as the regularisation weight and $\sigma^2$ as the prior scale. We first investigate BBAC's behavior in the \emph{MountainCar-Continuous-v0} task, where the environment's simplicity and low-dimensionality allows us to visually analyze the behavior of the randomized prior ensemble. We then analyze BBAC's sensitivity to randomized prior hyperaparameters in a slightly modified version of DeepMind Control Suite's~\citep{tassa2018deepmind} \mbox{\emph{cartpole-swingup\_sparse}} environment.

To improve the exploration capability of SAC in our challenging domains, we also introduce a variant of SAC called SAC* which uses a single $Q$-function to avoid \emph{pessimistic underexploration} \citep{ciosek2019better}. Here, instead of learning two soft $Q$-function approximators independently and choosing the minimum for the actor and critic gradient updates as specified by SAC, we train a single soft $Q$-function and use updates (6) and (13) of \citet{Haarnoja18} directly. This also reduces the inductive bias that SAC has for solving tasks with dense reward structures, making the comparison against BBAC fairer.

All the experiments in this section use the same hyperparameters from \autoref{tab:bbac_shared_params}. The policies and function approximators are parameterized as fully-connected neural networks.

\begin{table}[H]
\centering
\caption{Common Hyperparameters for BBAC and SAC.}
\label{tab:bbac_shared_params}

\begin{tabular}{l l}
    \toprule
    Hyperparameter &  Value \\
    \midrule
        Optimizer & Adam \\
        Learning rates & $3\mathrm{e}{-4}$ \\
        Discount &  0.99 \\
        Replay buffer size & $1\mathrm{e}{6}$ \\
        Nonlinearity & ReLU \\
        Hidden layers & 2 \\
        Hidden units/layer & 256 \\
        Batch size & 256 \\
        Target smoothing coefficient & $5\mathrm{e}{-3}$ \\
        Training steps per environment step & 1 \\
    \bottomrule
\end{tabular}

\end{table}

\subsection{BBAC MountainCar-Continuous-v0}
\label{app:bbac_mountain_car_continuous_v0}

\begin{figure*}[t]
    \centering

    \includegraphics[width=\linewidth, trim={0mm 0mm 0mm 0mm}, clip]{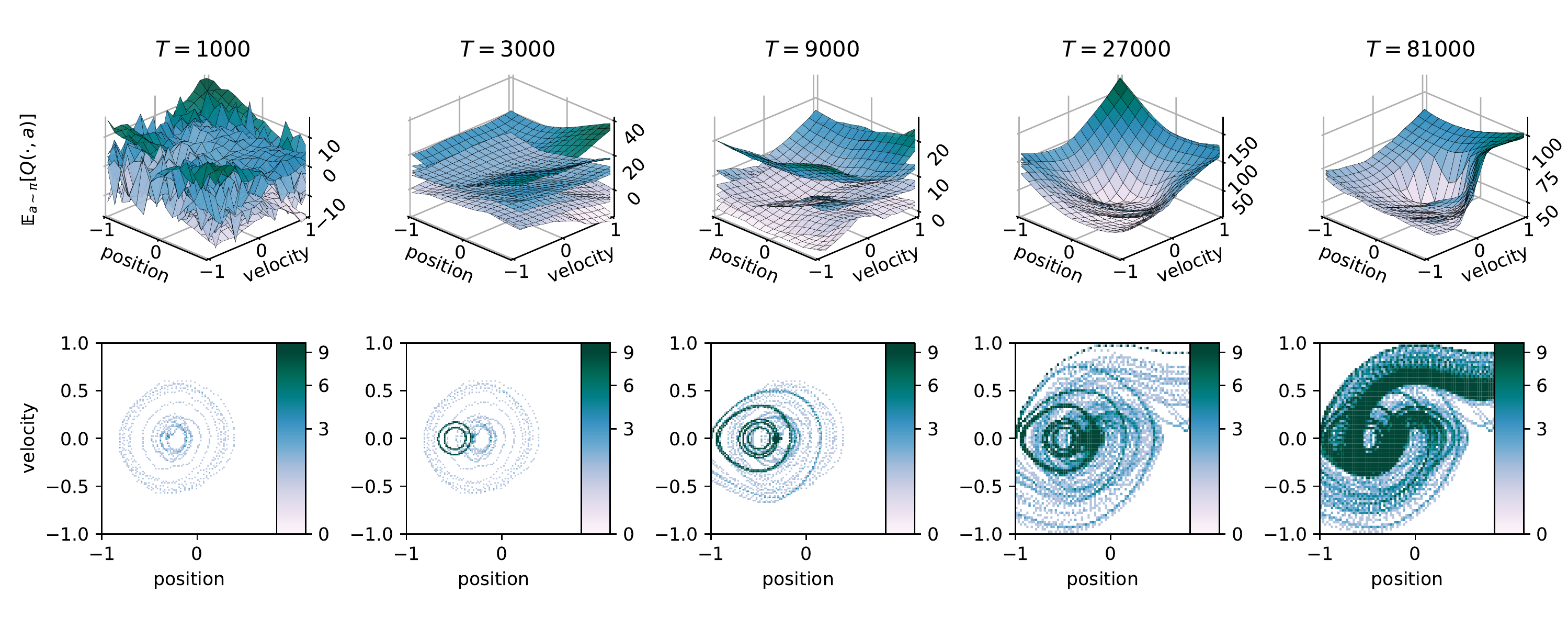}
    \caption{Diagnostics throughout learning of BBAC in MountainCar-Continuous-v0 environment. (Top) Expected Q-values for each ensemble member ($L=8$) over environment states $(position, velocity)$. At $T=9000$, the agent has not yet discovered any reward from the environment, but the disagreement of ensemble members drives the algorithm to deeply explore the state space. At around $T=14000$ (not shown in the plot), the agent achieves the goal for the first time and the value functions start to shape towards optimal solution. (Bottom) State visitation plots show how the covered states evolve during learning. The deep, adaptive exploration carried out by BBAC leads to the agent systematically exploring regions of the state-action space eventually leading to successful task completion.}
    \label{fig:bbac-mountain-car-continuous-v0}

\end{figure*}

\begin{figure*}[t]
    \centering

    \includegraphics[width=\linewidth, trim={0mm 0mm 0mm 0mm}, clip]{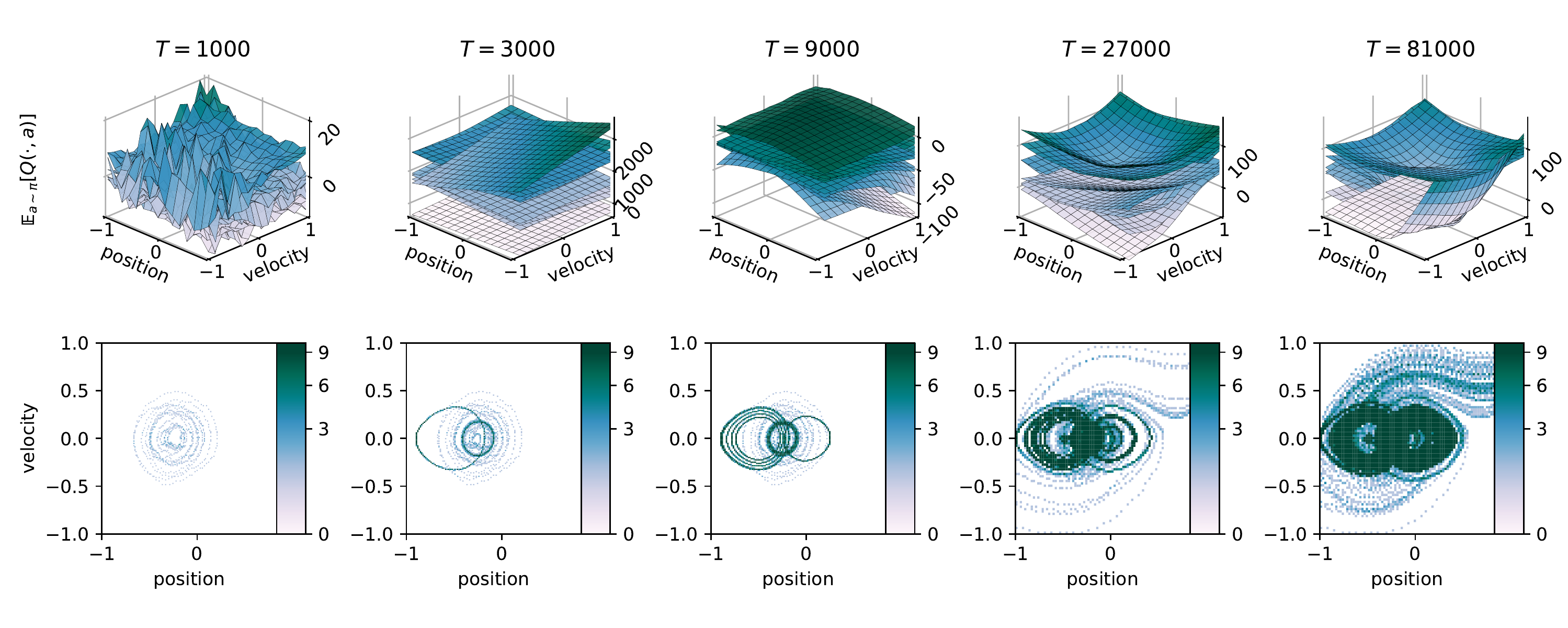}
    \caption{Diagnostics throughout learning of BAC in MountainCar-Continuous-v0 environment. (Top) Expected Q-values for each ensemble member ($L=8$) over environment states $(position, velocity)$. At $T=9000$, the agent has not yet discovered any reward from the environment and the disagreement of ensemble members drives the algorithm to explore the state space similarly as with BBAC (see \cref{fig:bbac-mountain-car-continuous-v0}). As corroborated by \cref{fig:bbac-results:mountain-car-v0}, the Q-function learning is relatively unstable without target networks and the agent spends a lot of time re-exploring the states around origin. (Bottom) State visitation plots show how the covered states evolve during learning. The agent is able to reach the goal but, due to unstable learning, spends more time exploring around the origin than BBAC.}
    \label{fig:bac-mountain-car-continuous-v0}

\end{figure*}

\begin{figure*}[h]
    \centering

    \includegraphics[width=\linewidth, trim={0mm 0mm 0mm 0mm}, clip]{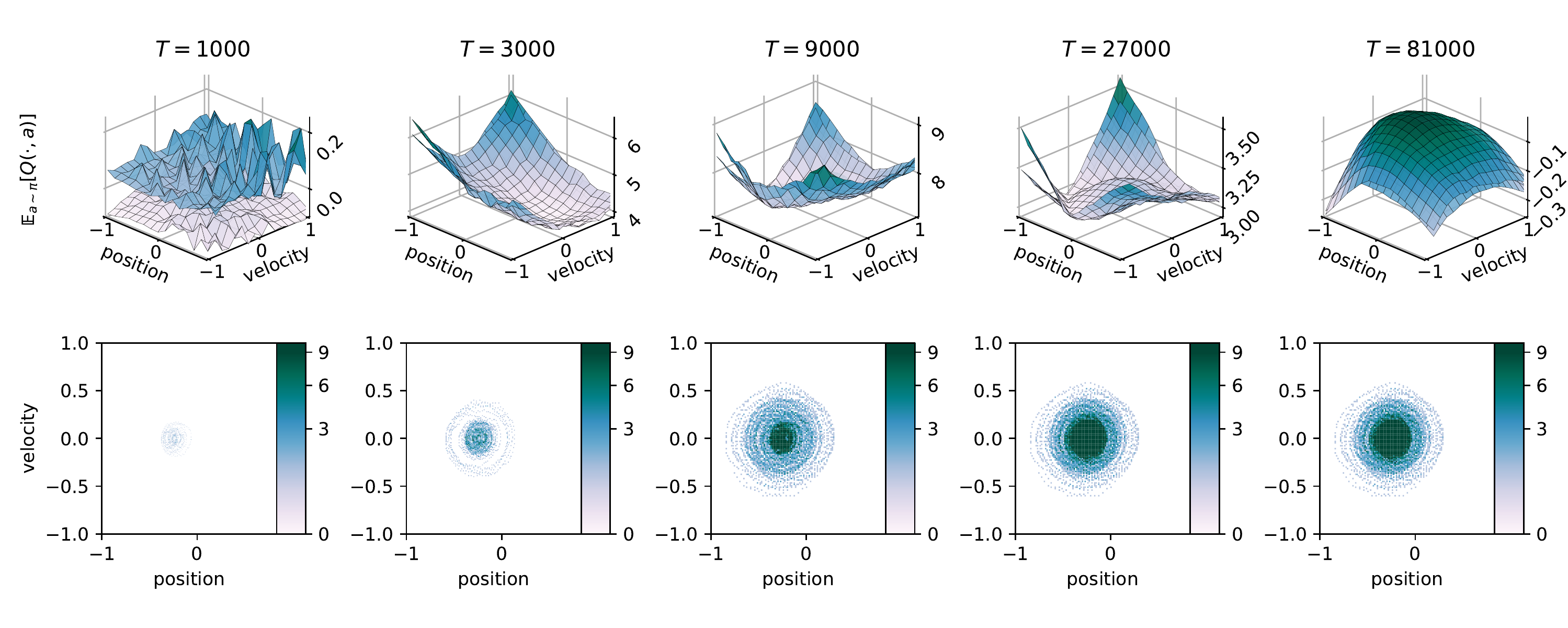}
    \caption{Diagnostics throughout learning of SAC in MountainCar-Continuous-v0 environment. (Top) Ensemble values over environment state $(position, velocity)$. Notice the local maximum in the Q functions around the initial state at $T = 27000$. Similar, but more subtle maximum exists at $T = 9000$. (Bottom) State visitation plots show how the covered states evolve during learning. Due to the na{\"i}ve exploration, SAC repeatedly explores actions that lead to poor performance and rarely explores beyond the initial state.}
    \label{fig:sac-mountain-car-continuous-v0}

\end{figure*}

In order to qualitatively measure how the agents explore the state space, we analyze the evolution of critics and coverage of state space throughout learning. We use the \href{https://github.com/openai/gym/blob/074bc269b5405c22e95856920e43a067a14302b1/gym/envs/classic_control/continuous_mountain_car.py}{standard MountainCar-Continuous-v0 implementation} from the OpenAI Gym suite~\citep{brockman2016openai}, and run each of the algorithms for $1e5$ timesteps, while checkpointing the replay pool and the critics on pre-specified intervals.

For the final SAC runs, we set the target entropy using the heuristic provided in \citep{Haarnoja18c} which is the negative number of action dimensions, i.e. $-1$ in this case. We also tested target entropy values $\{-8, -4, -2, 0, \frac{1}{2}, 1\}$, all of which behaved similarly as the default target. From the 35 runs with these 7 different entropies, a total of 3 seeds were able to achieve the goal. As our experiments are carried out in a tabula rasa 
 setting where the prior functions  are drawn as randomly initialized neural networks, we find that the choice of prior parameters affects the speed with which BBAC solves the tasks.  BBAC is relatively insensitive to the randomized prior hyperparameters in this environment, as long as the ensemble size $L\ge4$, and for the final results we use ensemble size $L=8$, prior scale $\sigma = 100$, and regularization weight $\lambda = 3\mathrm{e}{-5}$.  







The state support analysis, shown in \cref{fig:sac-mountain-car-continuous-v0}, confirms the inefficiency of the exploration typical of RL-as-inference algorithms like SAC: the agent eschews costly actions that would ultimately lead to rewarding states, thus rarely exploring beyond its initial state. This can also be seen in the value functions, which are prematurely driven to sub-optimal solution.

The same analysis for BBAC, in \cref{fig:bbac-mountain-car-continuous-v0}, shows how the deep, adaptive exploration leads to the agent systematically exploring regions of the state-action space with high uncertainty. The value function plots illustrate how the ensemble uncertainty drives the exploration. In the beginning, even when the actions are costly and no positive rewards are encountered, there still exists ensemble members whose value are optimistic under uncertainty. When at least one such optimistic value function exists for a given state, then that state will eventually be explored by the agent, and the agent will only stop exploring states whose uncertainty is driven down by visiting them.

The analysis for BAC in \cref{fig:bac-mountain-car-continuous-v0} confirms that BAC initially explores similarly to BBAC, but due to the convergence issues that stem from ignoring the posterior's dependence on $\omega$, the ensembles never concentrate with increasing number of samples. This means that the approximate posterior cannot characterise the epistemic uncertainty well and the residual stochasticity continues to drive exploration when the agent should have learnt to ignore actions that can't lead to increased returns.

\subsection{BBAC cartpole}
\label{app:bbac_cartpole}

\begin{figure*}[tb]

    \vspace{-3mm}
    \centering

    \begin{subfigure}[t]{0.32\linewidth}
        \centering
        \includegraphics[width=\textwidth, trim={0 0 0 0}, clip]{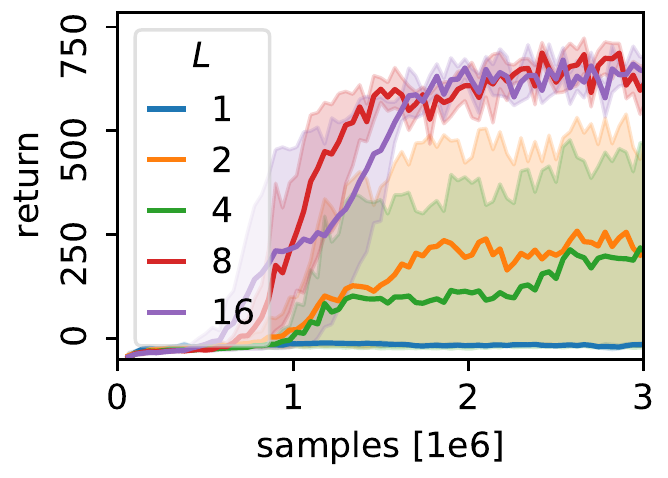}
        \caption{Ensemble size $L$. ${\lambda{=}{3\mathrm{e}{-5}}, \sigma{=}32}$.}
        \label{fig:bbac-ablation:ensemble-size}
    \end{subfigure}
    ~
    \begin{subfigure}[t]{0.32\linewidth}
        \centering
        \includegraphics[width=\textwidth, trim={0 0 0 0}, clip]{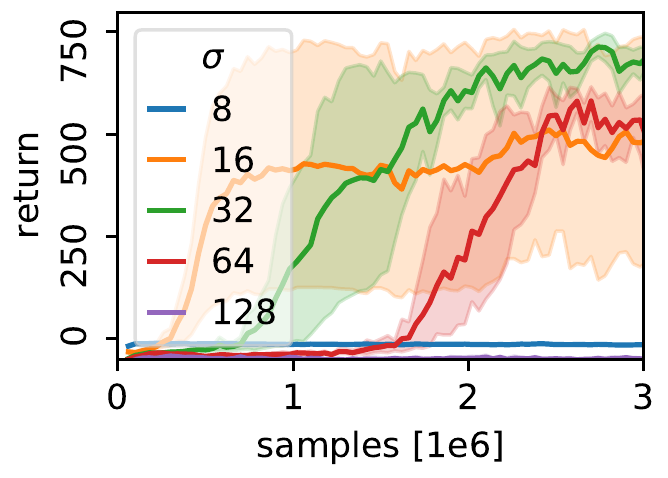}
        \caption{Prior scale $\sigma$. $L{=}8, \lambda{=}{3\mathrm{e}{-5}}$.}
        \label{fig:bbac-ablation:prior-scale}
    \end{subfigure}
    ~
    \begin{subfigure}[t]{0.32\linewidth}
        \centering
        \includegraphics[width=\textwidth, trim={0 0 0 0}, clip]{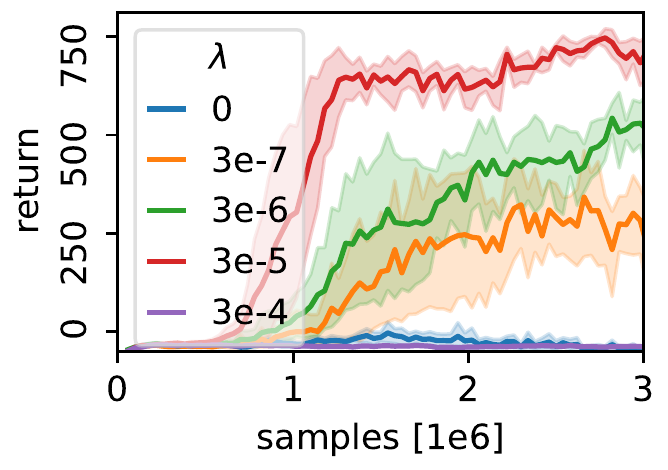}
        \caption{Regularization weight $\lambda$. ${L{=}8, \sigma{=}32}$.}
        \label{fig:bbac-ablation:regularization-weight}
    \end{subfigure}

    \caption{Evaluation of BBAC's sensitivity to randomized prior hyperparameters. We vary (\subref{fig:bbac-ablation:ensemble-size}) ensemble size $L$, (\subref{fig:bbac-ablation:prior-scale}) prior scale $\sigma$, and (\subref{fig:bbac-ablation:regularization-weight}) regularization weight $\lambda$, while keeping other hyperparameters fixed.}
    \label{fig:bbac-ablation}

    \vspace{-3mm}
\end{figure*}

We further investigate BBAC's sensitivity to the choice of randomized prior hyperparameters -- ensemble size ($L$), prior scale ($\sigma$), and regularization weight ($\lambda$) -- in a slightly modified version of DeepMind Control Suite's \citep{tassa2018deepmind} \mbox{\emph{cartpole-swingup\_sparse}} domain, as a continuous analog to the one presented in \citep{Osband19}.

The original \mbox{\emph{cartpole-swingup\_sparse}} is modified such that the reward function includes a control cost of $0.1a_{t}$ for action $a_{t}$ on each timestep $t$, and the agent receives a positive reward of $1$ only when both the cart is controllably centered and the pole is controllably upright. That is, the reward function $r(s_{t},a_{t})$ for action $a_{t}$ at state ${s_{t}=(\cos(\theta_{t}), \sin(\theta_{t}), \dot{\theta}_{t}, x_{t}, \dot{x}_{t})}$ is given by:

\begin{align}
    r(s_{t},a_{t}) = - 0.1|a_{t}| + (\mathbbm{1}(|x_{t}| < 0.1) \cdot\mathbbm{1}(0.95 < \cos(\theta_{t})) \cdot \mathbbm{1}(|\dot{x_{t}}| < 1) \cdot \mathbbm{1}(|\dot{\theta_{t}}| < 1)).
\end{align}

The pole is initialized to a stationary downright position, and the motor is not strong enough to turn the pole upright on a single pass, meaning that in order to reach the goal, the agent has to build momentum by moving the cart and swinging the pole back and forth. This requires executing costly actions for more than a hundred steps, making the exploration problem non-trivial.

SAC's performance, shown in \cref{fig:bbac-results:cartpole}, confirms that na{\"i}ve exploration strategies, such as noisy actions incorporated by maximum-entropy reinforcement learning, eschew costly exploration actions needed for task completion and converge to sub-optimal strategy. BBAC on the other hand is able to consistently solve the task.

The results for BBAC's hyperparameter sensitivity are presented in~\cref{fig:bbac-ablation}. Increasing the ensemble size ($L$;~\cref{fig:bbac-ablation:ensemble-size}) improves the likelihood of solving the task and, as expected, this effect plateaus and the task can be consistently solved with $L\ge8$. In the case of both prior scale ($\sigma$;~\cref{fig:bbac-ablation:prior-scale}) and regularization weight ($\lambda$;~\cref{fig:bbac-ablation:regularization-weight}), too small values limit the exploration and there is a sweet spot where task performance and exploration are well-balanced. While higher values of prior scale are unable to solve the task within the $3e6$ time steps shown here, we expect them to eventually converge to the optimal solution, whereas higher values of regularization weight are likely to constrain the learning too much to allow convergence.

All our experiments are carried out in a tabula rasa setting where the prior functions are drawn as randomly-initialized neural networks, which is why the effect of the hyperparameter choice to the speed of learning is expected. The range of working hyperparameters is relatively wide and easy to tune, however, and in a real-world scenario, we might have access to prior knowledge on the task, for example through transfer learning, which would further simplify the choice of BBAC's hyperparameters.